%% file: main.tex
\documentclass[lettersize,journal]{IEEEtran}
\usepackage[table]{xcolor}
\usepackage{generic}
\usepackage{cite}
\usepackage{amsmath,amssymb,amsfonts}
\usepackage{algorithmic}
\usepackage{graphicx}
\usepackage{algorithm,algorithmic}
\usepackage[hidelinks]{hyperref}
\usepackage{textcomp}

\usepackage{booktabs}
\usepackage{multirow}
\usepackage{subcaption}
\usepackage{tabularx}

\usepackage{dblfloatfix}
\usepackage{afterpage}

\usepackage{titlesec}

\titlespacing*{\section}
  {0pt} 
  {2.0ex plus 0.5ex minus 0.2ex} 
  {1.5ex plus 0.3ex minus 0.2ex} 

\titlespacing*{\subsection}
  {0pt}
  {1.5ex plus 0.3ex minus 0.2ex}
  {1.0ex plus 0.2ex minus 0.2ex}

\usepackage{siunitx}    
\usepackage{caption}
\usepackage{colortbl}
\usepackage{collcell} 
\usepackage{pgfkeys} 
\usepackage{pgfmath} 
\usepackage{array}

\definecolor{baselineBlue}{HTML}{BFD4F2} 
\definecolor{goodGreen}{HTML}{C6EFCE}      
\definecolor{badRed}{HTML}{F4CCCC}         

\newcommand{\base}[1]{\cellcolor{baselineBlue}{#1}}  
\newcommand{\downv}[1]{\cellcolor{goodGreen}{$\downarrow$\,#1}}
\newcommand{\upv}[1]{\cellcolor{badRed}{$\uparrow$\,#1}}

\begin{document}

\title{Comprehensive Evaluation of Large Multimodal Models for Nutrition Analysis: A New Benchmark Enriched with Contextual Metadata}

\author{
    \IEEEauthorblockN{Bruce Coburn\IEEEauthorrefmark{1}, Jiangpeng He\IEEEauthorrefmark{2}\IEEEauthorrefmark{4}, Megan E. Rollo\IEEEauthorrefmark{3}, Satvinder S. Dhaliwal\IEEEauthorrefmark{3}, Deborah A. Kerr\IEEEauthorrefmark{3}, Fengqing Zhu\IEEEauthorrefmark{1}}
    \thanks{
        \IEEEauthorrefmark{1}B. Coburn and F. Zhu are with Purdue University, West Lafayette, IN 47906 USA (e-mail: coburn6@purdue.edu; zhu0@purdue.edu).}
    \thanks{
        \IEEEauthorrefmark{2}J. He is with Indiana University, Bloomington, IN 47405 USA (e-mail: jhe2@iu.edu).}
    \thanks{
        \IEEEauthorrefmark{3}M. E. Rollo, S. S. Dhaliwal, and D. A. Kerr are with Curtin University, Bentley WA 6102, Australia (e-mail: Megan.Rollo@curtin.edu.au; S.Dhaliwal@curtin.edu.au; D.Kerr@curtin.edu.au).}
    \thanks{This study was funded by an Australian Research Council Discovery Project 190101723 entitled ``Accuracy and Cost-Effectiveness of Technology-Assisted Dietary Assessment".}
    \thanks{\ \IEEEauthorrefmark{4}Corresponding author \& Project lead.}
}

\maketitle

\input{00_abstract}

\input{01_introduction}

\begin{figure*}[htbp] 
  \centering

  \newcommand{\subfigureheight}{4.5cm} 

  \begin{subfigure}[t]{0.32\textwidth} 
    \centering
    \includegraphics[height=\subfigureheight, width=\linewidth, keepaspectratio]{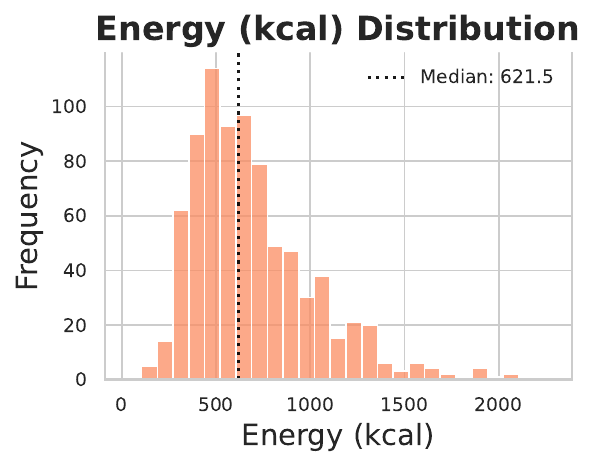}
    \caption{Energy (kcal) Distribution}
    \label{fig:combo_energy}
  \end{subfigure}\hfill
  \begin{subfigure}[t]{0.32\textwidth}
    \centering
    \includegraphics[height=\subfigureheight, width=\linewidth, keepaspectratio]{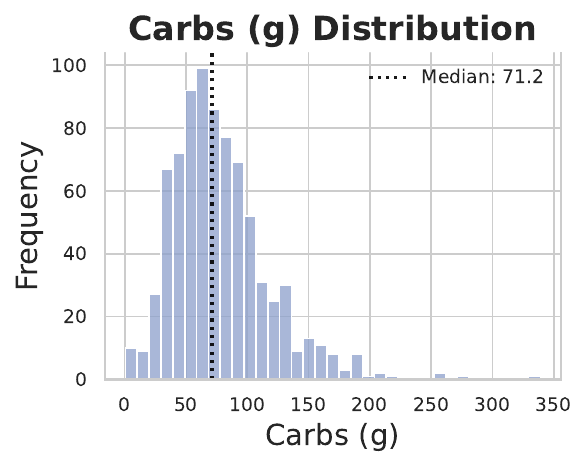}
    \caption{Carbohydrates (g) Distribution}
    \label{fig:combo_carbs}
  \end{subfigure}\hfill
  \begin{subfigure}[t]{0.32\textwidth}
    \centering
    \includegraphics[height=\subfigureheight, width=\linewidth, keepaspectratio]{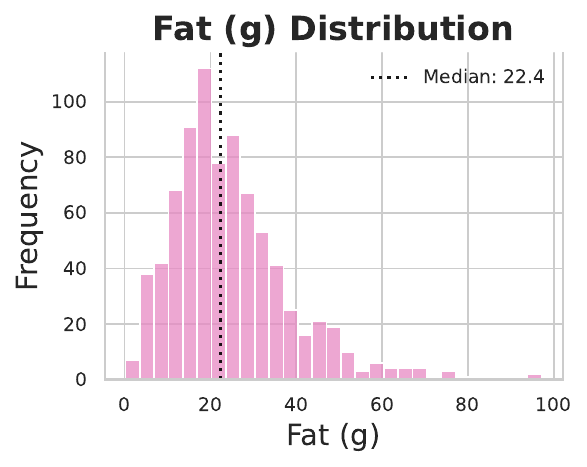}
    \caption{Fat (g) Distribution}
    \label{fig:combo_fat}
  \end{subfigure}

  \vspace{0.4em} 

  \begin{subfigure}[t]{0.32\textwidth}
    \centering
    \includegraphics[height=\subfigureheight, width=\linewidth, keepaspectratio]{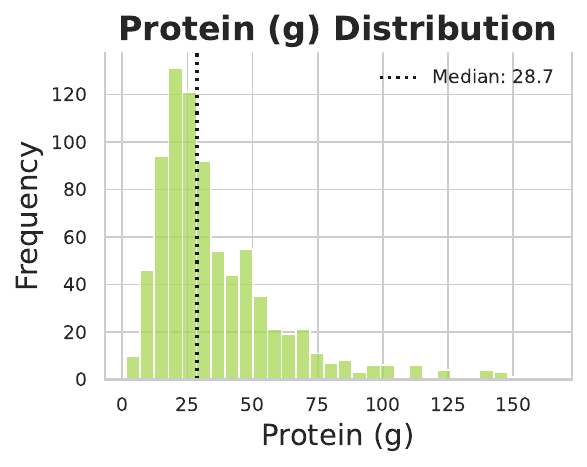}
    \caption{Protein (g) Distribution}
    \label{fig:combo_protein}
  \end{subfigure}\hfill
  \begin{subfigure}[t]{0.32\textwidth}
    \centering
    \includegraphics[height=\subfigureheight, width=\linewidth, keepaspectratio]{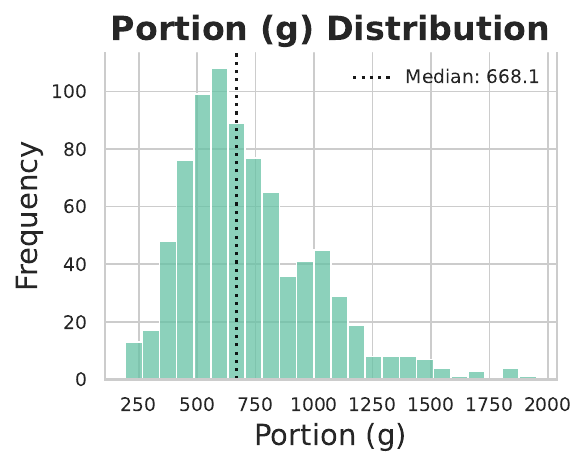}
    \caption{Overall Portion (g) Distribution}
    \label{fig:combo_portion}
  \end{subfigure}\hfill
  \begin{subfigure}[t]{0.32\textwidth}
    \centering
    \includegraphics[height=\subfigureheight, width=\linewidth, keepaspectratio]{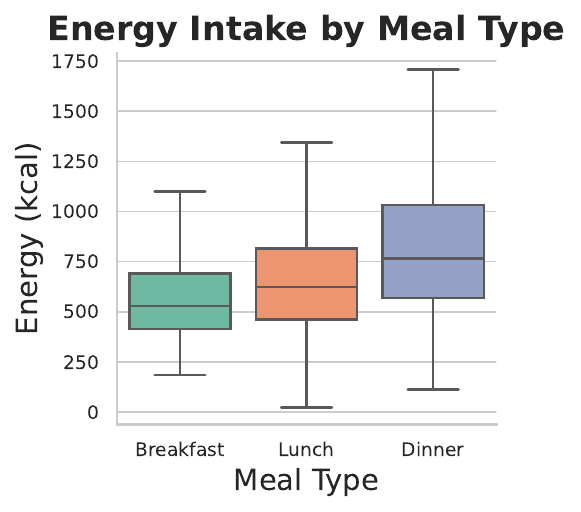}
    \caption{Energy (kcal) by Meal Type}
    \label{fig:combo_kcal_mealtype}
  \end{subfigure}

  \caption{Overview of nutrient, portion weight, and energy distributions in the ACETADA dataset. Subplots (a-e) are histograms where the y-axis represents frequency (count of meals), showing per-meal distributions for: (a) Energy (kcal), (b) Carbohydrates (g), (c) Fat (g), (d) Protein (g), and (e) Overall Portion weight (g). Subplot (f) is a box plot illustrating the distribution of Energy (kcal) for different meal types (Breakfast, Lunch, Dinner), where the y-axis represents Energy (kcal).}
  \label{fig:combined_distributions}
\end{figure*}

\input{02_related-work}

\input{03_methodology}

\input{04_results}

\input{05_discussions}

\input{06_conclusions_and_future_work}

\bibliographystyle{IEEEbib}
\bibliography{references}

\newpage
\clearpage
\input{07_supplemental_material}

\end{document}

%% file: 00_abstract.tex
\begin{abstract}
Large Multimodal Models (LMMs) are increasingly applied to meal images for nutrition analysis. However, existing work primarily evaluates proprietary models, such as GPT-4. This leaves the broad range of LLMs underexplored. Additionally, the influence of integrating contextual metadata and its interaction with various reasoning modifiers remains largely uncharted. This work investigates how interpreting contextual metadata derived from GPS coordinates (converted to location/venue type), timestamps (transformed into meal/day type), and the food items present can enhance LMM performance in estimating key nutritional values. These values include calories, macronutrients (protein, carbohydrates, fat), and portion sizes. We also introduce \textbf{ACETADA}, a new food-image dataset slated for public release. This open dataset provides nutrition information verified by the dietitian and serves as the foundation for our analysis. Our evaluation across eight LMMs (four open-weight and four closed-weight) first establishes the benefit of contextual metadata integration over straightforward prompting with images alone. We then demonstrate how this incorporation of contextual information enhances the efficacy of reasoning modifiers, such as Chain-of-Thought, Multimodal Chain-of-Thought, Scale Hint, Few-Shot, and Expert Persona. Empirical results show that integrating metadata intelligently, when applied through straightforward prompting strategies, can significantly reduce the Mean Absolute Error (MAE) and Mean Absolute Percentage Error (MAPE) in predicted nutritional values. This work highlights the potential of context-aware LMMs for improved nutrition analysis.
\end{abstract}


%% file: 01_introduction.tex
\section{Introduction}
\label{sec:intro}

\IEEEPARstart{I}mage-based nutrition analysis is increasingly being adopted as a practical and automated alternative to self-report methods such as weighed food records, 24-h recalls (ASA24), and manual logging apps (e.g., MyFitnessPal). Early deep-learning systems—Im2Calories \cite{lu2014im2calories}, He \emph{et al.}'s multi-task framework which jointly recognizes foods and infers portion size \cite{He2020MultiTask}, and the smartphone-centric MUSEFood which fuses RGB-D and inertial cues \cite{zhang2020multi}—demonstrated automatic nutrient estimation with a single meal image, resulting in \(\sim\!\)15-20 \% mean error. Building upon such foundational work, the field is increasingly leveraging Large Multimodal Models (LMMs). These advanced models, which can reason over several different modalities of data, such as images and text, offer substantial promise for further advancing image-based nutrition analysis. However, their application faces two primary challenges:

\begin{enumerate}
  \item \textbf{Context vulnerability.} Image-only LMMs often hallucinate portion sizes or misidentify region-specific dishes when crucial cues such as meal time, geolocation, or short ingredient lists are absent \cite{Lo2024Dietary, OHara2021MealPattern, Wang2018,gps_menu}. Dietitians routinely use these signals, yet they are missing from most public datasets and evaluations.
  \item \textbf{Prompt sensitivity.} Model accuracy is highly dependent on how the request is phrased.  Naïve prompts under-exploit the model’s reasoning, whereas expert-persona or chain-of-thought prompts can cut calorie MAE by up to 50\% on small sets \cite{shanahan2023nature, Lo2024Dietary, OHara2025Evaluation}. Prior studies, however, draw on fewer than 200 images, so evidence remains anecdotal.
\end{enumerate}

\noindent
Additionally, there is no open benchmark that (i) pairs dietitian-verified nutrients with precise timestamps and GPS coordinates and (ii) evaluates how modern LMMs exploit that metadata under diverse prompting schemes. Even the newest multimodal nutrition analysis datasets (e.g., MetaFood3D \cite{Chen2024MetaFood3D}) focus on 3-D geometry rather than contextual metadata.

\begin{figure*}[htbp]
    \centering
    \includegraphics[width=\linewidth]{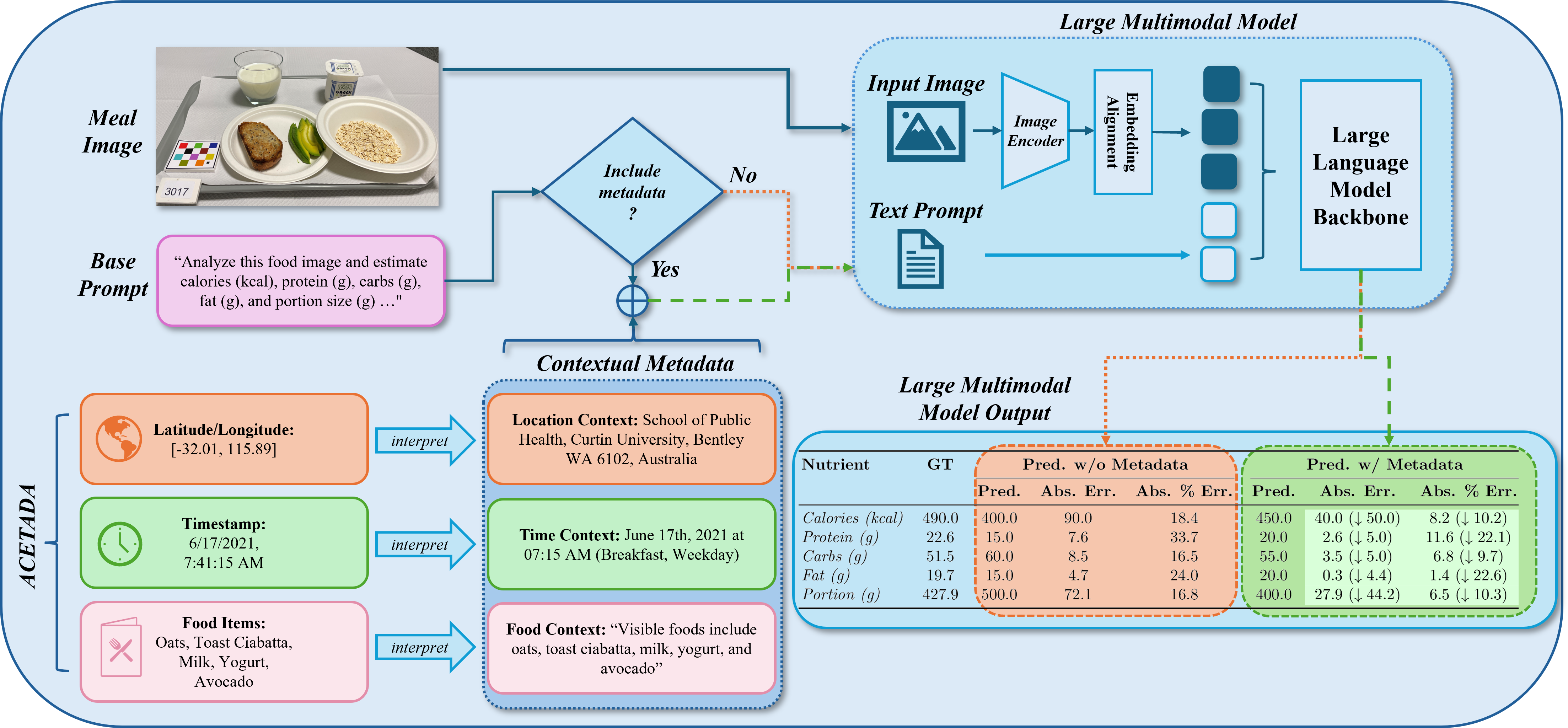}
    \caption{Contextual metadata overview. Location, time, and food context can be combined with the meal photo and the ``base prompt" (``Analyze the food image and estimate ..."). This enriched prompt is passed to an LMM to enhance absolute error and absolute percentage error. In this instance, caloric absolute error and caloric absolute percentage error improve by 100 and 23.42 points, respectively. Aggregated results appear in the Results section.}
    \label{fig:main}
\end{figure*}

\vspace{4pt}
\noindent
Our study aims to address these gaps by making three primary contributions:
\begin{enumerate}
  \item \emph{\textbf{Broad LMM Benchmark}}: We deliver the first evaluation of eight LMMs—four closed-weight APIs (GPT-4o, GPT-4.1, Claude 3 Sonnet, Gemini 2.5 Pro) and four open-weight checkpoints (DeepSeek Janus-Pro-VL, Google Gemma-3-IT, Meta Llama-3.2-VI, Qwen 2.5-VL)—extending performance evidence beyond GPT-4, which is the only model studied extensively to date for nutrition analysis. 
  
  \item \emph{\textbf{Metadata Efficacy Study}}: For every model, we evaluate all combinations of geolocation, timestamp, and individual food items, crossed with five reasoning-oriented prompting methods (CoT, multimodal-CoT, scale-hint, expert persona, few-shot).
  
  \item \emph{\textbf{Open Dataset}}: We plan to release meal images from a controlled-feeding dietary study, providing ground truth data \cite{Whitton2021, Whitton2024}. This dataset details dietitian-verified food and beverage types, their precise gram-level consumed weights, and associated nutrient information confirmed by dietitians. Further enriched with second-level timestamps and GPS coordinates, this will be the first public corpus to unite these four data streams, offering context-aware insights relative to actual intake across breakfast, lunch, and dinner meals.
\end{enumerate}

By systematically disentangling the effects of intelligently interpreted and integrated contextual metadata and diverse prompting strategies across a wide array of LMM architectures, this study provides insights for practitioners aiming to deploy these models for large-scale, automated dietary monitoring. We illuminate how different forms of contextual information and prompting approaches contribute to enhancing accuracy in nutrition analysis.

%% file: 02_related-work.tex
\section{Related Work}
\label{sec:related}

\subsection{Traditional Nutrition Analysis Methods}
Despite rapid progress in computer vision, most population-level nutrition studies still rely on self-report instruments. The ASA24 automated 24-h recall platform \cite{Subar2012} and derivative tools from the National Cancer Institute \cite{NCI2024asa24quickstart} provide low-cost scalability but inherit recall bias.  Mobile food diaries, such as MyFitnessPal, achieve a finer temporal resolution yet show systematic nutrient misreporting relative to weighed records \cite{Boushey2017ReportedEnergy}. Meta-analyses consistently find large under- and over-estimations, motivating new passive approaches \cite{Ravelli2020Traditional}. Controlled-feeding protocols remain the gold standard for validation \cite{hand2019evaluating} but are labor-intensive for both researchers and participants completing the task. Complementary sensing modalities—acoustic, inertial, or biochemical—have been reviewed as promising passive monitors, though they often require wearable form factors and still struggle to quantify portion size \cite{bi2023wearable}.

\subsection{Image-Based Nutrition Analysis}
\label{ssec:image-nutrition}

Early image-based nutrition systems~\cite{Shao2021Integrated} treated each meal image as an isolated monocular vision problem. \emph{Im2Calories} first showed that a deep‐ranking network coupled with hand-crafted portion priors could predict energy from a single image \cite{lu2014im2calories}. Subsequent work introduced explicit geometric cues to tame monocular scale ambiguity: Fang \emph{et al.} compared stereo geometry with commodity depth cameras for portion-size regression \cite{fang2015evaluation}, while \emph{DietCam} used structure-from-motion and ingredient templates to improve volume accuracy in free-living settings \cite{pouladzadeh2017dietcam}. The energy distribution map through generative models is utilized in~\cite{fang2018single, ma2023improved, shao2021towards} to enhance the portion estimation performance.\emph{MUSEFood} later fused inertial and RGB–D signals on smartphones, achieving sub-10 g error for homogeneous foods \cite{zhang2020multi}. The most recent work~\cite{shao2023singleimage3d, vinod2024food, ma2024mfp3d} leverages 3D information of input 2D food images to calculate the portion size. 

To lessen user burden and filter irrelevant frames, Sazonov’s group proposed the \emph{Automatic Ingestion Monitor (AIM-2)}, a smart eyewear device that detects chewing and captures photos only during eating episodes, reducing the image load by 95\% while reaching an 82\% F1 score for meal detection \cite{Doulah2021AIM2, huang2024automatic}. Complementary multimodal strategies have emerged: Mortazavi and colleagues embedded food photos jointly with continuous glucose‐monitor traces, cutting absolute calorie error by roughly 15\% compared with vision-only baselines \cite{Zhang2023JointEmbedding}. Ghasemzadeh’s \emph{DeepFood} system takes a single meal image, detects multiple food items, and outputs per-item nutrient reports, illustrating how deep detectors can automate large parts of the dietary‐logging workflow \cite{RahimiAzghan2025DeepFood}.

Alongside these multimodal efforts, existing image-based dietary assessment methods also focus on addressing core challenges of food data from different real-world perspectives, such as continual learning~\cite{yang2024learning, raghavan2024online, he2021online, he2023long}, long-tailed learning~\cite{he2023longtail, he2023single}, personalization classification~\cite{pan2023personalized, pan2022simulating}, and fine-grained classification~\cite{pan2023hierarchical, mao2021visual, mao2021improving}. Together, these advances in selective image capture, sensor fusion, holistic pipelines, and long-tail learning provide the backdrop for our study, which explores an orthogonal direction: enriching Large Multimodal Models with contextual metadata at \emph{prompt} time to improve nutrition estimation from images.

\subsection{Large Multimodal Models for Nutrition Analysis}

Lo \emph{et al.} provided the first systematic analysis of GPT-4V for nutrition estimation and food classification, highlighting implicit scale reasoning and hallucination pitfalls \cite{Lo2024Dietary}. Kim \emph{et al.} demonstrated that textual descriptors produced by an LMM can be cross-attended with vision transformers to raise Food-101 accuracy by seven percentage points \cite{electronics13224552}. O’Hara \emph{et al.} reported that chain-of-thought and expert-persona prompts halve macro-nutrient error for simple dishes but still under-estimate complex meals \cite{OHara2025Evaluation}.

\begin{figure*}[!t]
  \centering 

  \newcommand{\subfigureheight}{5.5cm} 

  \begin{subfigure}[t]{0.48\linewidth} 
    \centering
    \includegraphics[height=\subfigureheight, width=\linewidth, keepaspectratio]{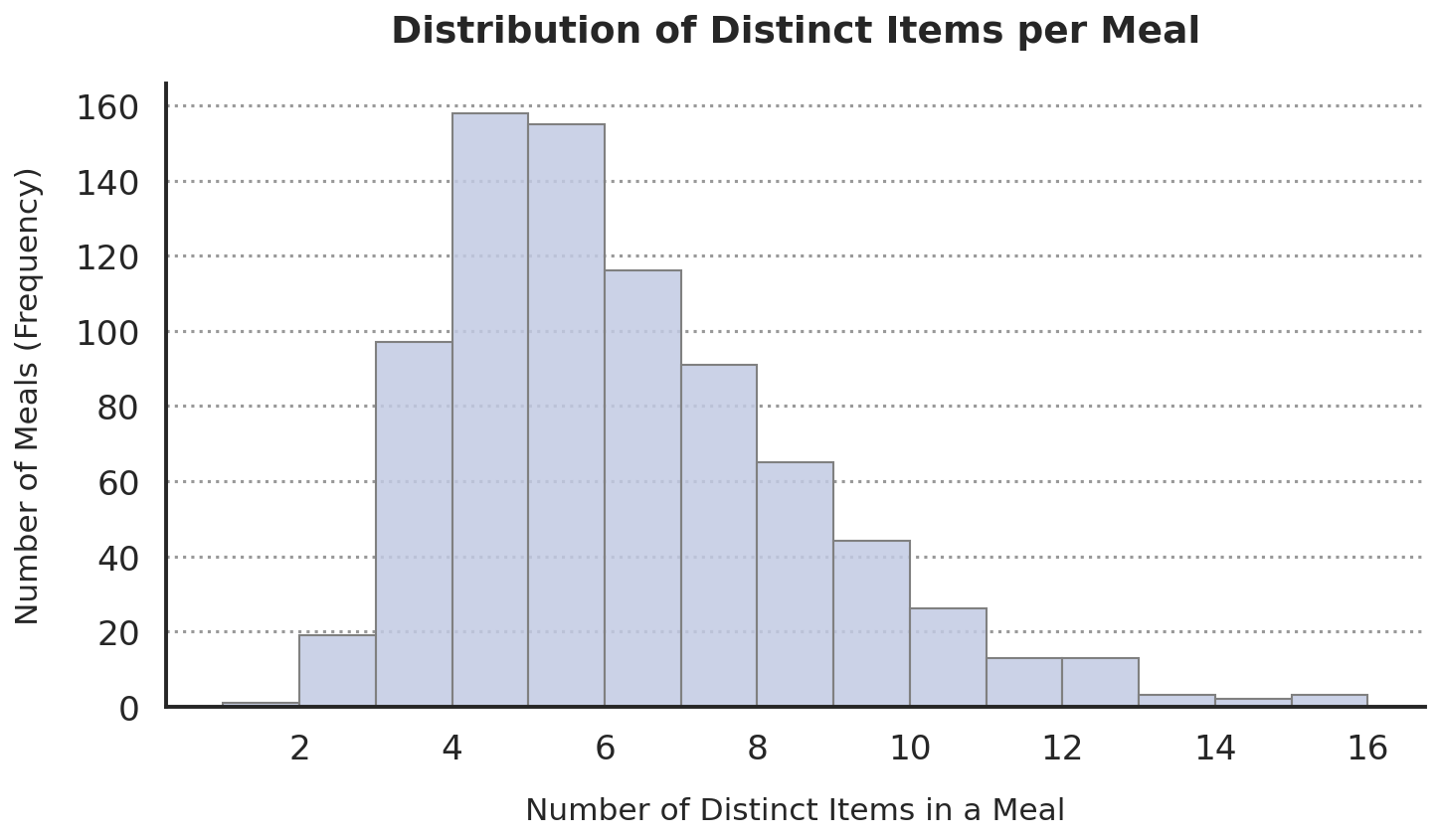}
    \caption{Items per meal distribution.}
    \label{fig:items_per_meal}
  \end{subfigure}\hfill
  \begin{subfigure}[t]{0.48\linewidth} 
    \centering
    \includegraphics[height=\subfigureheight, width=\linewidth, keepaspectratio]{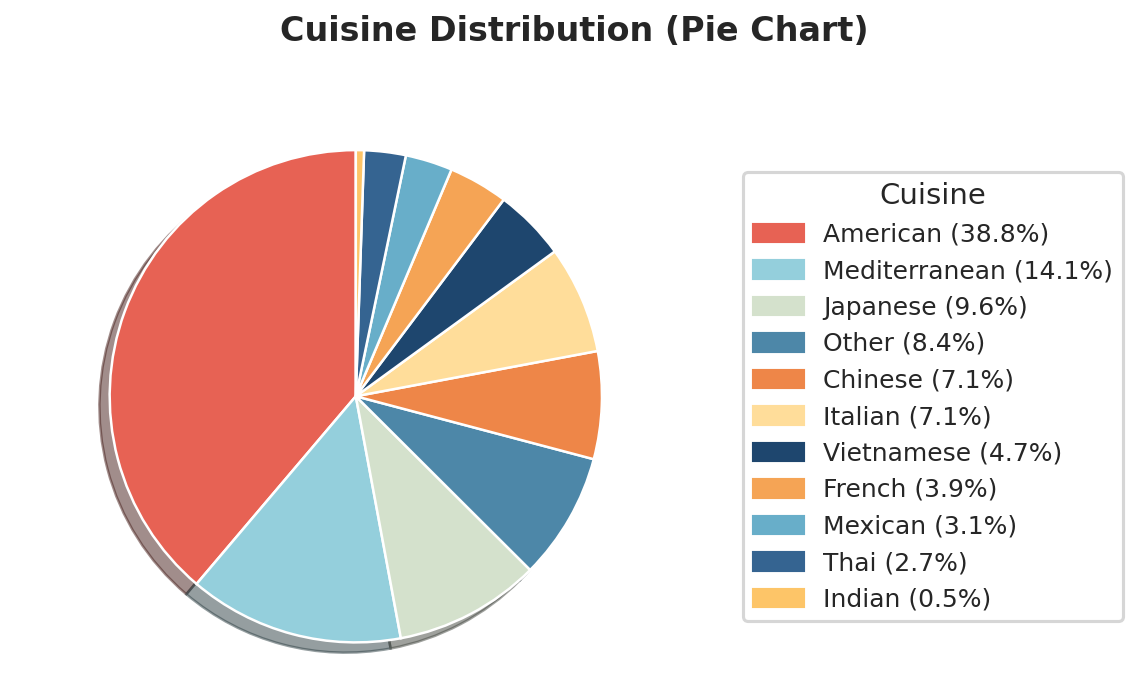}
    \caption{Cuisine category breakdown.}
    \label{fig:cuisine_pie}
  \end{subfigure}

  \caption{Meal composition characteristics in the ACETADA dataset: (a) Histogram showing the distribution of the number of distinct food items recorded per meal. (b) Pie chart illustrating the proportional breakdown of identified cuisine categories.}
  \label{fig:meal_composition_and_cuisine}
\end{figure*}

%% file: 03_methodology.tex
\section{Methodology}

\subsection{The \textsc{ACETADA} Dataset}
\label{sec:dataset}

The \textsc{ACETADA} dietary study is a controlled-feeding, randomised crossover trial that enrolled 152 adults in Perth, Western Australia (55 \% women; \(32 \pm 11\)\,y, BMI \(26 \pm 5\)\, kg m\(^{-2}\))  Over three feeding days, scheduled one week apart, participants consumed laboratory-prepared breakfasts, lunches, and dinners with unobtrusive weighing of foods and beverages consumed to the nearest 0.1g  These weighed records constitute the ``true” intake against which four technology-assisted 24-h dietary‐recall methods were benchmarked: ASA24-Australia, Intake24-Australia, the mobile Food Record with trained-analyst coding (mFR-TA), and an image-assisted interviewer-administered recall (IA-24HR) that reused the images captured for mFR-TA \cite{Whitton2021, Whitton2024}.

\begin{figure*}[!htbp]
  \centering
  \includegraphics[width=\linewidth]{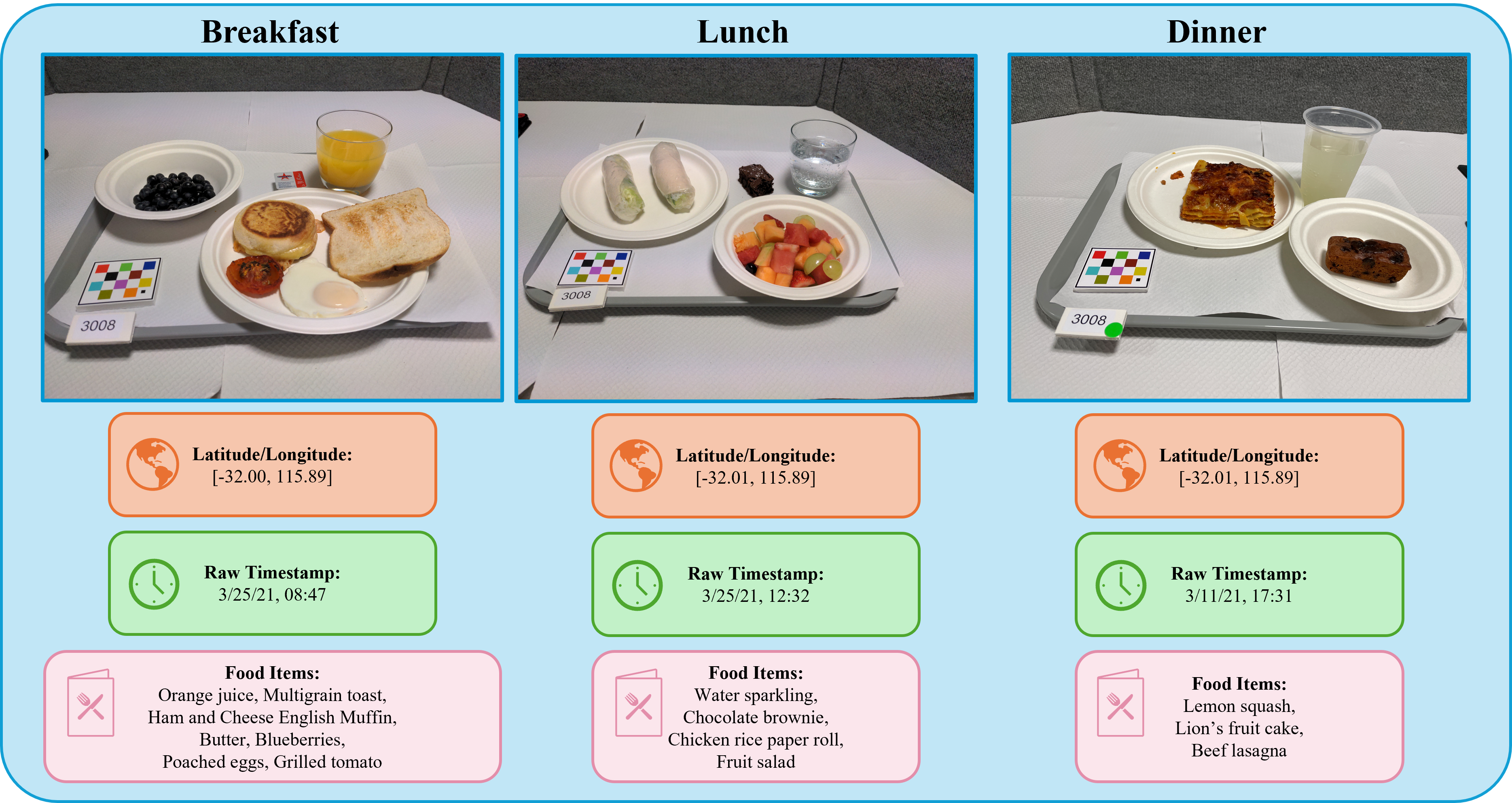}

  \caption{Example of ACETADA images across breakfast, lunch, and dinner with corresponding available contextual metadata. In this instance, images are taken by the same participant.}
  \label{fig:ex_images} 
\end{figure*}

Meal images were acquired with the mFR24 smartphone application \cite{Ahmad2016MFR, Zhu2010ImageAnalysis} immediately before and after consumption  Each frame includes a fiducial marker for scale calibration, the device-local timestamp, and—when available—a latitude–longitude coordinate provided by the on-board GNSS chip  Accredited practising dietitians reviewed every image pair, enumerated the visible foods, and assigned portion weights and macronutrient profiles using AUSNUT 2011–2013 factors \cite{FSANZ2014AUSNUT}  Because the ground-truth labels reflect the consumed mass (served minus leftovers), the pre-meal images represent an upper bound for portion-size estimation tasks.

Our experiments draw on 806 ``before-meal" images—36 \% breakfast, 32 \% lunch, and 32 \% dinner. The paired post-meal frames will be included in the public release, but were unnecessary here because nutrient estimation requires only the full ``before” portion.

Figure~\ref{fig:meal_composition_and_cuisine} summarizes dataset diversity.  Meals contain a median of five items (Fig.~\ref{fig:items_per_meal}) and span 11 cuisines predicted by a zero-shot BART classifier \cite{Lewis2020BART} applied after name normalization (e.g., converting ``Thai Basil Chicken" to ``thai basil chicken" through lowercasing and punctuation removal); the classifier leverages natural-language classification \cite{Yin2019Benchmarking} to effectively classify our cuisine types into predefined cuisine labels (such as American, Chinese, etc.). Example frames with metadata appear in Fig.~\ref{fig:ex_images}. Aggregate nutrient and portion-weight distributions are provided in Fig.~\ref{fig:combined_distributions}.

\textsc{ACETADA} stands out due to three key attributes not jointly available elsewhere: 1) nutrient labels derived from weighed and dietitian-verified measurements, not merely served portions or self-reports; 2) paired pre- and post-consumption smartphone images taken in free-living conditions, each with a fiducial marker, timestamp, and available GPS, enabling rigorous evaluation under realistic user conditions; and 3) the planned public release of all images, annotations, and preprocessing code, establishing an open benchmark.

Alternative datasets were deemed unsuitable. For instance, those used by Lo \emph{et al.} \cite{Lo2024Dietary} and O’Hara \emph{et al.} \cite{OHara2025Evaluation} are not publicly available, hindering reproducible comparisons. Nutrition5k \cite{thames2021nutrition5kautomaticnutritionalunderstanding}, although public, employs studio-based image capture. \textsc{ACETADA} is uniquely positioned as the most suitable testbed for our investigation because it is the only one that combines public access, realistic smartphone image capture in free-living settings, a suite of contextual metadata, and laboratory-verified consumed-mass labels.

\subsection{Prompt Engineering for Contextual Metadata}
\label{ssec:metadata_prompting}

Modern LMMs interpret a mixture of pixels and natural-language instructions, yet their accuracy hinges on how well the prompt surfaces the right priors. Consequently, for our study, we formulate prompt construction as the controlled combination of two sub-methods:

\begin{enumerate}
  \item \textbf{Contextual metadata facets} that situate the image according to
        \emph{geolocation} (\texttt{gps}), \emph{time context}
        (\texttt{timestamp}), and \emph{individual food items}
        (\texttt{food}) - leveraging our ACETADA data.
  \item \textbf{Established prompting methods}—drawn from zero-shot literature—that situate the model’s reasoning style (\texttt{cot}, \texttt{multimodal\_cot}, \texttt{scale}, \texttt{few\_shot}, \texttt{expert}).
\end{enumerate}

The three contextual metadata flags (\texttt{gps}, \texttt{timestamp}, \texttt{food}) append location information, meal-time information, and specific meal components, respectively. These metadata facets further enrich the meal image with context that the image alone may not convey. The five established prompting modifiers provide a modern baseline of established techniques: classic Chain-of-Thought \cite{wei2023chainofthoughtpromptingelicitsreasoning} (\texttt{cot}) elicits step-wise inference; \texttt{multimodal\_cot} grounds each numeric guess in the pixels \cite{zhang2023multimodal}; \texttt{scale} urges explicit size anchoring \cite{lester2021power}; \texttt{few\_shot} provides numerical exemplars \cite{petroni-etal-2019-language}; and \texttt{expert} casts the model as a registered dietitian \cite{shanahan2023nature}. Table~\ref{tab:prompt_parts} summarizes each of our available prompting flags. 

\begin{table}[htbp]
  \centering
  \caption{Prompt flags used in our experiments.}
  \label{tab:prompt_parts}
  \begin{tabular}{@{}llp{4.5cm}@{}}
    \toprule
    \textbf{Flag} & \textbf{Class} & \textbf{Effect / Example}\\
    \midrule
    \texttt{gps}            & Metadata   & Reverse-geocoded venue (“café, Perth, Australia”).\\
    \texttt{timestamp}      & Metadata   & Human-readable time + meal type (“07:43 AM – Breakfast”).\\
    \texttt{food}           & Metadata   & Dietitian-verified list (“scrambled egg, toast”).\\
    \texttt{cot}            & Reasoning  & Classic Chain-of-Thought cue.\\
    \texttt{multimodal\_cot}& Reasoning  & Step-wise \emph{visual} CoT scaffold.\\
    \texttt{scale}          & Reasoning  & Ask model to list size references.\\
    \texttt{few\_shot}      & Reasoning  & Provide two worked exemplars.\\
    \texttt{expert}         & Reasoning  & Prefix: “You are a nutrition expert…”.\\
    \bottomrule
  \end{tabular}
\end{table}

\begin{figure}[htbp]
  \centering
  \includegraphics[width=\linewidth]{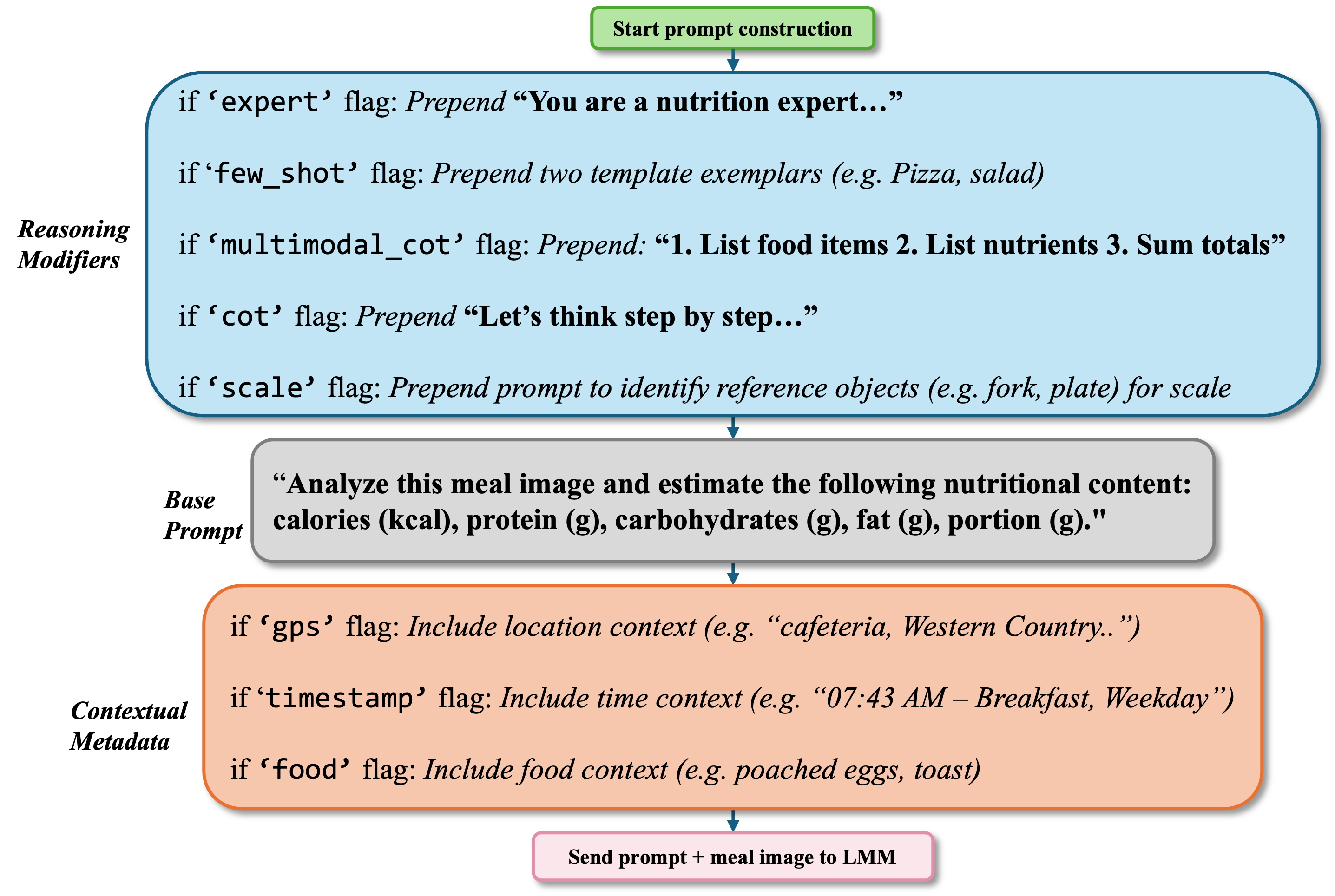}
  \caption{Prompt-construction flowchart.  
           \textbf{Metadata flags} (orange)—\texttt{gps}, \texttt{timestamp}, \texttt{food}—and \textbf{reasoning modifiers} (blue)—\texttt{cot}, \texttt{mmcot}, \texttt{scale}, \texttt{fewshot}, \texttt{expert}—optionally augment a base nutrition-analysis prompt (grey) before being sent, with the meal image, to the LMM.}
  \label{fig:prompt-flow}
\end{figure}

Figure \ref{fig:prompt-flow} visualizes our prompt construction  Metadata flags (orange) can be utilized to append context to the base prompt, whereas reasoning modifiers (blue) can prepend the base prompt  This enhanced prompt is then merged with a concise base nutrition analysis prompt (gray) and the corresponding meal image before being sent to the LMM  The fixed ordering guarantees byte-identical token sequences across models so that any performance gap can be attributed to the model itself rather than prompt permutation effects.

Every experimental run is launched by specifying a compact \texttt{scheme} string (such as \texttt{gps+timestamp+cot}).  The backend wrapper parses this string, instantiates the pipeline in Fig.~\ref{fig:prompt-flow}, and ensures that all models receive byte-identical text payloads and identical image resolutions. This isolation allows us to measure the intrinsic value of each metadata facet and prompting method in terms of nutrition analysis accuracy.

%% file: 04_results.tex
\section{Results}
\label{sec:results}

This section quantifies the impact of contextual metadata and established prompting methods on the accuracy of nutrition and portion-size estimation. For our evaluation, we primarily conduct two experiments. Experiment 1 isolates the stand‑alone benefit of metadata, while Experiment 2 explores how that benefit compounds when metadata is combined with the reasoning-modifying methods introduced in Section~\ref{ssec:metadata_prompting}. 

\subsection{Experimental Setup}
\label{ssec:exp-setup}

All evaluations are conducted on all GPS-valid meal images of the ACETADA dataset. Each image is paired with a text prompt depicted in Figure \ref{fig:main}. For open‑weight checkpoints we use \texttt{transformers v4.41} with bfloat16 activations and restrict GPU memory usage to 90 \% on one or two NVIDIA H100 cards, depending on model size. Closed‑weight models—GPT‑4o, GPT‑4.1, Claude 3.7 Sonnet, and Gemini 2.5 Pro— are queried at temperature 0.1 with the most recent checkpoint available at the time of writing. Default values of vendor-specific tokenizers, image-resolution pipelines, and generation hyperparameters remain untouched, so that any performance difference can be attributed to the incorporation of contextual metadata prompts rather than low-level tuning. Following model inference, error is reported as Mean Absolute Error (MAE) and  Mean Absolute Percentage Error (MAPE) for each nutritional attribute (kilocalories, protein, carbohydrates, fat, portion). 

\subsection{Model Suite}
\label{sec:models}

To understand how openness, parameter count, and vision-tower architecture shape a model’s ability to infer calories and macronutrients from images, we evaluate eight modern LMMs. Each model takes an RGB meal image and a text prompt as inputs, and returns a single natural-language response containing the model's estimates.

\textbf{Closed-weight APIs.} 
We adopt GPT-4o as our high-end baseline model. Released in May 2024, GPT-4o unifies vision and language in a single network, offers a 128 k-token context window, and delivers a higher throughput at a ~50\% lower cost than the retired ``gpt-4-vision" checkpoint \cite{OpenAI_GPT4o_System_Card}. The earlier GPT-4 Vision checkpoints were the model evaluated in the two prior nutrition analysis studies by O'Hara et al.\ and Lo et l.\ \cite{OHara2025Evaluation, Lo2024Dietary}; it is no longer accessible to new API users. Using GPT-4o updates the state of the art and preserves reproducibility for future evaluations. GPT-4.1 extends the same architecture to a one-million-token window and cuts inference cost by 26 \%~\cite{OpenAI_GPT4_1}. Claude 3.7 Sonnet~\cite{Anthropic_Claude_3_7_Sonnet_2025} offers a 200k window and markedly low hallucination rates. Gemini 2.5 Pro combines a sparse Mixture-of-Experts backbone with a high-capacity vision encoder and supports up to one million tokens, allowing the test of whether stronger visual features can offset limited contextual metadata \cite{gemini25pro}.

\textbf{Open-weight checkpoints.} 
At the lightweight end, Janus-Pro-VL (7B) runs on a single 24 GB GPU and is designed for efficient vision–language inference \cite{chen2025januspro}. Mid-scale options include Gemma-3-IT (27B), which introduces multimodal support with a long 128 k-token context window \cite{kamath2025gemma3}. At the high end, LLaMA-3.2-VI (90B) and Qwen 2.5-VL (72B) both offer powerful vision backbones and large language capacity; Qwen 2.5-VL in particular supports multilingual reasoning and is designed for fine-grained image–text alignment \cite{meta2024llama32vision, bai2025qwen25vl}.

\subsection{Evaluation Protocol and Metrics}
\label{ssec:metrics}

For each nutritional attribute $y\in\{\text{kcal},\text{protein},\text{carbs},\text{fat},$ $\text{portion}\}$ we report the mean-absolute error and the mean-absolute percentage error:
\[
  \mathrm{MAE}(y)=\frac{1}{N}\sum_{i=1}^{N}\lvert y_i-\hat y_i\rvert,
\qquad
  \mathrm{MAPE}(y)=\frac{100}{N}\sum_{i=1}^{N}\Bigl\lvert \frac{y_i-\hat y_i}{y_i}\Bigr\rvert,
\]
where $N$ is the number of evaluation images, $y_i$ the dietitian ground truth, and $\hat y_i$ the model prediction.  MAE reflects absolute deviation, while MAPE normalises this error by the true value.

Unless otherwise noted, Section~\ref{sec:results} shows macro-averaged scores rather than per-model results. Specifically, for any subset of models $M$ (all models, only open-weight, only closed-weight, or any set of prompting methods), the scheme-level MAE is:
\[
  \overline{\mathrm{MAE}}_{\text{scheme}}
  =\frac{1}{|M|}\sum_{m\in M}\;
   \frac{1}{|N|}\sum_{i\in\mathcal{I}}
   \lvert y_{i}-\hat y_{i}^{(m)}\rvert,
\]
with a similar definition for MAPE. These metrics form the basis of all tables and radar plots found in the Results section.

\subsection{Experiment 1: Impact of Metadata on Simple Nutrition Analysis Prompting}
\label{ssec:exp1}

\input{tables/experiment1}

We first quantify how much contextual metadata can support LMMs in a straightforward prompting setting. For every model, we compared a baseline that showed the meal image plus a simple nutrition estimation prompt, further modified with metadata-aware variants by toggling the \texttt{gps}, \texttt{timestamp}, and \texttt{food} flags. The table reports, for each model, the combination that produced the largest reduction in MAE and, when tied, in MAPE. 

Across the eight models, adding context consistently lowered calorie error. The average decrease in \emph{calorie MAE} was ${\sim}76\,$ kcal. The best case was Janus-Pro with the \texttt{gps+timestamp} flag, lowering calorie MAE by 246 kcal and calorie MAPE by 52 percentage points. Portion size estimates also improved substantially, dropping by ${\sim}53\,$g on average and by 124 g for LLaMA-3.1-Vision-Instruct. Protein, carbohydrate, and fat benefited more modestly, with typical MAE reductions of 2-5 g; the only notable regression was a 0.68 g rise in protein MAE for Janus-Pro, which is negligible compared the positive impact on its calorie MAE.

Performance patterns varied by model family. Among the open-weight systems, Janus-Pro and LLaMA recorded the most dramatic improvements. Closed-weight APIs showed more moderate yet consistent benefits: Gemini 2.5-Pro lowered every nutrient metric, Claude-3.7-Sonnet posted double-digit drops for both calorie and portion errors, and the two GPT-4 variants registered steady 12-17 kcal calorie reductions. A single adverse effect surfaced for GPT-4o, whose carb MAE rose by 1 kcal even as all of its other metrics improved.

Despite these differences, every ``best metadata" combination contained either \texttt{gps} or \texttt{timestamp}, suggesting that location-based and meal-time cues are the most universally valuable annotations. Open-weight models tended to prefer either the \texttt{gps} or \texttt{gps+timestamp} incorporations, whereas closed-weight models each performed best when food items were a commonality. Taken together, the results establish that even without elaborate reasoning chains, contextual metadata integration provides a measurable boost to nutrition estimates.

\begin{figure*}[t]
  \centering
  \begin{subfigure}[t]{0.32\textwidth}
    \centering
    \includegraphics[width=\linewidth]{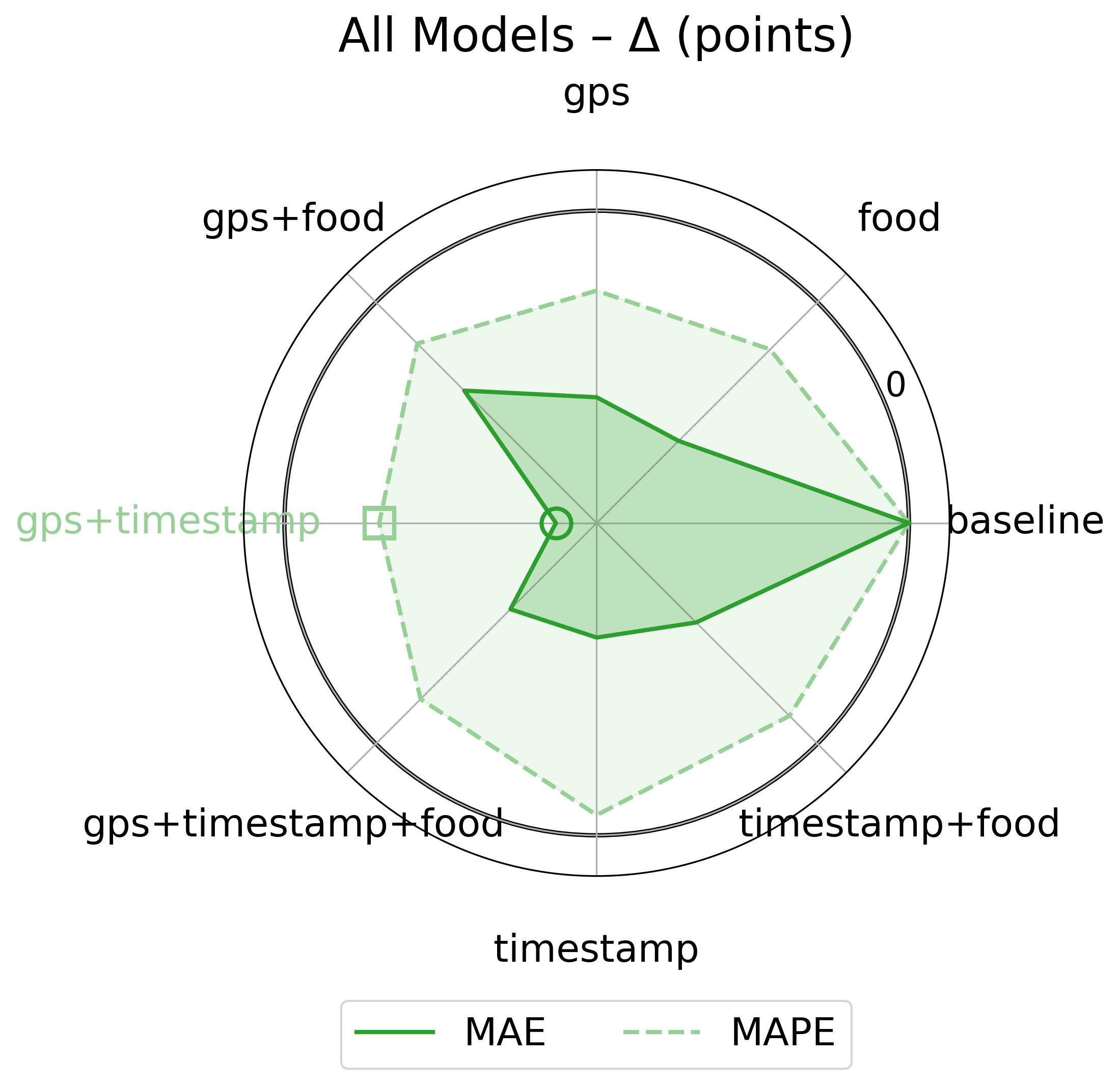}
    \caption{All models}
    \label{fig:radar_all_models}
  \end{subfigure}
  \hfill
  \begin{subfigure}[t]{0.32\textwidth}
    \centering
    \includegraphics[width=\linewidth]{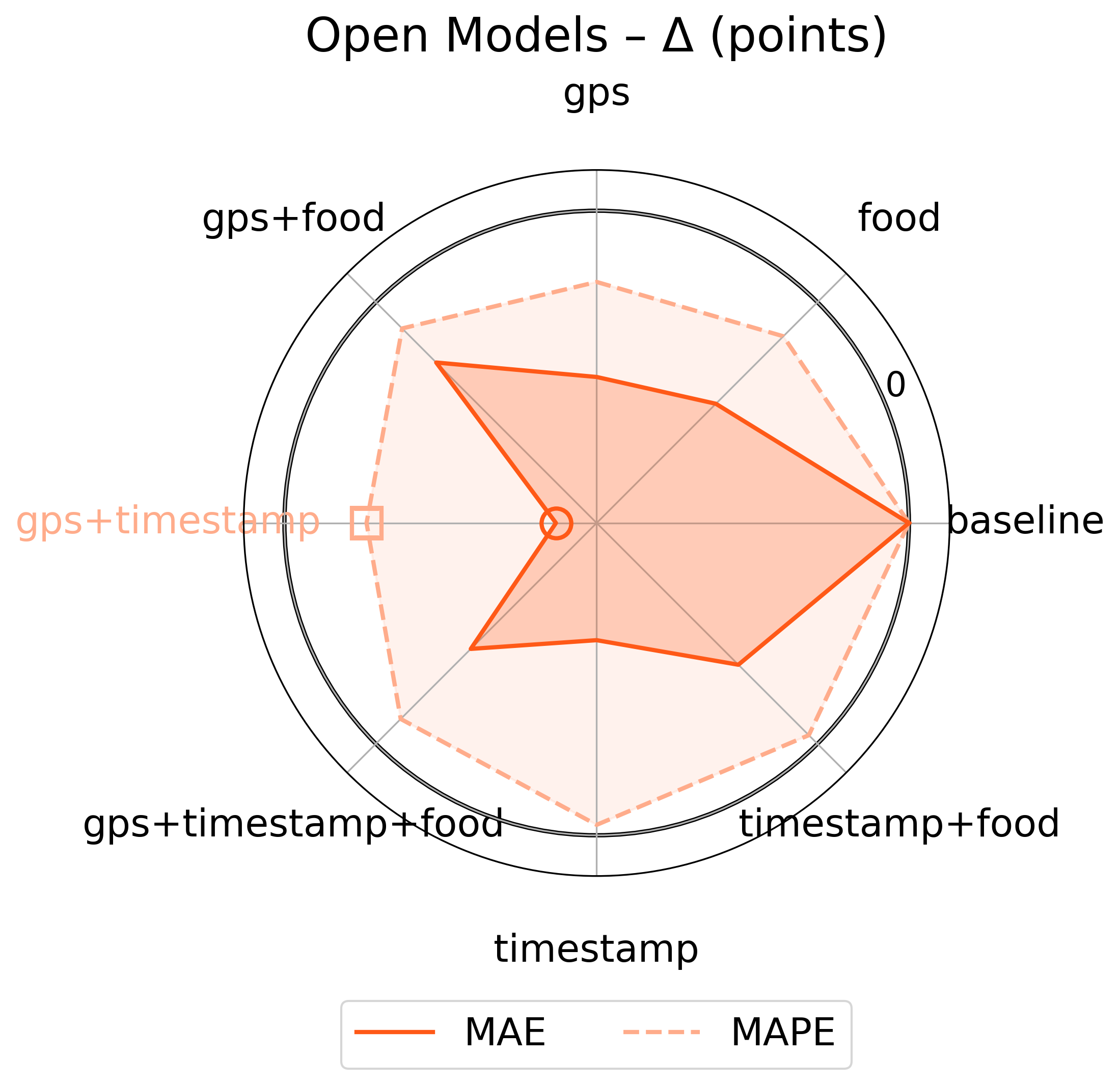}
    \caption{Open-weight checkpoints}
    \label{fig:radar_open_models}
  \end{subfigure}
  \hfill
  \begin{subfigure}[t]{0.32\textwidth}
    \centering
    \includegraphics[width=\linewidth]{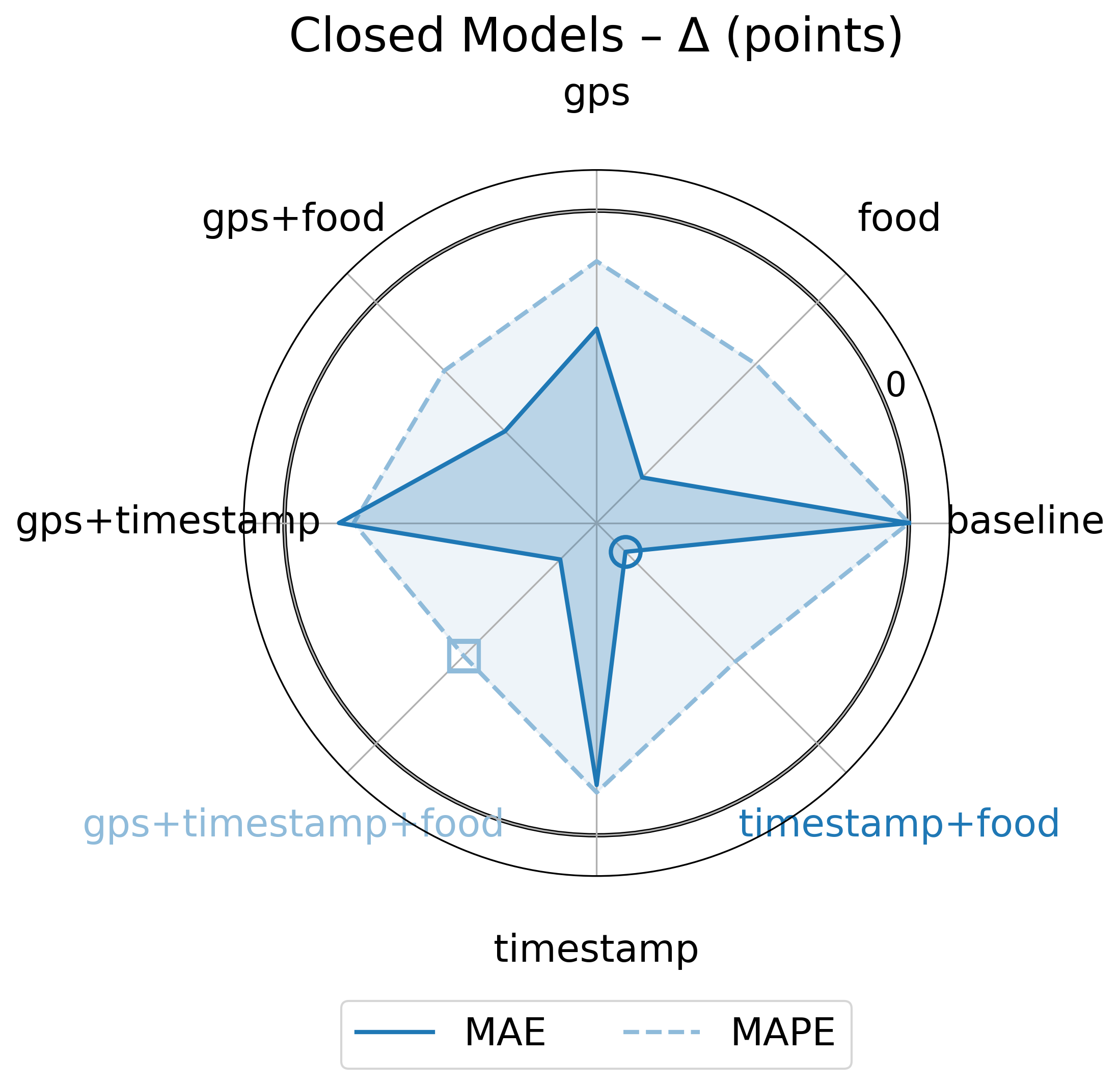}
    \caption{Closed-weight APIs}
    \label{fig:radar_closed_models}
  \end{subfigure}

  \vspace{4pt}
  \caption{\textbf{Averaged Experiment 1 Results According to Weight-Type.} MAE (solid lines) and MAPE (dashed lines) are plotted for various contextual metadata combinations. Each spoke represents a metadata combination's error; closer proximity to the center signifies a reduction in error relative to the baseline prompt. Colored markers denote the \textsc{Best-Metadata} configuration for each metric.}
  \label{fig:exp1_radar_suite}
\end{figure*}

\subsection{Experiment 2: Impact of Metadata on Reasoning Modifiers for Nutrition Analysis}
\label{ssec:exp2}

To determine whether the contextual block introduced in Section~\ref{ssec:exp1} can complement more sophisticated prompting strategies, we paired it with five widely used reasoning modifiers: \emph{Chain-of-Thought} (CoT), \emph{Multimodal CoT}, \emph{Scale-Hint}, \emph{Few-Shot Exemplars}, and \emph{Expert-Persona}.  For each modifier, we compared a baseline prompt, which contained only the meal image and the modifier-specific baseline prompt, with a metadata-enriched prompt where metadata was included from all combinations of \texttt{gps},\texttt{timestamp}, and \texttt{food}.  The most effective combination for every modifier is reported in Table~\ref{tab:rel_change_abridged_metadata}.

Across all five modifiers, attaching contextual metadata lowered calorie-prediction error.  The decrease in calorie MAE ranged from $21.31$ kcal (Few-Shot) to $75.39$ kcal (Expert-Persona), while the associated reduction in calorie MAPE spanned $4.26$ to $10.38$ percentage points.  Portion-size estimates also benefited, falling by $7.72$ g for the few-shot prompt and by as much as $45.28$ g for the expert persona prompt. Only one adverse effect was observed: the \emph{Multimodal CoT} prompt incurred a marginal increase of $0.25$ g in protein MAE.  Similarly, the \emph{Scale-Hint} prompt raised carbohydrate MAPE by a modest $0.48$ percentage points, but still reduced all other nutrient metrics.

The size of the benefit depended strongly on the modifier and on the metadata string that paired best with it. The \emph{Expert-Persona} template achieved the largest calorie reduction with the full \texttt{gps+food+timestamp} string, indicating that persona-style reasoning can exploit context. \emph{Multimodal CoT} required only \texttt{gps+timestamp} to cut calorie error by $64.46$ kcal, whereas standard CoT derived its $51.08$ kcal improvement from the \texttt{food+timestamp} pair. The exemplar prompt benefited primarily from geolocation alone, and the scale-hint prompt again preferred the full three-facet string. Despite these differences, the best-performing variant for every modifier contained at least one of the \texttt{gps} or \texttt{timestamp} flags, underscoring the importance of meal-time and location cues for effective nutrition estimation with large multimodal models.

\begin{figure*}[t]
  \centering
  \begin{subfigure}[t]{0.30\textwidth}
    \centering
    \includegraphics[width=\linewidth]{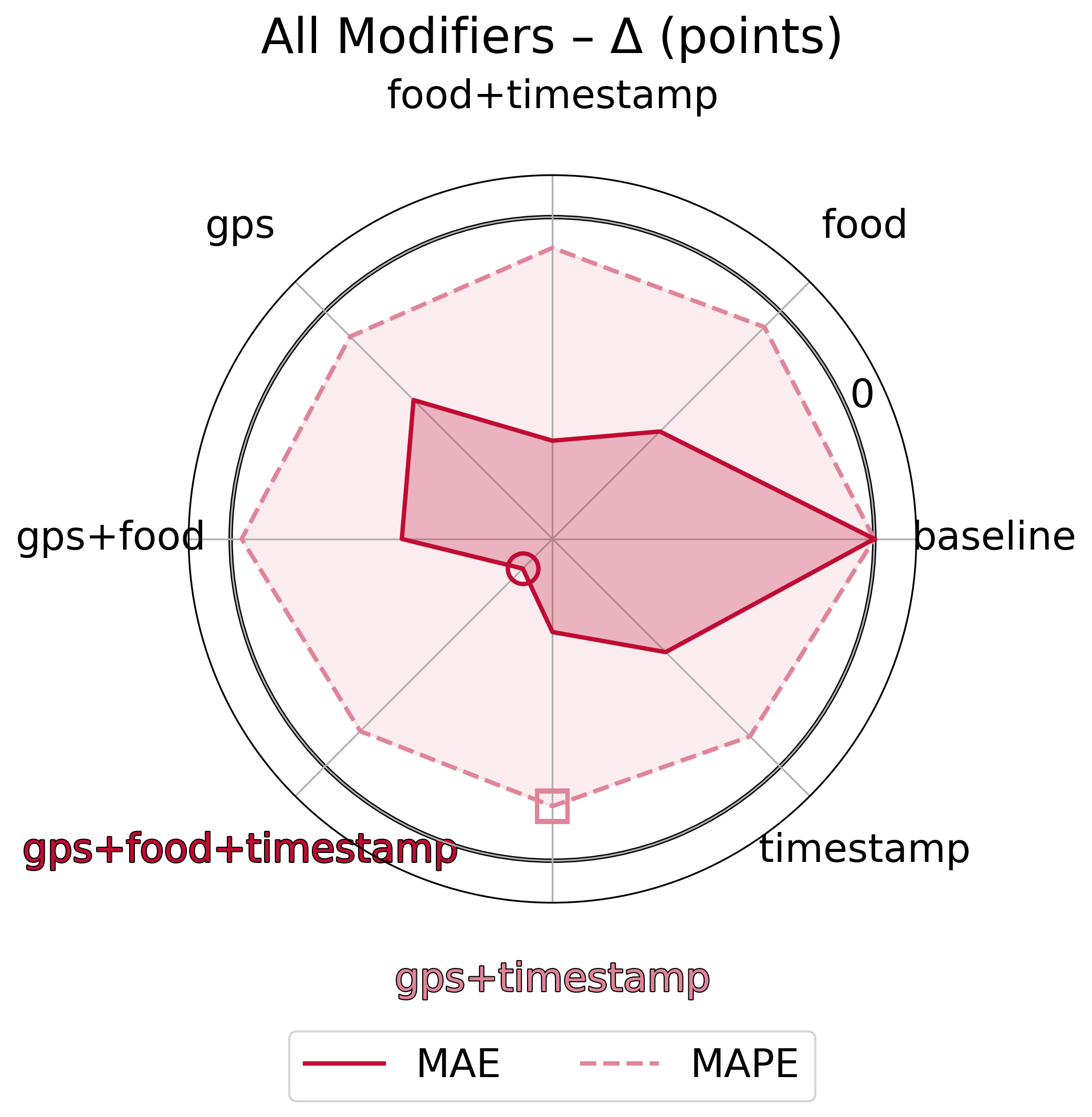}
    \caption{All reasoning modifiers combined}
    \label{fig:exp2_all}
  \end{subfigure}\hfill
  \begin{subfigure}[t]{0.30\textwidth}
    \centering
    \includegraphics[width=\linewidth]{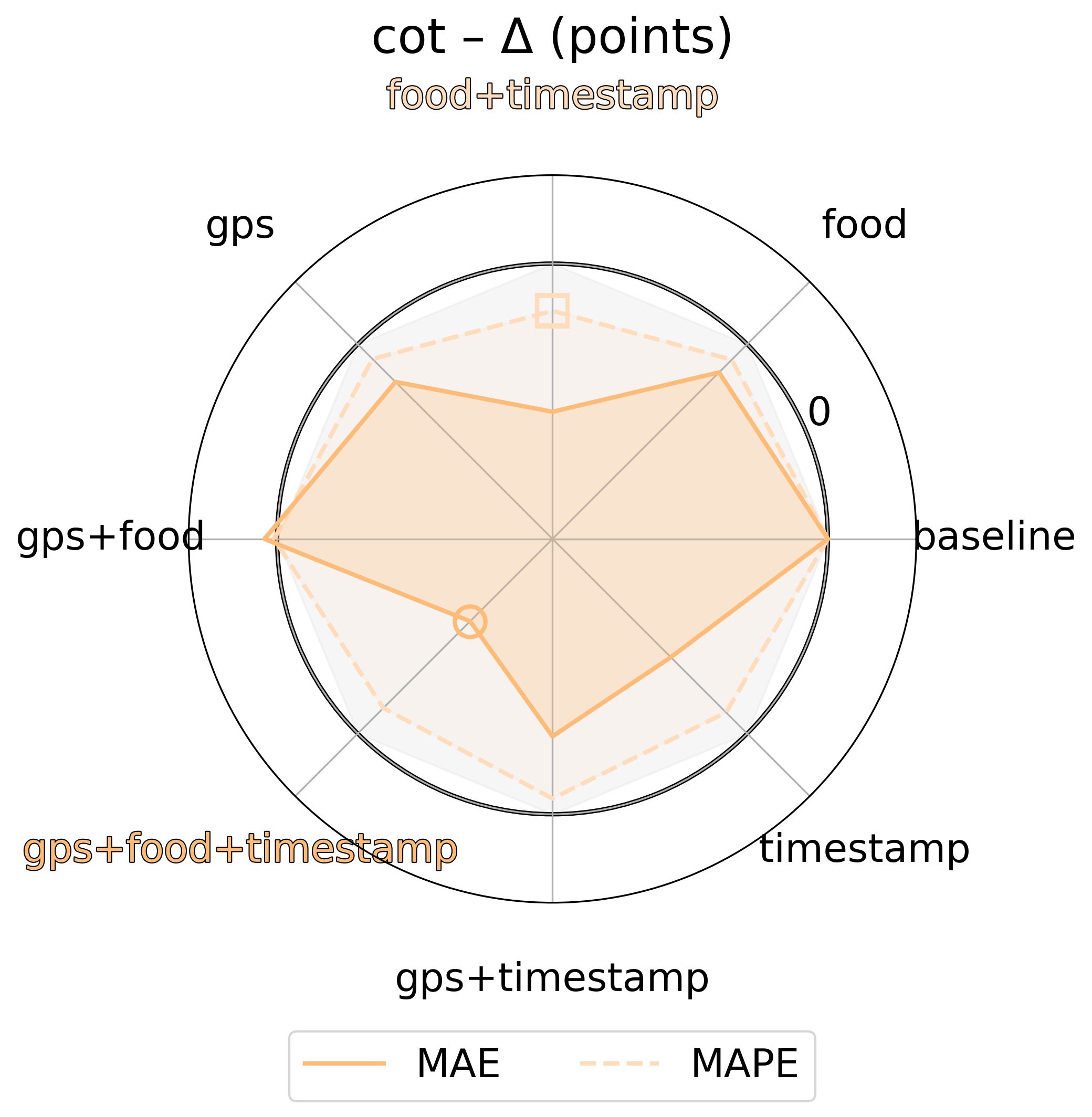}
    \caption{Chain-of-Thought}
    \label{fig:radar_cot}
  \end{subfigure}\hfill
  \begin{subfigure}[t]{0.30\textwidth}
    \centering
    \includegraphics[width=\linewidth]{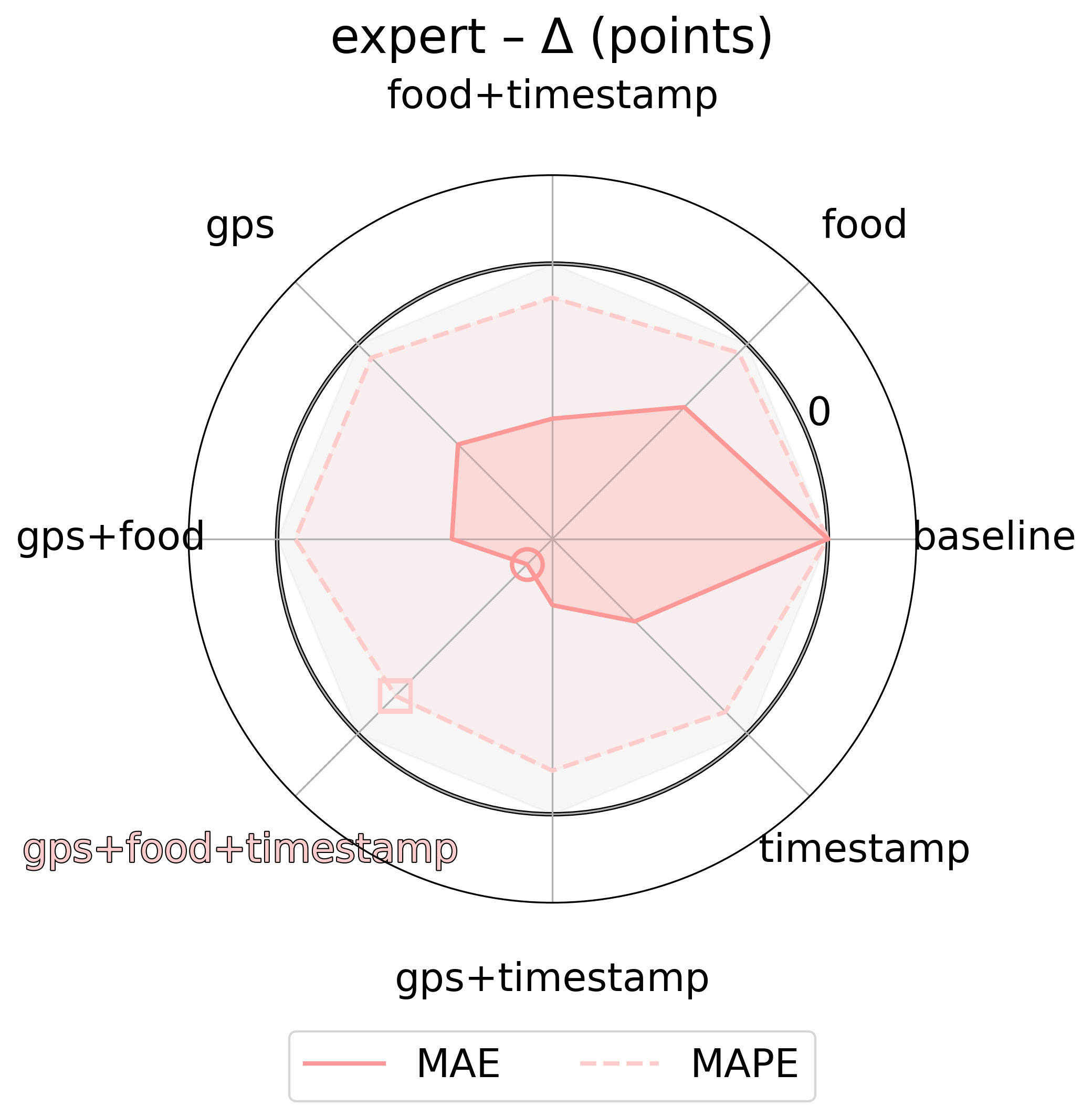}
    \caption{Expert Persona}
    \label{fig:radar_expert}
  \end{subfigure}

  \makebox[\textwidth][c]{%
    \begin{subfigure}[t]{0.30\textwidth}
        \centering
        \includegraphics[width=\linewidth]{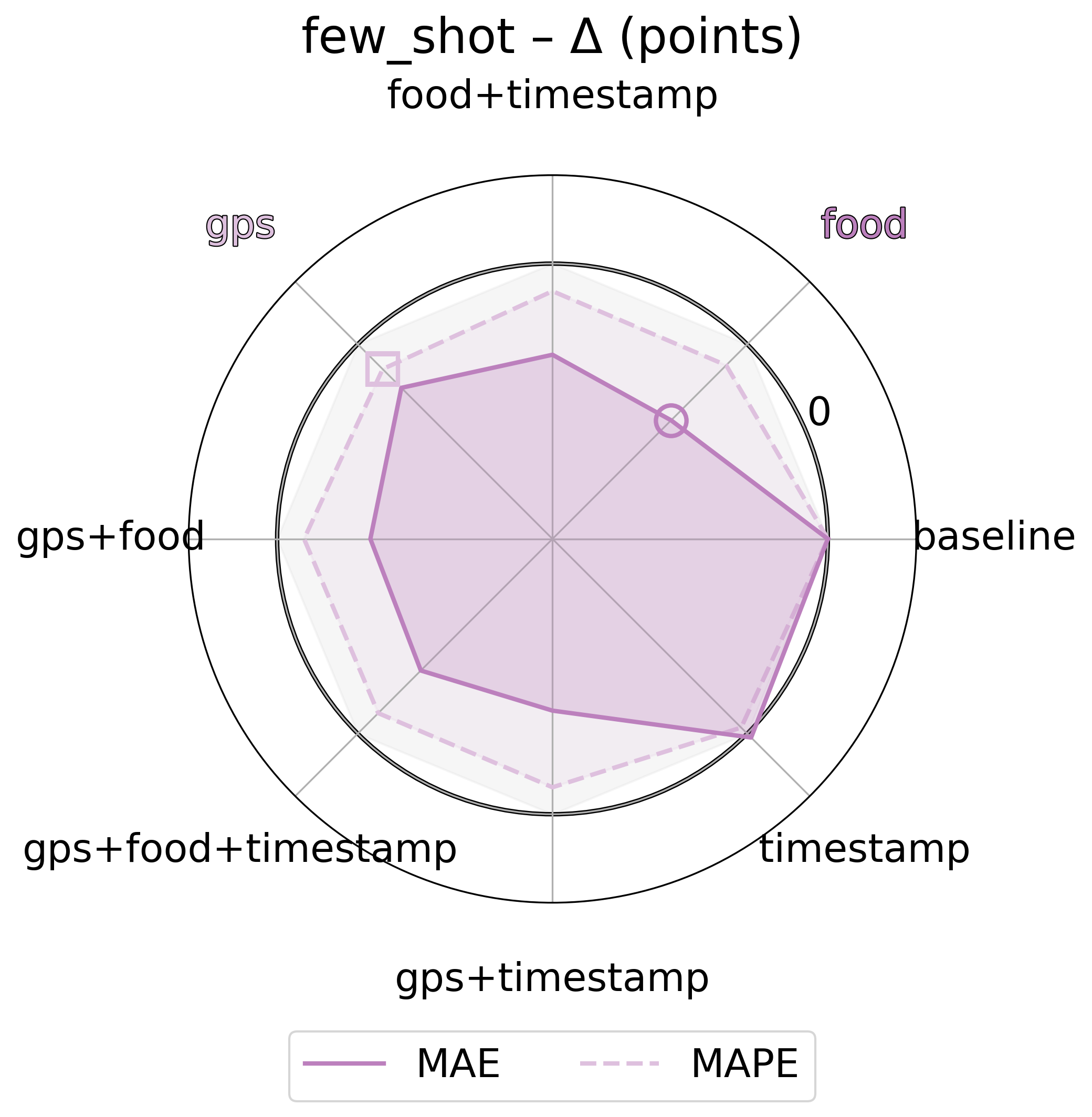}
        \caption{Few-Shot Exemplars}
        \label{fig:radar_few_shot}
      \end{subfigure}\hfill
    \begin{subfigure}[t]{0.30\textwidth}
      \centering
      \includegraphics[width=\linewidth]{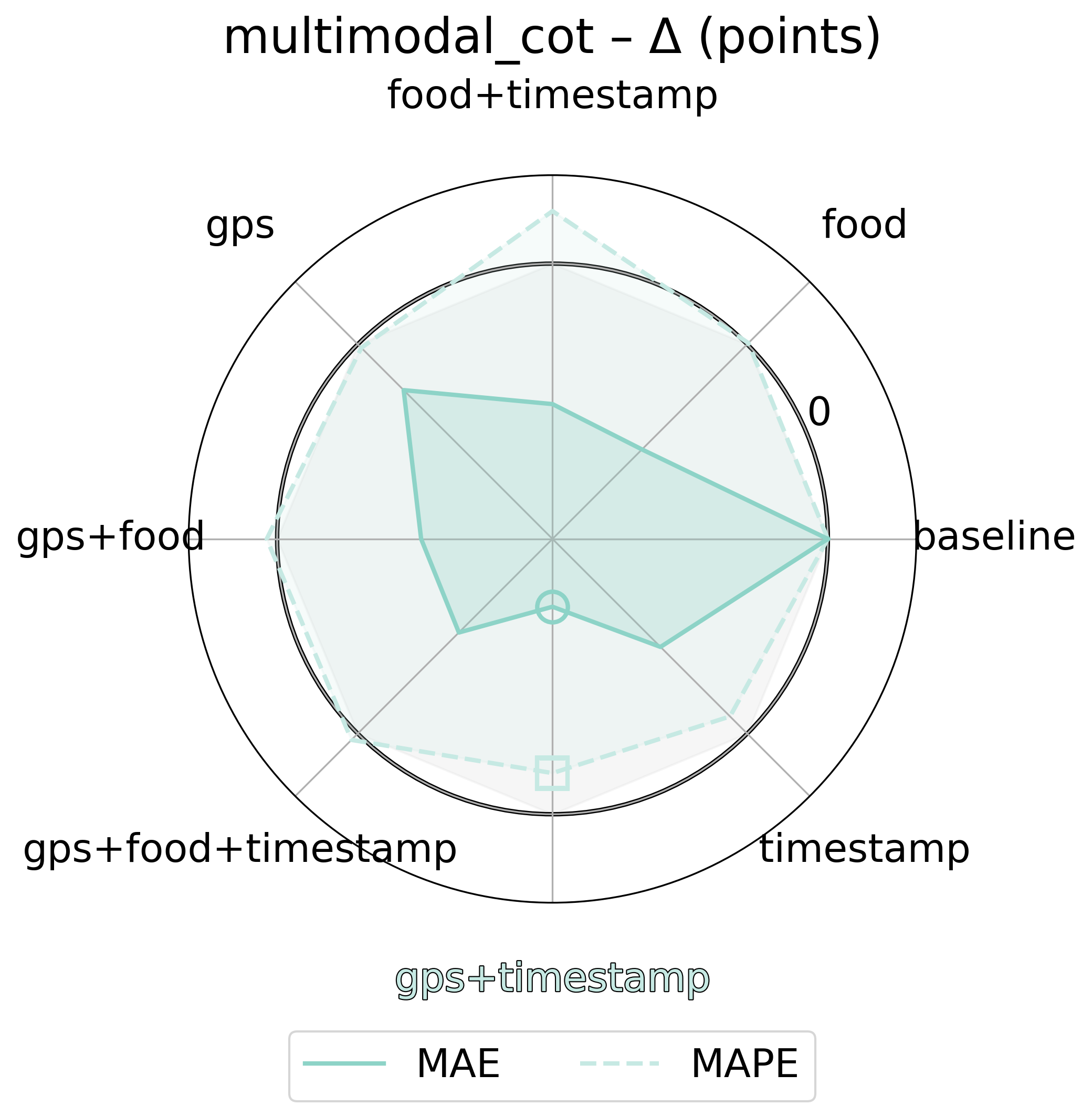}
      \caption{Multimodal Chain-of-Thought}
      \label{fig:radar_multimodal_cot}
    \end{subfigure}\hfill
    \begin{subfigure}[t]{0.30\textwidth}
      \centering
      \includegraphics[width=\linewidth]{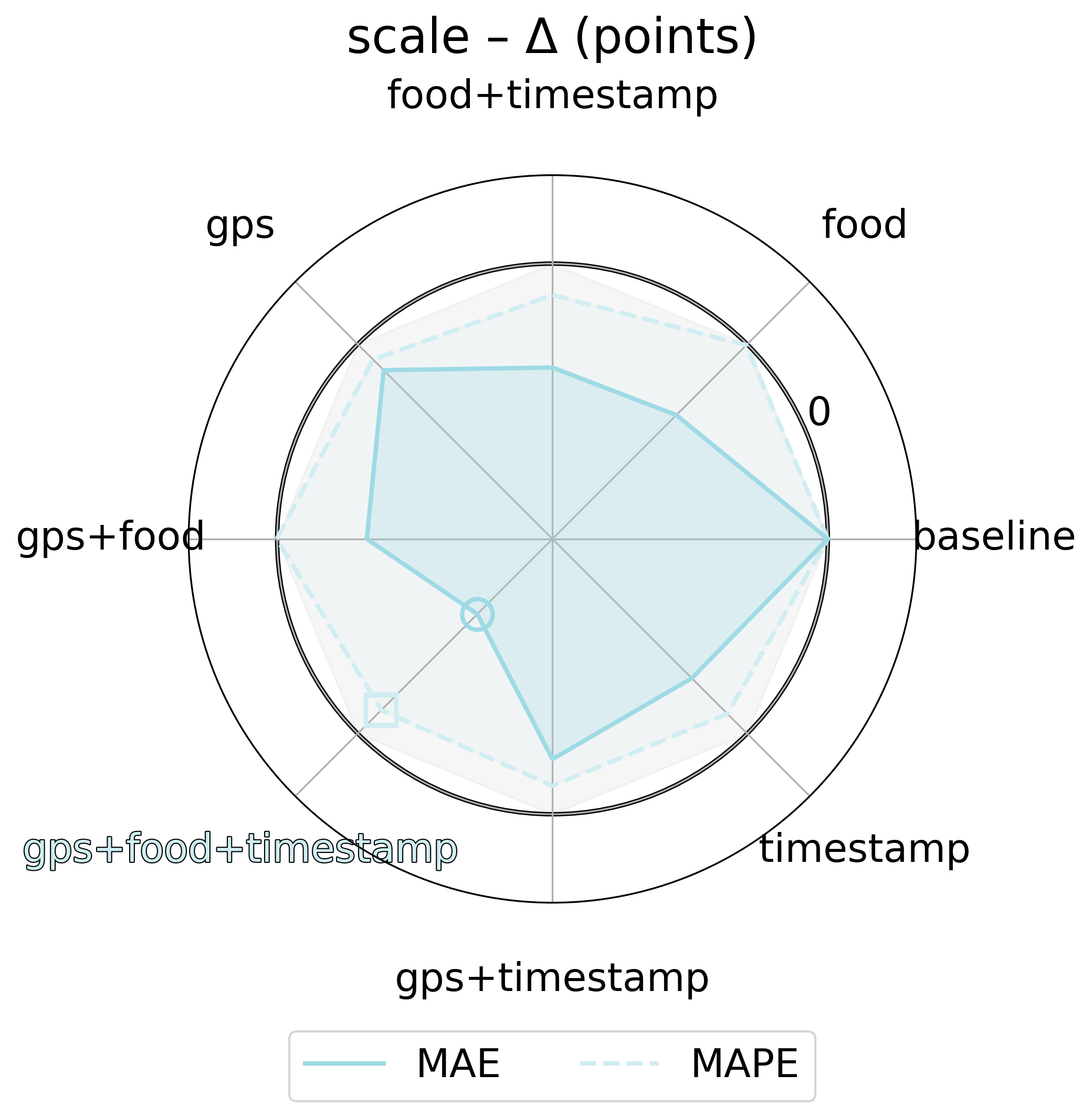}
      \caption{Scale-Hint}
      \label{fig:radar_scale}
    \end{subfigure}%
  }

  \vspace{4pt}
  \caption{\textbf{Averaged Experiment 2 Results According to Reasoning Modifier.} MAE (solid lines) and MAPE (dashed lines) are plotted for various contextual metadata combinations with a given reasoning modifier. Each spoke represents a metadata combination's error; closer proximity to the center signifies a reduction in error relative to the baseline reasoning modifier prompt. Colored markers denote the \textsc{Best-Metadata} configuration for each metric.}
  \label{fig:exp2_radar_suite}
\end{figure*}

\input{tables/experiment2}

\begin{figure*}[htbp]
\centering
\includegraphics[width=\linewidth]{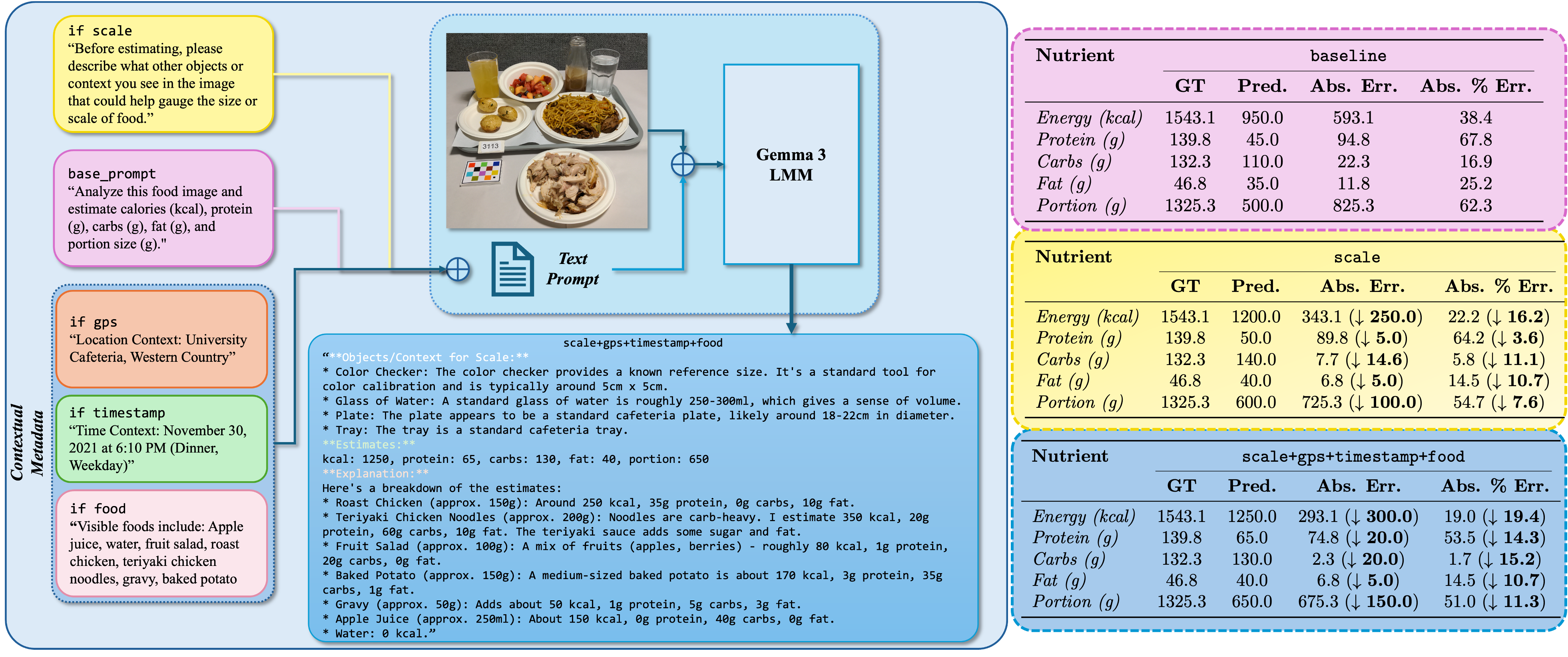}
\caption{Example from broader LMM evaluations for Experiment 2, showing Gemma 3 with the \texttt{scale} reasoning modifier. This figure compares baseline (purple), \texttt{scale} (yellow), and full metadata (blue) prompt strategies. For this Gemma 3 instance, error tables demonstrate full metadata (blue) improves accuracy over baseline/scale for most nutrients, except fat (g).}
\label{fig:prompt-flow}
\end{figure*}

%% file: tables/experiment1.tex
\begin{table*}[htbp]
  \centering
  \caption{\textbf{Impact of Metadata Integration on LMM Nutritional Estimation Error.} This table presents MAE and MAPE changes from a baseline prompt to the \emph{best performing} metadata combination scheme. Downward arrows ($\downarrow$) and bolded values signify an improvement (i.e., reduced error). The best baseline for each metric is highlighted in blue. Cells are colored green for error reduction and red for error increase, with the largest error reduction being bolded for each nutrient measure.}
  \label{tab:rel_change_abridged_metadata}
  \resizebox{\textwidth}{!}{%
  \begin{tabular}{@{}llcccccccccccc@{}}
    \toprule
    \multirow{2}{*}{\textbf{Model}} & \multirow{2}{*}{\textbf{Weight-Type}} & \multirow{2}{*}{\textbf{Scheme}} & \multicolumn{2}{c}{\textbf{Energy (kcal)}} & \multicolumn{2}{c}{\textbf{Protein (g)}} & \multicolumn{2}{c}{\textbf{Carbs (g)}} & \multicolumn{2}{c}{\textbf{Fat (g)}} & \multicolumn{2}{c}{\textbf{Portion (g)}} \\
    & & & \emph{($\downarrow$) MAE (kcal)} & \emph{($\downarrow$) MAPE (\%)} & \emph{($\downarrow$) MAE (g)} & \emph{($\downarrow$) MAPE (\%)} & \emph{MAE (g)} & \emph{($\downarrow$) MAPE (\%)} & \emph{($\downarrow$) MAE (g)} & \emph{($\downarrow$) MAPE (\%)} & \emph{MAE (g)} & \emph{($\downarrow$) MAPE (\%)} \\
    \midrule
    \multirow{3}{*}{Claude-3.7-Sonnet} & \multirow{3}{*}{Closed} &  Baseline   & 181.68 & 23.88 & 15.41 & 36.02 & 25.01 & \base{\textbf{37.22}} & 7.82 & 41.91 & 329.89 & 42.81 \\
     &  & \textbf{w/ (\texttt{gps+timestamp+food})} & \downv{27.91} & \downv{2.56} & \downv{4.99} & \downv{9.05} & \downv{3.82} & \downv{2.32} & \downv{1.74} & \downv{14.45} & \downv{55.52} & \downv{7.36} \\
    \midrule
    \multirow{3}{*}{GPT-4.1} & \multirow{3}{*}{Closed} &  Baseline   & 170.02 & \base{\textbf{23.29}} & 14.33 & 33.00 & \base{\textbf{22.9}}0 & 39.34 & 7.58 & \base{\textbf{32.21}} & 315.83 & 40.62 \\
     &  & \textbf{w/ (\texttt{timestamp+food})} & \downv{16.83} & \downv{2.03} & \downv{2.09} & \downv{6.06} & \downv{2.54} & \downv{4.38} & \downv{0.22} & \downv{1.16} & \downv{19.73} & \downv{2.53} \\
    \midrule 
    \multirow{3}{*}{GPT-4o} & \multirow{3}{*}{Closed} &  Baseline   & \base{\textbf{165.77}} & 24.48 & 12.67 & \base{\textbf{31.15}} & 23.37 & 42.08 & \base{\textbf{7.53}} & 37.67 & 259.96 & 33.37 \\
     &  & \textbf{w/ (\texttt{gps+food})} & \downv{11.61} & \downv{1.53} & \downv{2.63} & \downv{5.85} & \upv{1.00} & \downv{1.04} & \downv{0.21} & \downv{3.82} & \downv{49.46} & \downv{6.79} \\
    \midrule
    \multirow{3}{*}{Gemini-2.5-pro} & \multirow{3}{*}{Closed} &  Baseline   & 211.04 & 34.11 & \base{\textbf{12.45}} & 39.67 & 28.47 & 50.15 & 10.84 & 53.96 & \base{\textbf{188.12}} & \base{\textbf{25.77}} \\
     &  & \textbf{w/ (\texttt{gps+timestamp+food})} & \downv{41.37} & \downv{6.28} & \downv{2.31} & \downv{9.19} & \downv{1.80} & \downv{3.51} & \downv{2.20} & \downv{12.37} & \downv{11.99} & \downv{1.01} \\
    \midrule
    \multirow{3}{*}{Gemma3} & \multirow{3}{*}{Open} &  Baseline   & 187.54 & 27.67 & 13.72 & 36.49 & 24.51 & 51.15 & 9.47 & 61.96 & 332.00 & 41.61 \\
     &  & \textbf{w/ (\texttt{gps})} & \downv{10.35}  & \upv{2.49} & \downv{2.22} & \upv{1.39} & \downv{2.30}  & \downv{1.94} & \upv{1.12}  & \upv{7.00}  & \downv{67.44}  & \downv{9.39}  \\
    \midrule
    \multirow{3}{*}{Janus-Pro} & \multirow{3}{*}{Open} &  Baseline   & 477.79 & 86.29 & 20.68 & 50.75 & 53.31 & 67.23 & 14.58 & 58.70 & 656.88 & 90.02 \\
     &  & \textbf{w/ (\texttt{gps+timestamp})} & \downv{\textbf{246.46}}  & \downv{\textbf{52.20}}  & \upv{0.68} & \downv{0.64} & \downv{2.62}  & \downv{4.28}  & \downv{1.70}  & \downv{9.09} & \downv{70.87}  & \downv{9.35} \\
    \midrule
    \multirow{3}{*}{LLaMA-3.2-Vision-Instruct} & \multirow{3}{*}{Open} &  Baseline   & 496.47 & 66.22 & 38.46 & 117.71 & 45.61 & 66.79 & 45.74 & 200.99 & 483.03 & 62.31 \\
     &  & \textbf{w/ (\texttt{gps+timestamp})} & \downv{193.25}  & \downv{28.90}  & \downv{\textbf{7.61}}  & \downv{\textbf{20.12}}  & \downv{\textbf{16.82}}  & \downv{\textbf{13.13}}  & \downv{\textbf{14.45}}  & \downv{\textbf{52.44}}  & \downv{\textbf{123.67}}  & \downv{\textbf{16.23}}  \\
    \midrule
    \multirow{3}{*}{Qwen2.5-VL} & \multirow{3}{*}{Open} &  Baseline   & 276.04 & 33.90 & 15.70 & 37.16 & 38.35 & 50.02 & 10.86 & 43.18 & 497.17 & 66.66 \\
     &  & \textbf{w/ (\texttt{timestamp+food})} & \downv{59.58}  & \downv{6.20}  & \downv{2.68}  & \downv{1.57}  & \downv{6.10}  & \upv{0.30}  & \downv{2.37}  & \downv{6.52}  & \downv{23.26}  & \downv{3.39}  \\
    \midrule
    \bottomrule
  \end{tabular}}
\end{table*}

%% file: tables/experiment2.tex
\begin{table*}[htbp]
  \centering
  \caption{\textbf{Impact of Metadata Integration on Reasoning Modifier Nutritional Estimation Error.} This table presents MAE and MAPE changes from a baseline reasoning modifier to the \emph{best performing} metadata combination (averaged across all models). Downward arrows ($\downarrow$) and accompanying values signify an improvement (i.e., reduced error). The best baseline for each metric is highlighted in blue. Cells are colored green for error reduction and red for error increase, with the largest error reduction being bolded for each nutrient measure.}
  \label{tab:avg_across_models_rel_change}
  \resizebox{\textwidth}{!}{%
  \begin{tabular}{@{}llcccccccccc@{}}
    \toprule
    \multirow{2}{*}{\textbf{Reasoning Modifier}} & \multirow{2}{*}{\textbf{Scheme}} & \multicolumn{2}{c}{\textbf{Energy (kcal)}} & \multicolumn{2}{c}{\textbf{Protein (g)}} & \multicolumn{2}{c}{\textbf{Carbs (g)}} & \multicolumn{2}{c}{\textbf{Fat (g)}} & \multicolumn{2}{c}{\textbf{Portion (g)}} \\
    & & \emph{($\downarrow$) MAE (kcal)} & \emph{($\downarrow$) MAPE (\%)} & \emph{($\downarrow$) MAE (g)} & \emph{($\downarrow$) MAPE (\%)} & \emph{($\downarrow$) MAE (g)} & \emph{($\downarrow$) MAPE (\%)} & \emph{($\downarrow$) MAE (g)} & \emph{($\downarrow$) MAPE (\%)} & \emph{($\downarrow$) MAE (g)} & \emph{($\downarrow$) MAPE (\%)} \\
    \midrule
    \multirow{3}{*}{Chain-of-Thought} &  Baseline   & 255.00 & 37.07 & 16.26 & 41.08 & \base{\textbf{30.39}} & \base{\textbf{46.23}} & 10.15 & 46.67 & \base{\textbf{350.20}} & \base{\textbf{46.18}} \\
     &  \textbf{w/ (\texttt{food+timestamp})} & \downv{51.08} & \downv{7.09} & \downv{2.26} & \downv{\textbf{5.14}} & \downv{\textbf{3.68}} & \downv{3.62} & \downv{1.31} & \downv{\textbf{7.77}} & \downv{20.92} & \downv{1.61} \\
    \midrule
    \multirow{3}{*}{Expert Persona} &  Baseline   & 284.63 & 40.56 & 16.28 & 39.77 & 32.51 & 48.69 & 10.59 & 48.56 & 386.38 & 51.47 \\
     &  \textbf{w/ (\texttt{gps+food+timestamp})} & \downv{\textbf{75.39}} & \downv{\textbf{10.38}} & \downv{2.29} & \downv{4.09} & \downv{3.44} & \downv{0.74} & \downv{\textbf{1.67}} & \downv{7.31} & \downv{\textbf{45.28}} & \downv{\textbf{6.00}} \\
    \midrule
    \multirow{3}{*}{Few Shot Exemplars} &  Baseline   & 240.44 & \base{\textbf{33.05}} & 15.78 & \base{\textbf{38.27}} & 31.86 & 47.28 & 10.05 & 46.37 & 381.45 & 50.96 \\
     &  \textbf{w/ (\texttt{gps})} & \downv{21.31} & \downv{4.26} & \downv{0.72} & \downv{2.06} & \downv{2.53} & \downv{\textbf{4.60}} & \downv{0.68} & \downv{5.74} & \downv{7.72} & \downv{2.15} \\
    \midrule
    \multirow{3}{*}{Multimodal CoT} &  Baseline   & 274.43 & 40.54 & \base{\textbf{15.62}} & 40.09 & 32.44 & 52.16 & 11.54 & 51.14 & 360.51 & 47.49 \\
     &  \textbf{w/ (\texttt{gps+timestamp})} & \downv{64.46} & \downv{7.72} & \upv{0.25} & \downv{0.64} & \downv{1.54} & \downv{3.63} & \downv{1.11} & \downv{5.60} & \downv{44.14} & \downv{4.41} \\
    \midrule
    \multirow{3}{*}{Scale-Hint} &  Baseline   & \base{\textbf{238.79}} & 34.44 & 15.68 & 38.46 & 30.74 & 47.57 & \base{\textbf{9.91}} & \base{\textbf{44.81}} & 376.27 & 50.54 \\
     &  \textbf{w/ (\texttt{gps+food+timestamp})} & \downv{40.49} & \downv{5.67} & \downv{\textbf{2.42}} & \downv{2.96} & \downv{2.60} & \upv{0.48} & \downv{1.22} & \downv{4.49} & \downv{43.64} & \downv{5.09} \\
    \midrule
    \bottomrule
  \end{tabular}}
\end{table*}

%% file: 05_discussions.tex
\section{Discussion}
\label{sec:discussion}

The present study demonstrates that even minimal contextual metadata, specifically location and meal-time information, improves nutritional estimates produced by LMMs. The improvement persists across eight different models, multiple reasoning templates, and across all nutrient metrics. Because mobile devices routinely collect GPS coordinates and timestamps, this finding points to a readily deployable performance intervention for applications such as dietary self-monitoring.

Differences between open-weight and closed-weight models offer additional insight into how contextual knowledge is stored and accessed. Open models exhibited the largest absolute gains once metadata was injected, sometimes overtaking closed-weight models that initially led from baseline estimates. A plausible explanation is that proprietary models have already internalized parts of the same contextual signal through Reinforcement Learning from Human Feedback (RLHF) or extensive curation, leaving less margin for improvement from user-provided cues. Open-weight models, by contrast, carry weaker default priors yet remain highly receptive to these provided cues. This asymmetry suggests a trade-off: closed-weight models may deliver strong out-of-the-box performance, but open models can close the gap (and even surpass it) when supplied with lightweight retrieval of information at inference.

One of the more interesting results is that integrating location and meal-time metadata (\texttt{gps} and \texttt {timestamp}) led to an increase in performance among most models and reasoning modifiers. Incorporating these signals incurs negligible latency and could further be integrated into existing mobile nutrition analysis pipelines (assuming that these metadata dimensions can be provided). Crucially, the gains require no additional fine-tuning, offering a low-cost path to more accurate dietary assessment.

%% file: 06_conclusions_and_future_work.tex
\section{Conclusions and Future Work}
\label{sec:conclusion}

This study introduces ACETADA, the first publicly available food-image dataset that pairs dietitian-verified nutrition labels with GPS coordinates, timestamps, and ground-truth food lists. Leveraging this resource, we conduct a cross-model evaluation encompassing eight LMMs, three metadata facets, and five reasoning modifiers. This study also demonstrates that incorporating constructed context into prompts can reduce MAE and MAPE across nutrition metrics, even when coupled with reasoning modifiers. This metadata injection rarely harms accuracy and is particularly effective for open-weight checkpoints, highlighting its value in privacy-constrained or on-premise deployments. 

Several directions for future work follow naturally. The metadata palette could be expanded to include social or behavioral cues, such as dining companions or habitual eating patterns, to better contextualize portion predictions. Architecturally, future models might benefit from integrating metadata into hidden representations (with such methods as cross-attention, gated adapters, or prompt tuning) rather than relying solely on text concatenation. Finally, assessing reliability under degraded conditions (e.g., blurred images or missing metadata) would help evaluate readiness for clinical or real-world deployment.

%% file: 07_supplemental_material.tex
\section{Supplemental Material}
\label{sec:supplemental}

This supplementary document furnishes additional figures and methodological details that extend and contextualize the results reported in the main paper.

\subsection{Additional ACETADA Visualizations}

\begin{figure}[htbp]
    \centering
    \includegraphics[width=\linewidth]{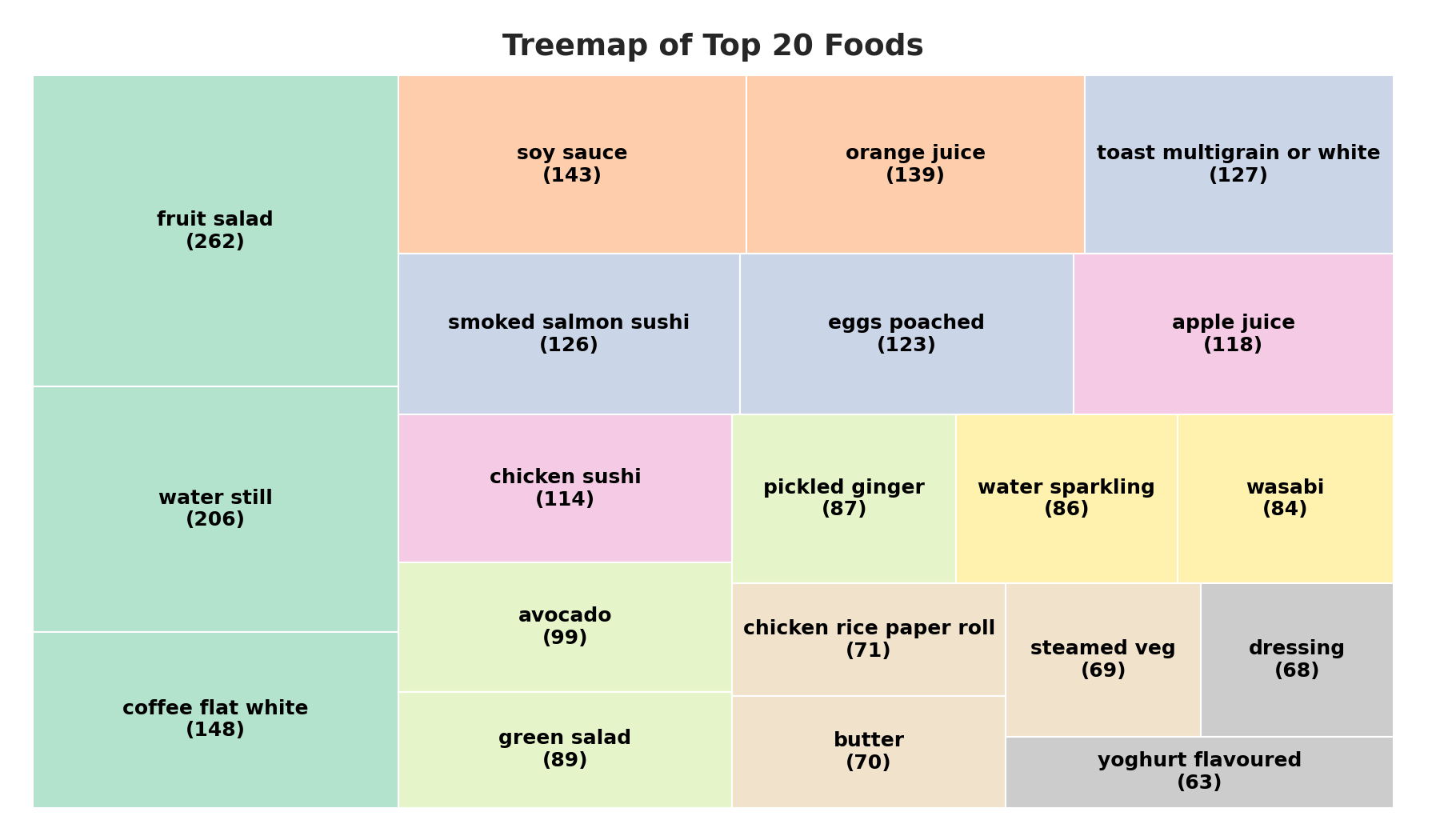}
    \caption{Treemap of the twenty most common food items in ACETADA, as identified by dietitian-verified labels. Block area is proportional to the food item frequency.}
    \label{fig:food_treemap}
\end{figure}

\begin{figure}[htbp]
    \centering
    \includegraphics[width=\linewidth]{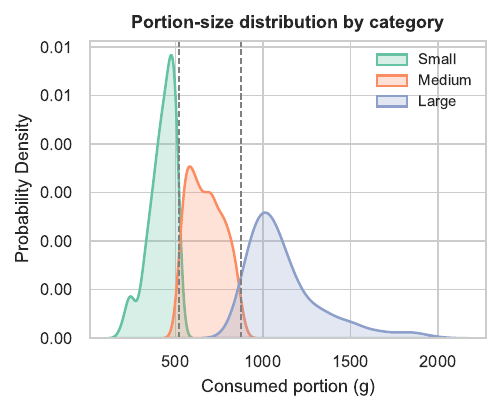}
    \caption{Kernel-density estimate of consumed portion sizes in ACETADA, stratified by ``small", ``medium", and ``large" consumption categories.}
    \label{fig:portion_size_kde_by_cat}
\end{figure}

To provide a more comprehensive description of the ACETADA dataset, we include two visualizations that are not included in the main text of our study. Figure~\ref{fig:food_treemap} presents a treemap of the twenty most frequent food items that appear from the dietitian-verified annotations. Figure~\ref{fig:portion_size_kde_by_cat} shows a kernel-density estimate of consumed portion sizes derived from paired before/after meal images. Portions are stratified into ``small" ($\leq 500\,$g), ``medium" ($500\,$g-$750\,$g), and ``large" ($>750\,$g) categories. Together, these plots further highlight the diversity of represented foods and the broad range of portion sizes in the ACETADA dataset.

\subsection{Experiment 1: Model-Averaged Metadata Effects}

\begin{figure*}[htbp] 
  \centering

  \begin{subfigure}[t]{0.24\linewidth}
    \centering
    \includegraphics[width=\linewidth]{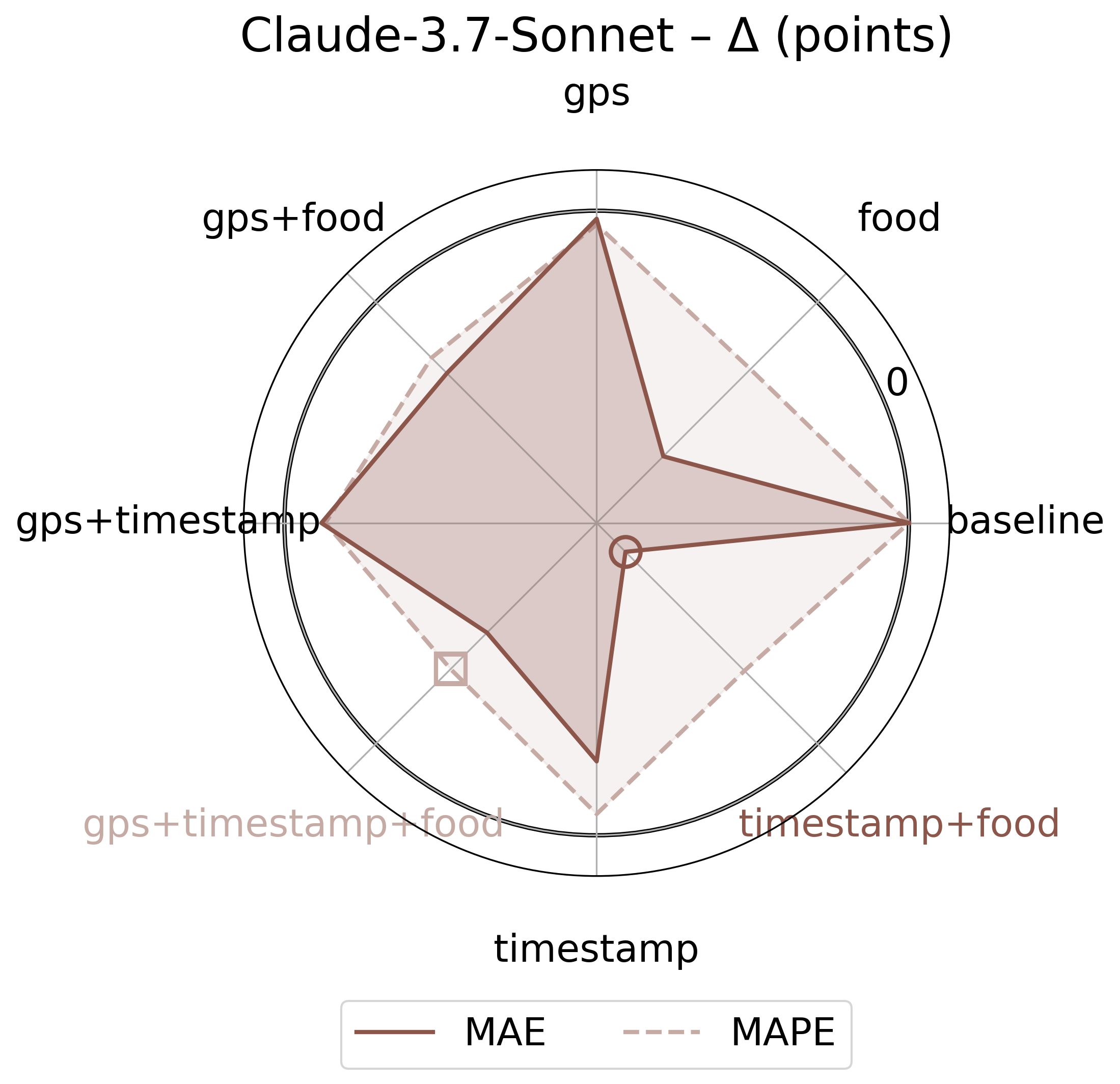}
    \caption{Claude-3.7-Sonnet}
    \label{fig:img1}
  \end{subfigure}
  \begin{subfigure}[t]{0.24\linewidth}
    \centering
    \includegraphics[width=\linewidth]{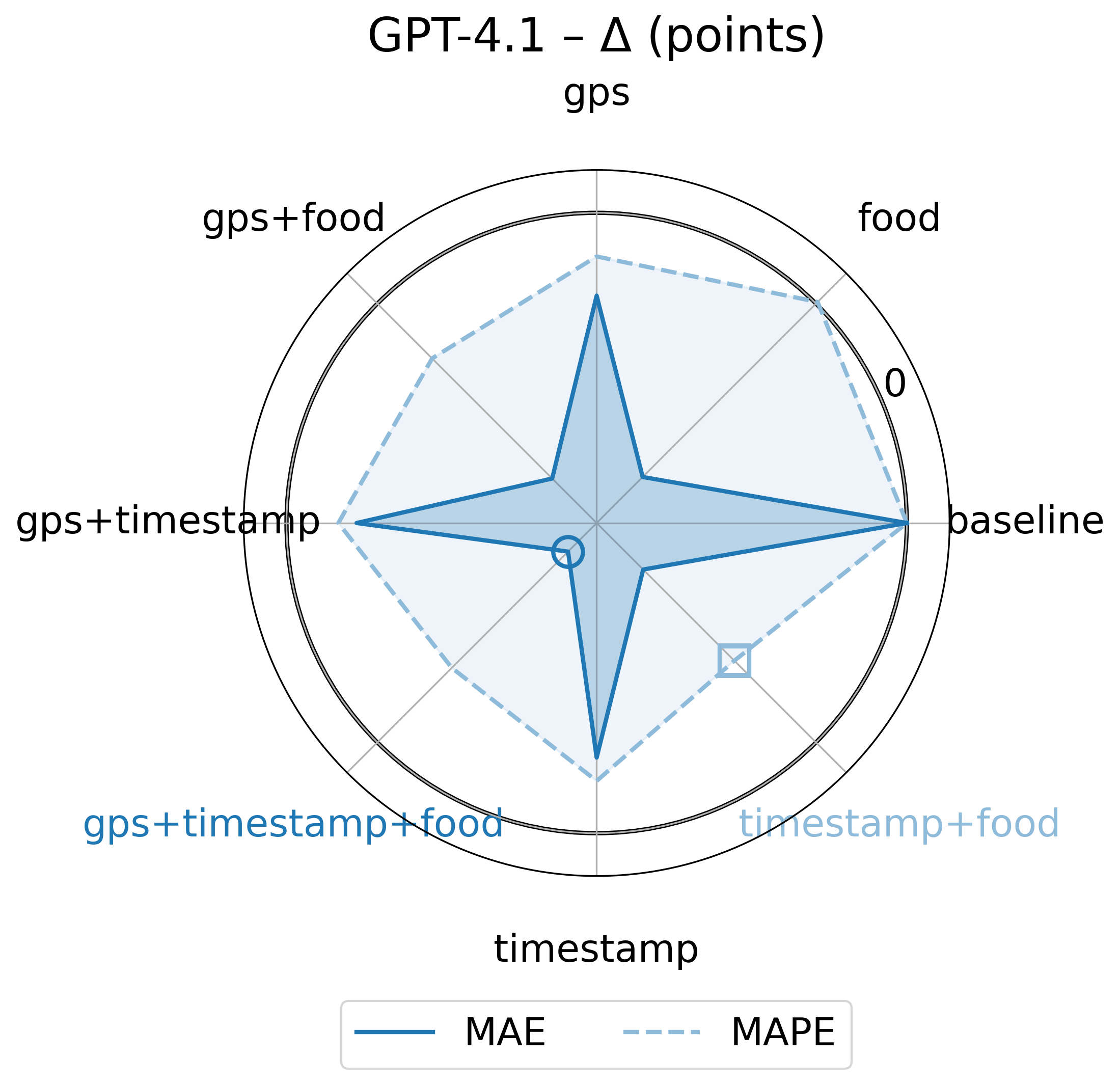}
    \caption{GPT-4.1}
    \label{fig:img2}
  \end{subfigure}
  \begin{subfigure}[t]{0.24\linewidth}
    \centering
    \includegraphics[width=\linewidth]{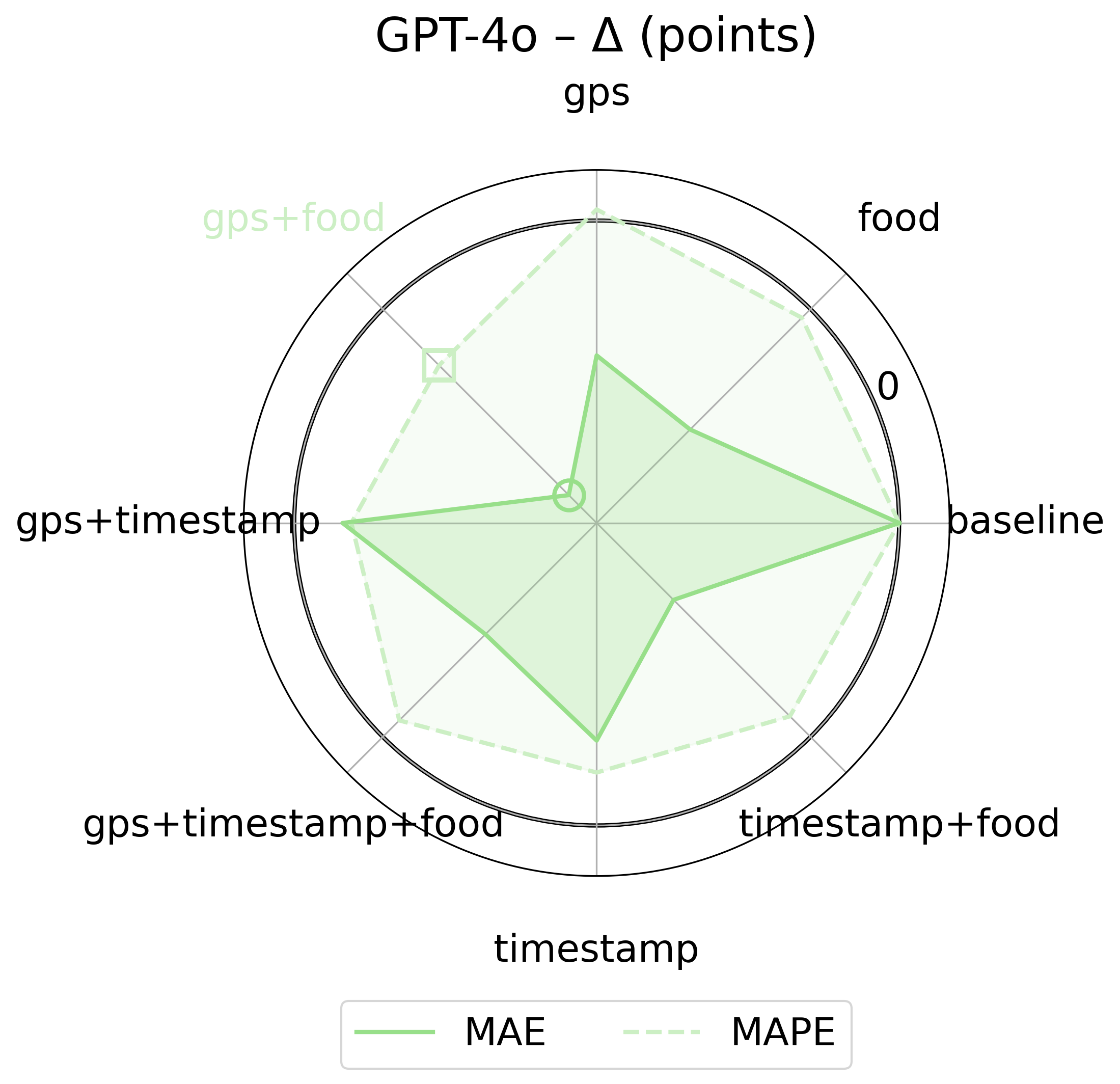}
    \caption{GPT-4o}
    \label{fig:img3}
  \end{subfigure}
  \begin{subfigure}[t]{0.24\linewidth}
    \centering
    \includegraphics[width=\linewidth]{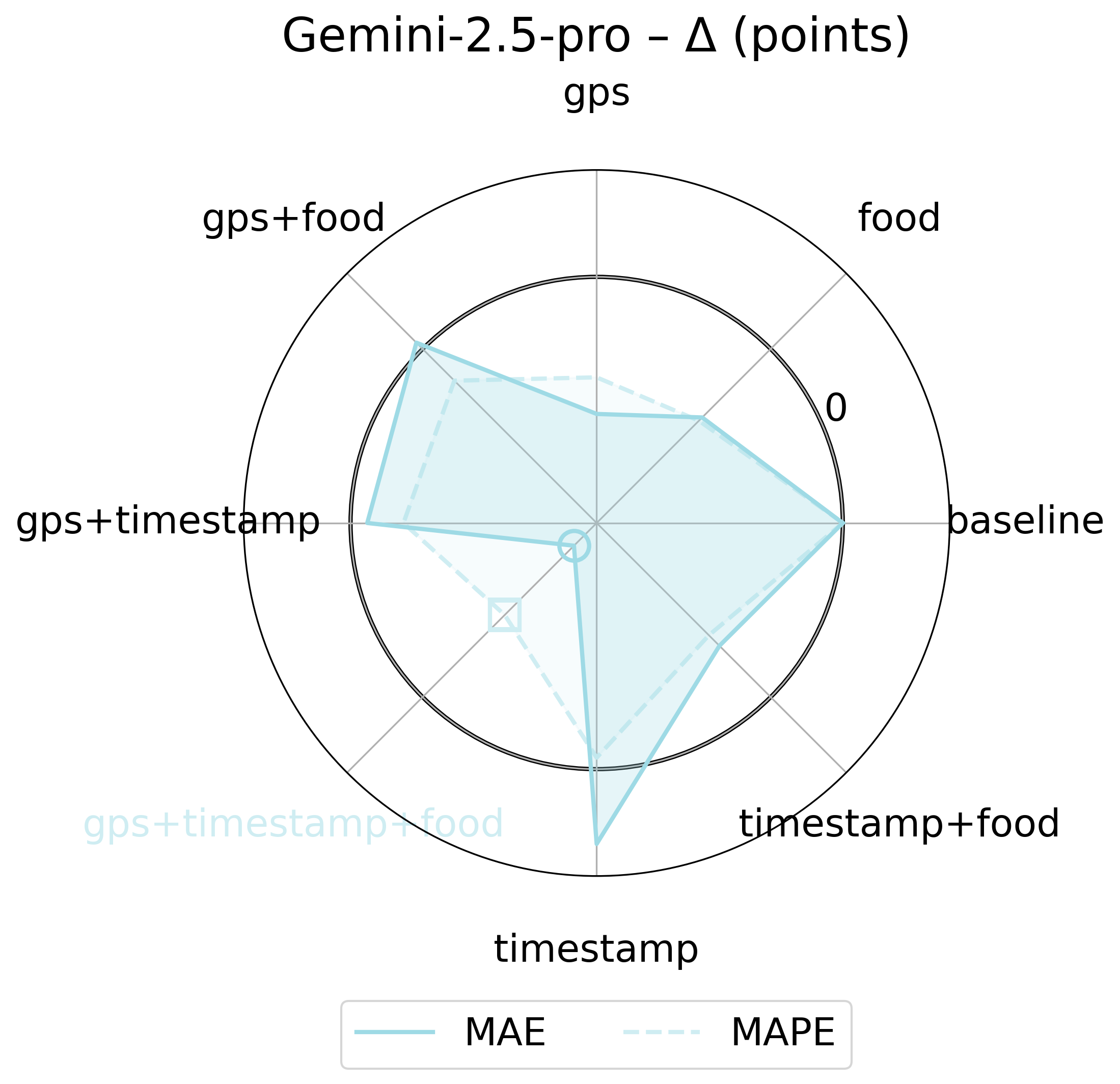}
    \caption{Gemini-2.5-Pro}
    \label{fig:img4}
  \end{subfigure}

  \begin{subfigure}[t]{0.24\linewidth}
    \centering
    \includegraphics[width=\linewidth]{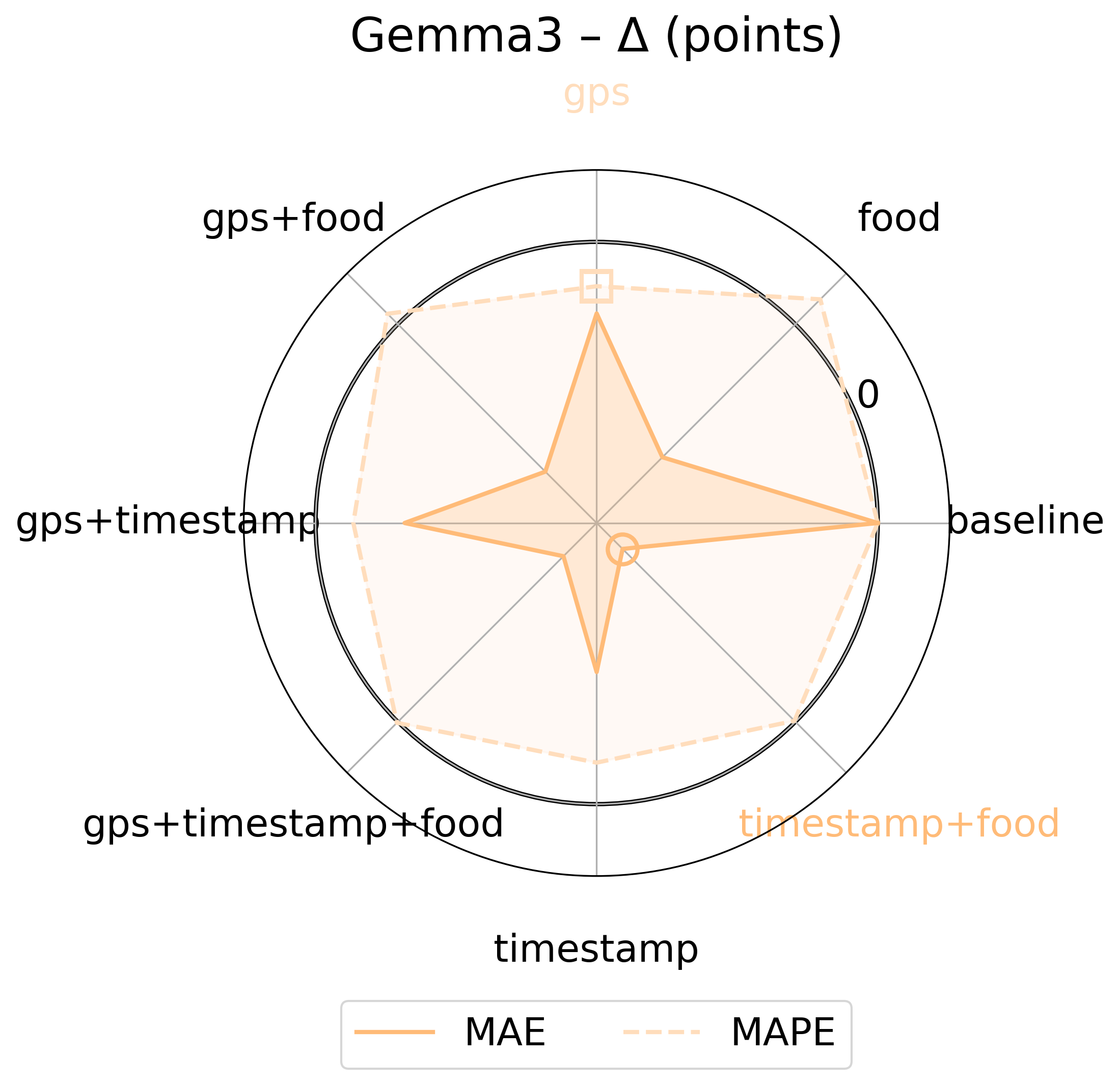}
    \caption{Gemma3}
    \label{fig:img5}
  \end{subfigure}
  \begin{subfigure}[t]{0.24\linewidth}
    \centering
    \includegraphics[width=\linewidth]{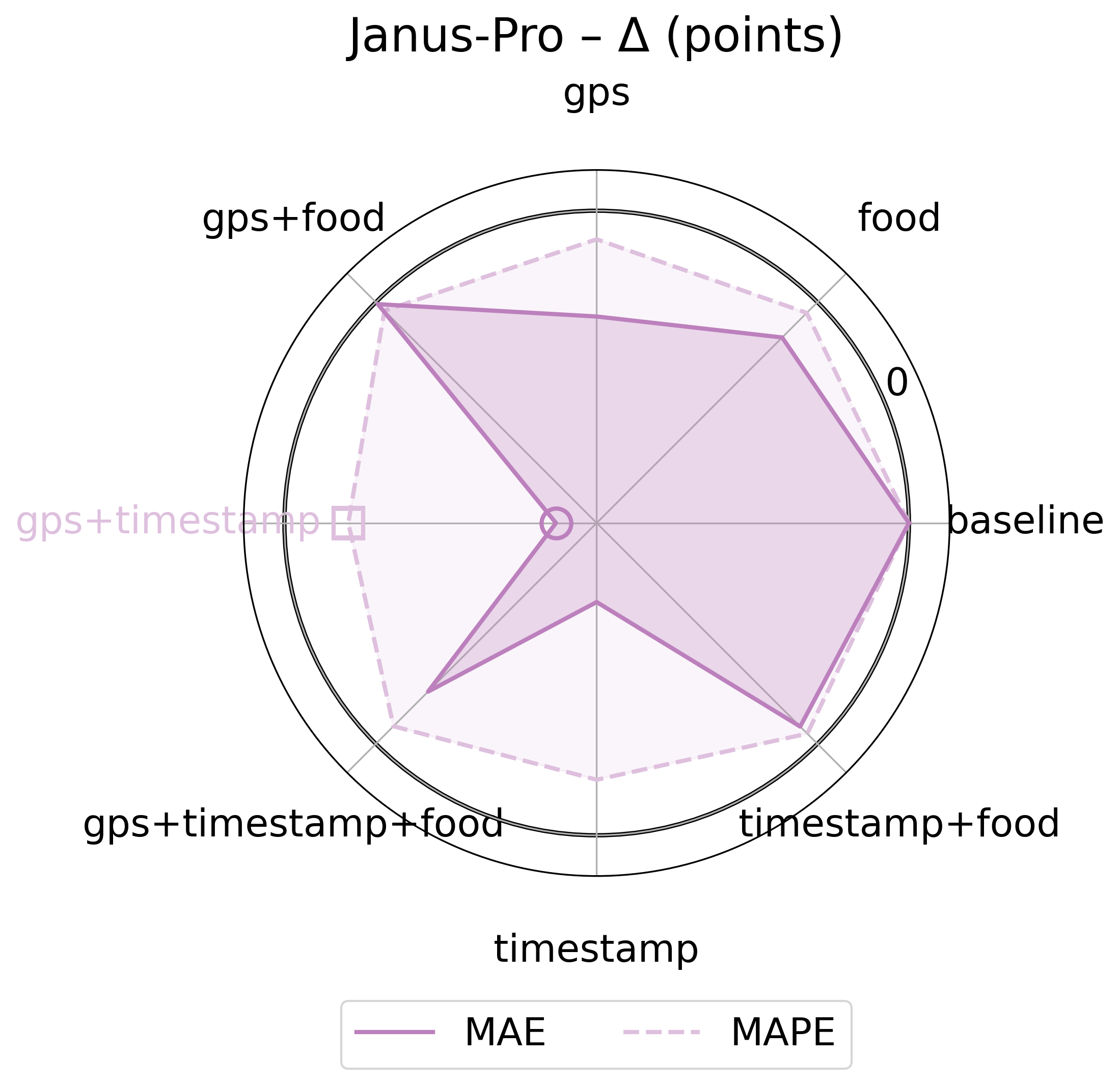}
    \caption{Janus-Pro}
    \label{fig:img6}
  \end{subfigure}
  \begin{subfigure}[t]{0.24\linewidth}
    \centering
    \includegraphics[width=\linewidth]{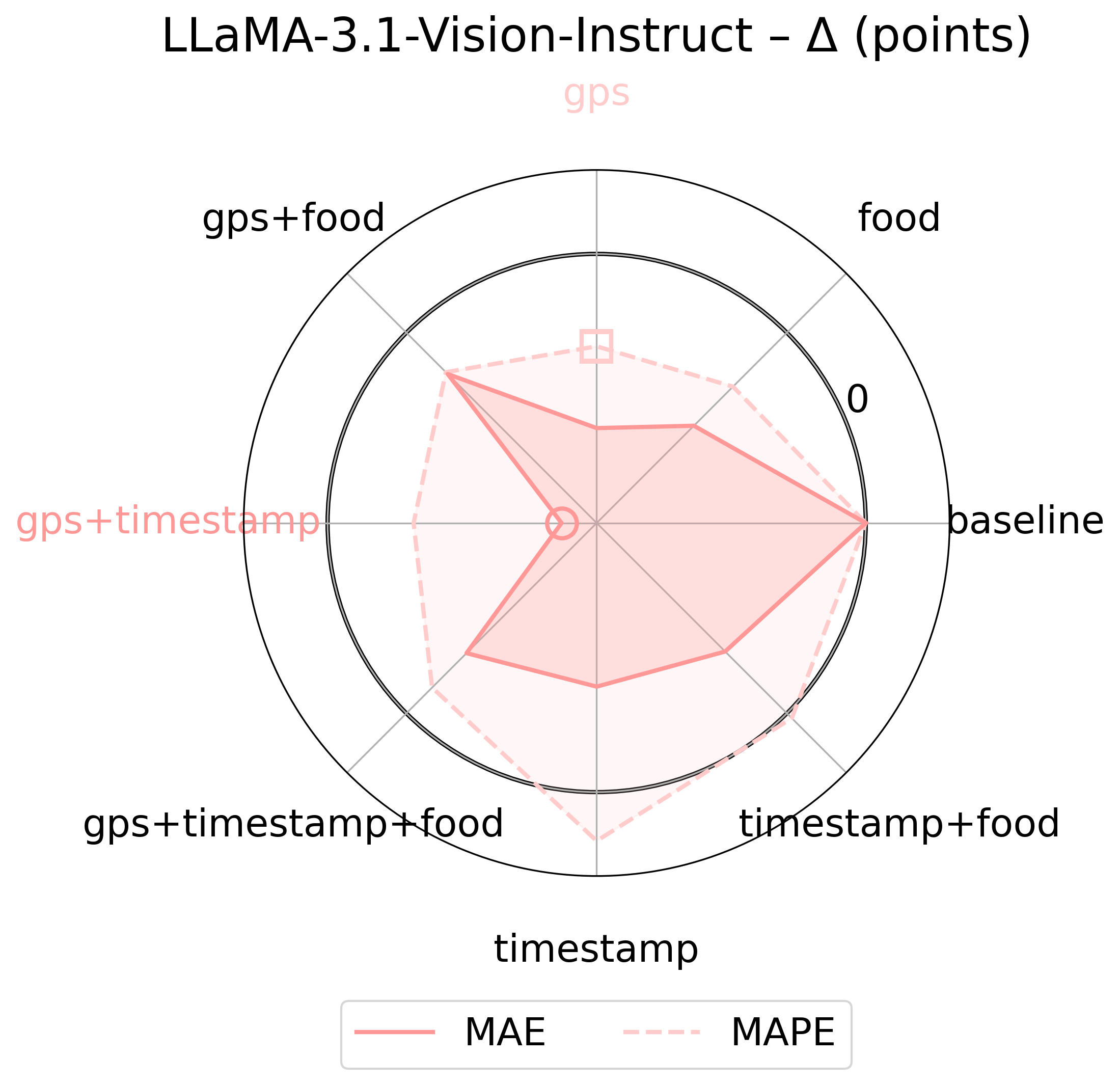}
    \caption{LLaMA-3.2-Vision-Instruct}
    \label{fig:img7}
  \end{subfigure}
  \begin{subfigure}[t]{0.24\linewidth}
    \centering
    \includegraphics[width=\linewidth]{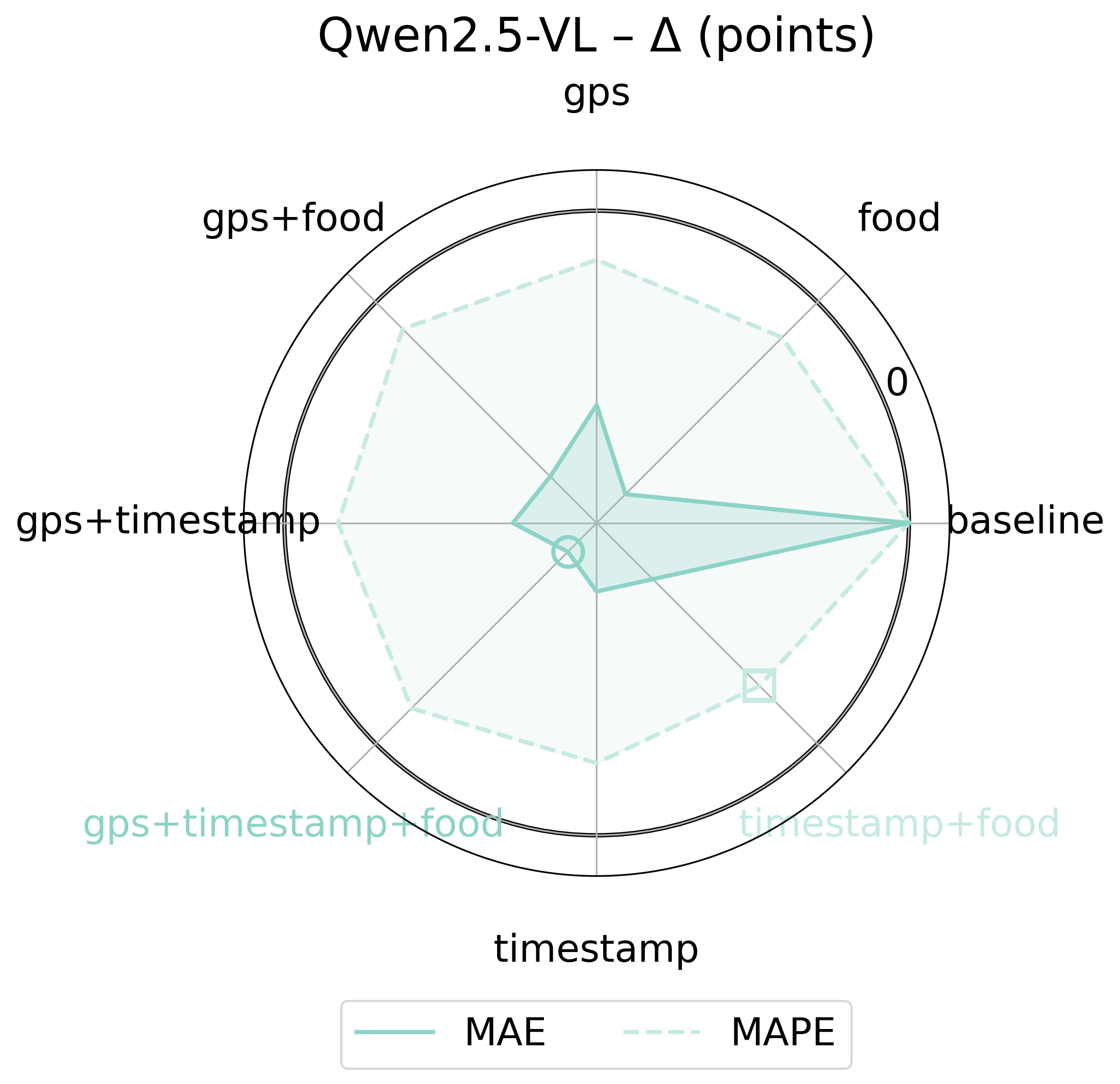}
    \caption{Qwen2.5-VL}
    \label{fig:img8}
  \end{subfigure}

  \caption{\textbf{Averaged Experiment 1 Results For Each Model.} MAE (solid lines) and MAPE (dashed lines) are plotted for various contextual metadata combinations for a given model. Each spoke represents an error in a metadata combination; the closer proximity to the center signifies a reduction in error relative to the baseline prompt. Colored markers denote the \textsc{Best-Metadata} configuration for each metric.}
  \label{fig:exp_1_all_models}
\end{figure*}

Figure~\ref{fig:exp_1_all_models} illustrates the impact of metadata on MAE and MAPE across all models and metadata combinations. This figure is presented similarly to the radar plots in Figures~\ref{fig:radar_all_models}-\ref{fig:radar_closed_models}. With minimal exceptions, incorporating metadata consistently reduces both MAE and MAPE across all models.

\subsection{Experiment 2: Modifier and Model-Averaged Metadata Effects}

\begin{figure*}[htbp] 
  \centering

  \begin{subfigure}[t]{0.195\linewidth}
    \centering
    \includegraphics[width=\linewidth]{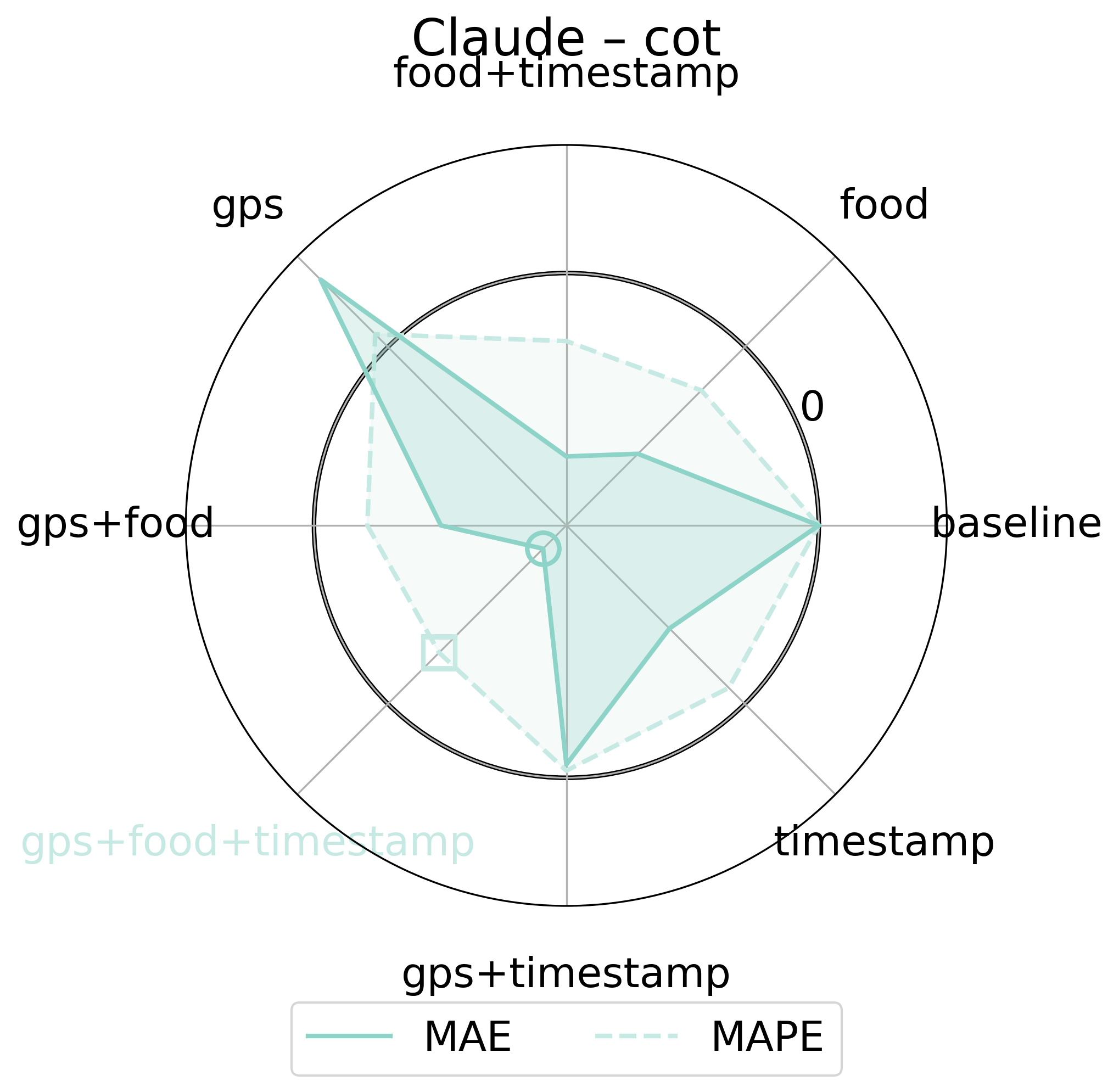}
    \caption{Claude Chain-of-Thought}
    \label{fig:img1}
  \end{subfigure}
  \begin{subfigure}[t]{0.195\linewidth}
    \centering
    \includegraphics[width=\linewidth]{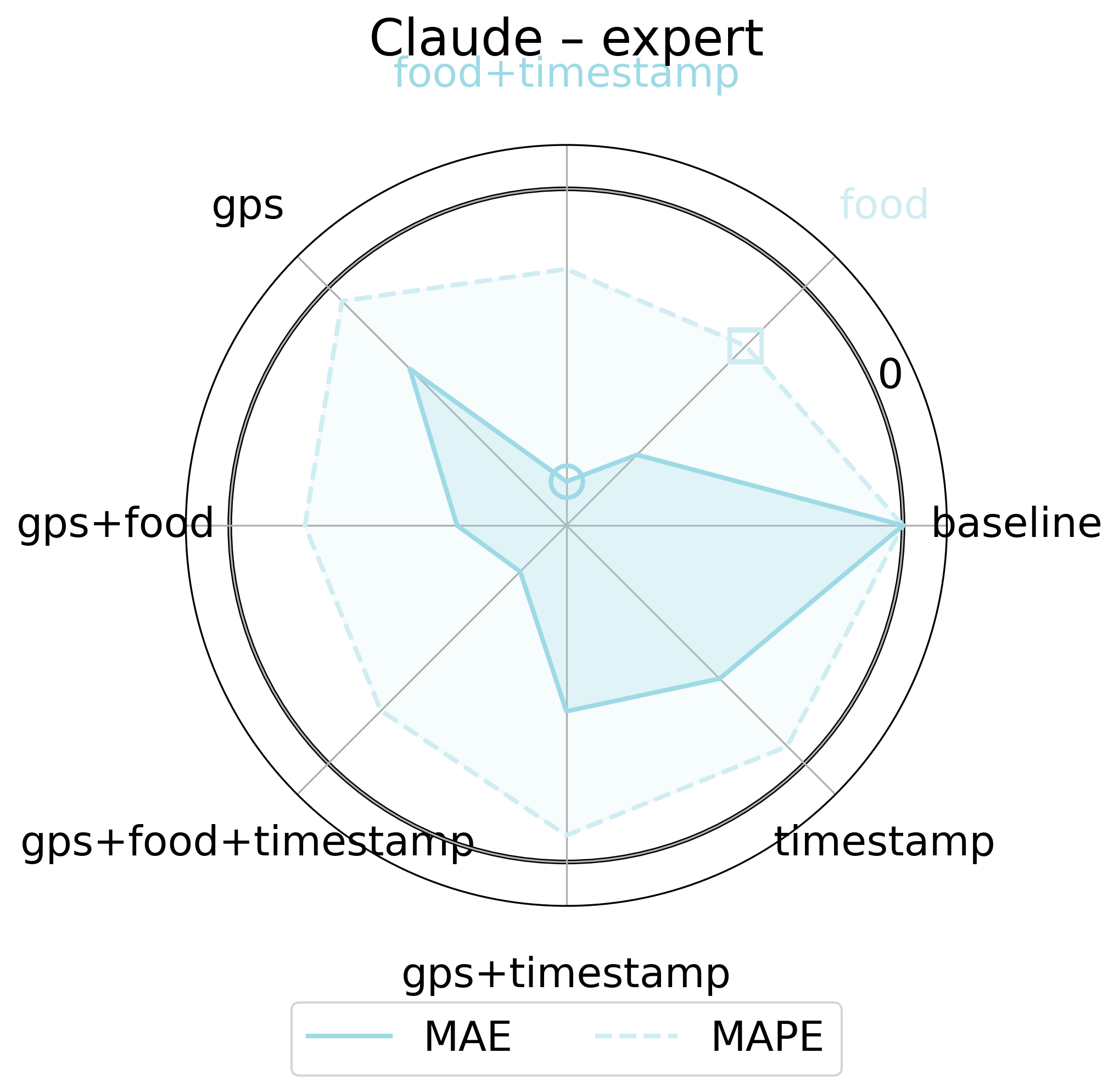}
    \caption{Claude Expert Persona}
    \label{fig:img2}
  \end{subfigure}
  \begin{subfigure}[t]{0.195\linewidth}
    \centering
    \includegraphics[width=\linewidth]{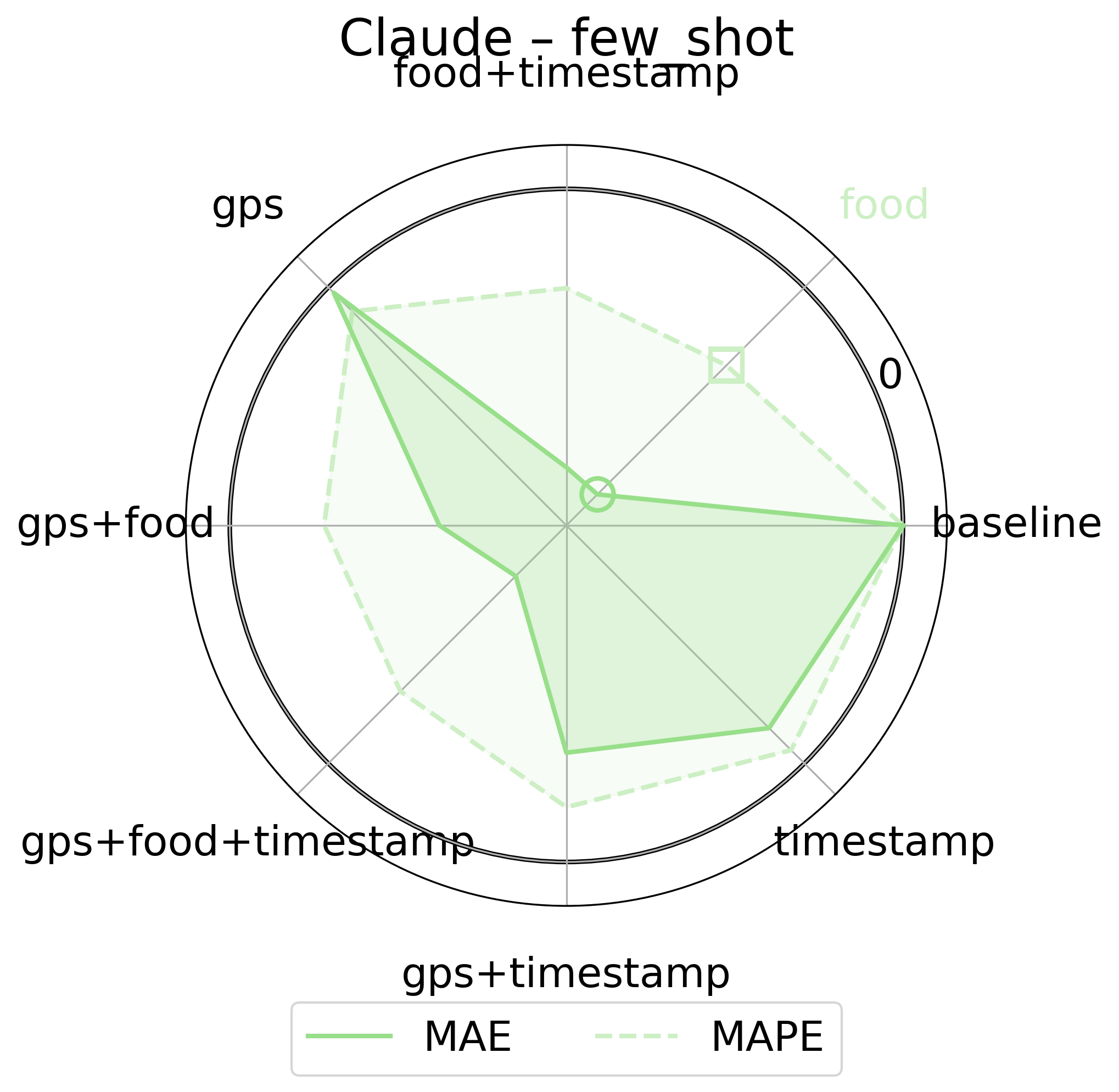}
    \caption{Claude Few-Shot}
    \label{fig:img3}
  \end{subfigure}
  \begin{subfigure}[t]{0.195\linewidth}
    \centering
    \includegraphics[width=\linewidth]{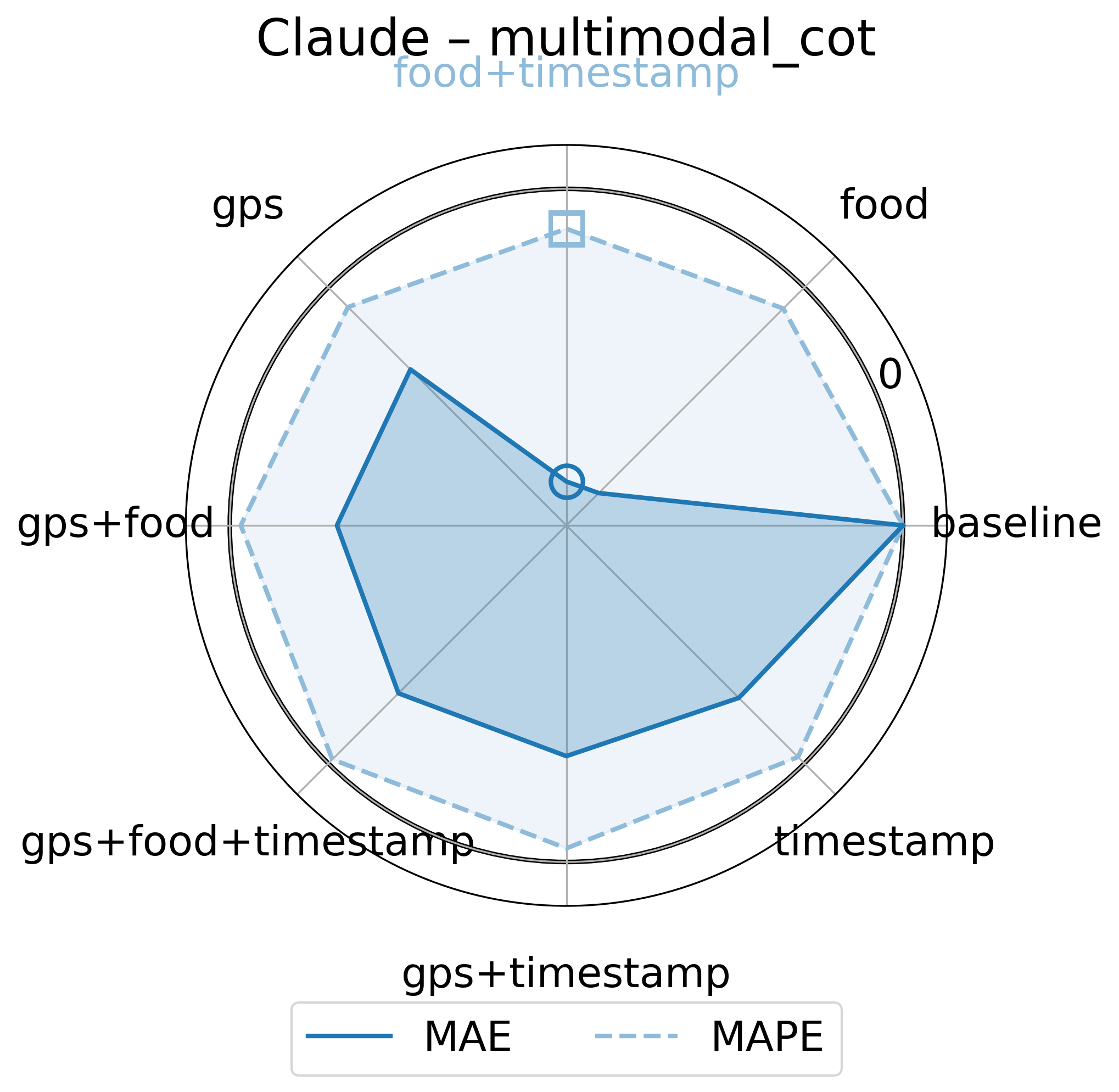}
    \caption{Claude Multimodal CoT}
    \label{fig:img4}
  \end{subfigure}
  \begin{subfigure}[t]{0.195\linewidth}
    \centering
    \includegraphics[width=\linewidth]{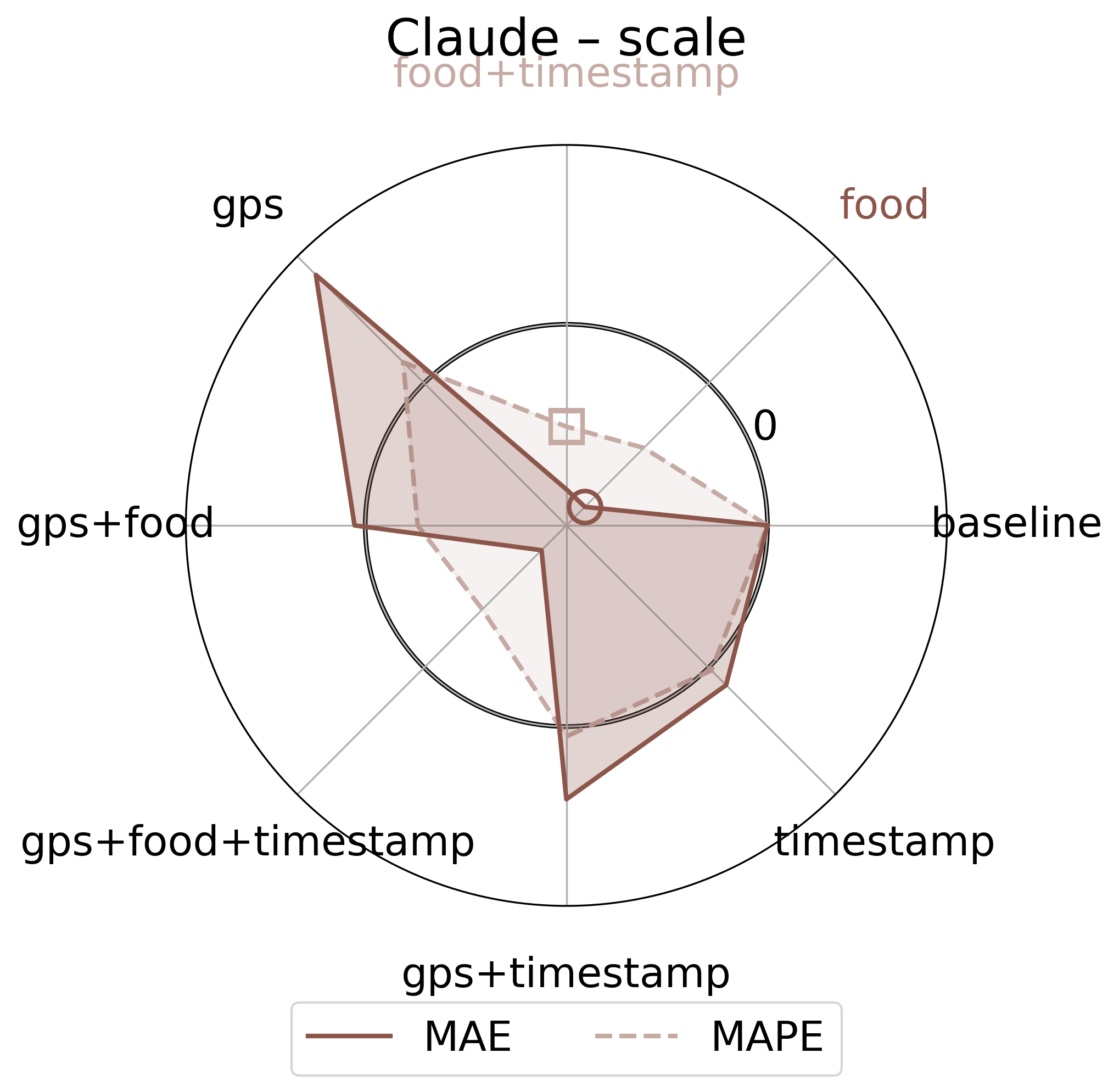}
    \caption{Claude Scale-Hint}
    \label{fig:img4}
  \end{subfigure}

  \begin{subfigure}[t]{0.195\linewidth}
    \centering
    \includegraphics[width=\linewidth]{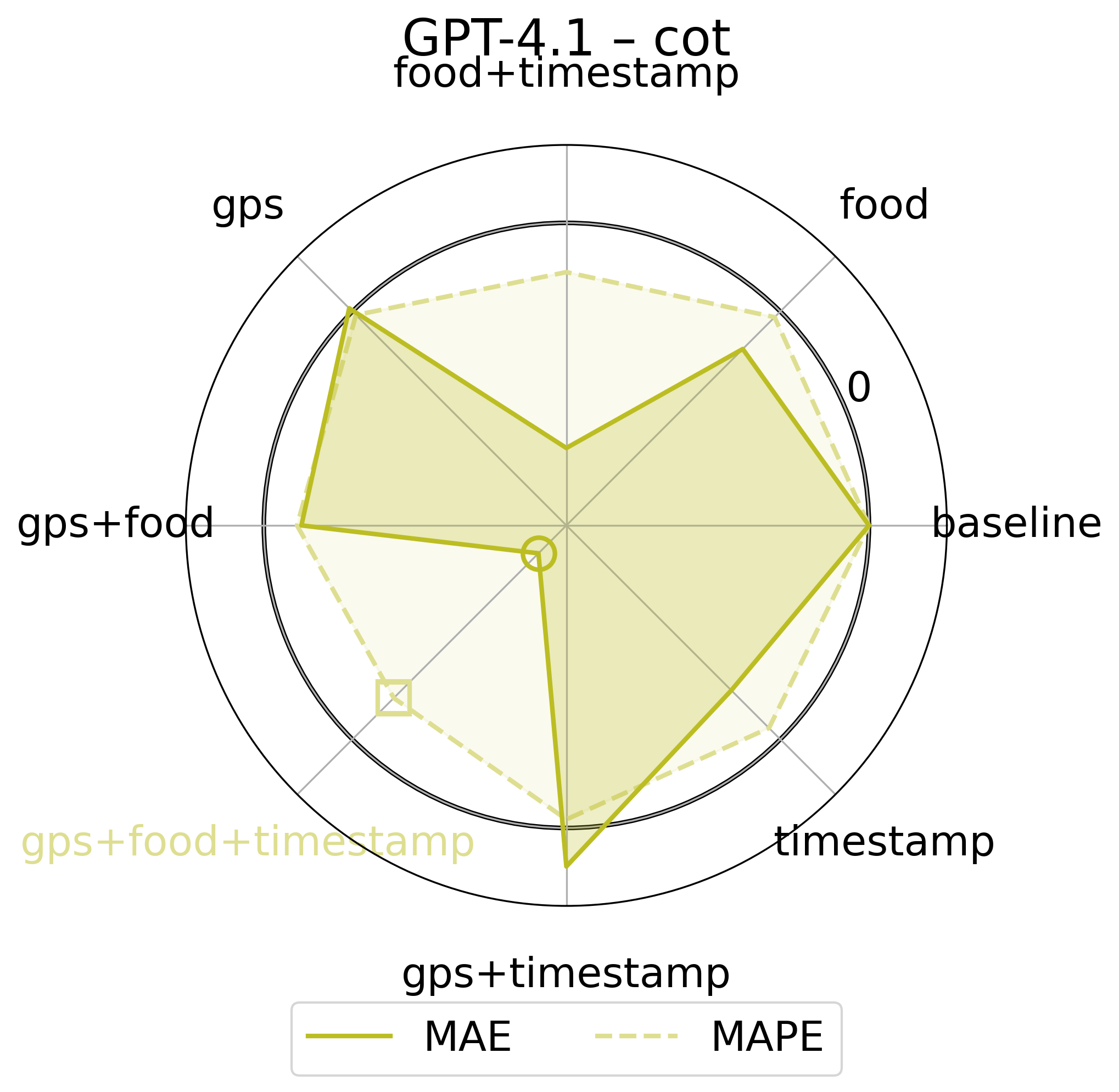}
    \caption{GPT-4.1 Chain-of-Thought}
    \label{fig:img1}
  \end{subfigure}
  \begin{subfigure}[t]{0.195\linewidth}
    \centering
    \includegraphics[width=\linewidth]{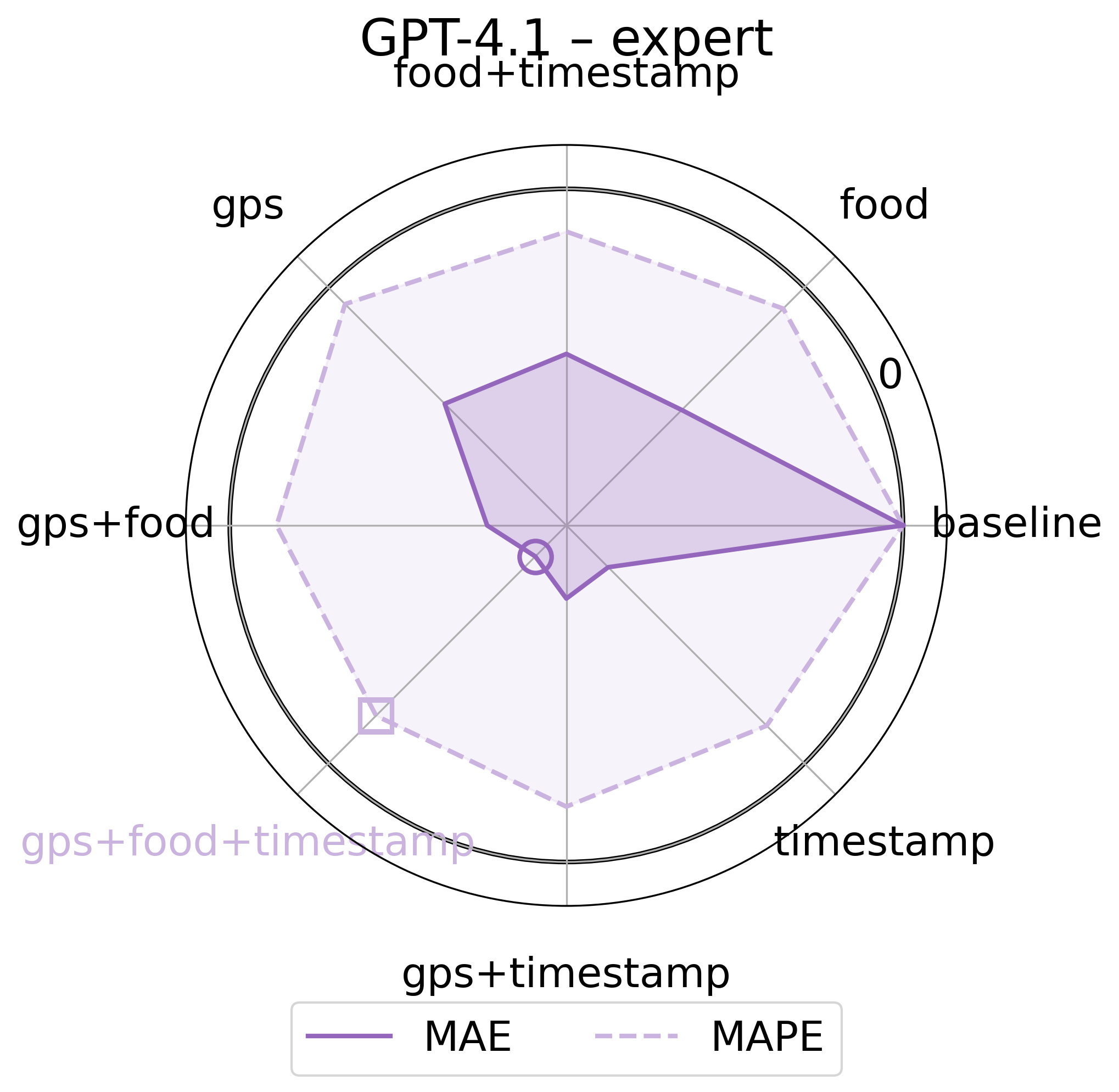}
    \caption{GPT-4.1 Expert Persona}
    \label{fig:img2}
  \end{subfigure}
  \begin{subfigure}[t]{0.195\linewidth}
    \centering
    \includegraphics[width=\linewidth]{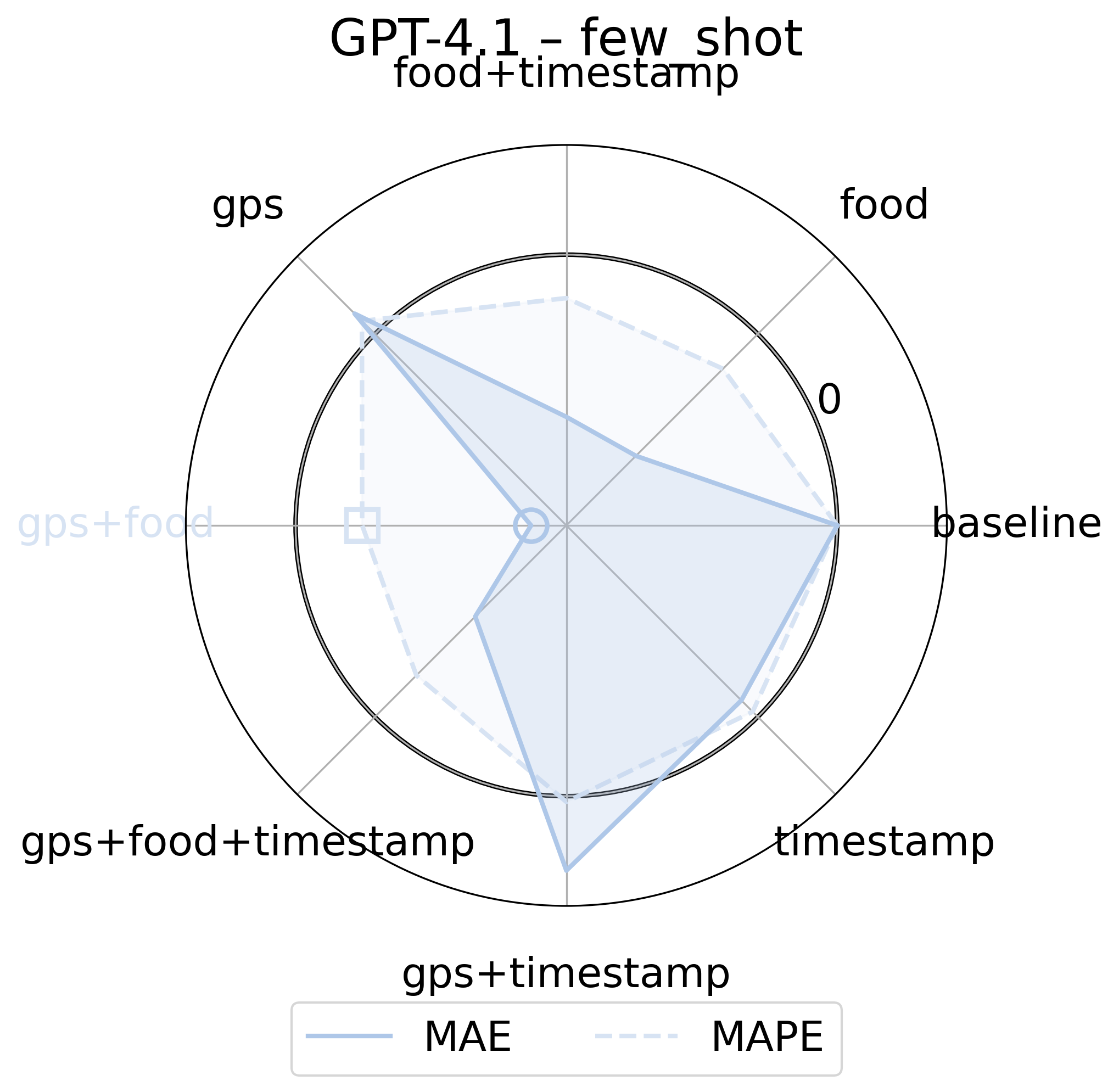}
    \caption{GPT-4.1 Few-Shot}
    \label{fig:img3}
  \end{subfigure}
  \begin{subfigure}[t]{0.195\linewidth}
    \centering
    \includegraphics[width=\linewidth]{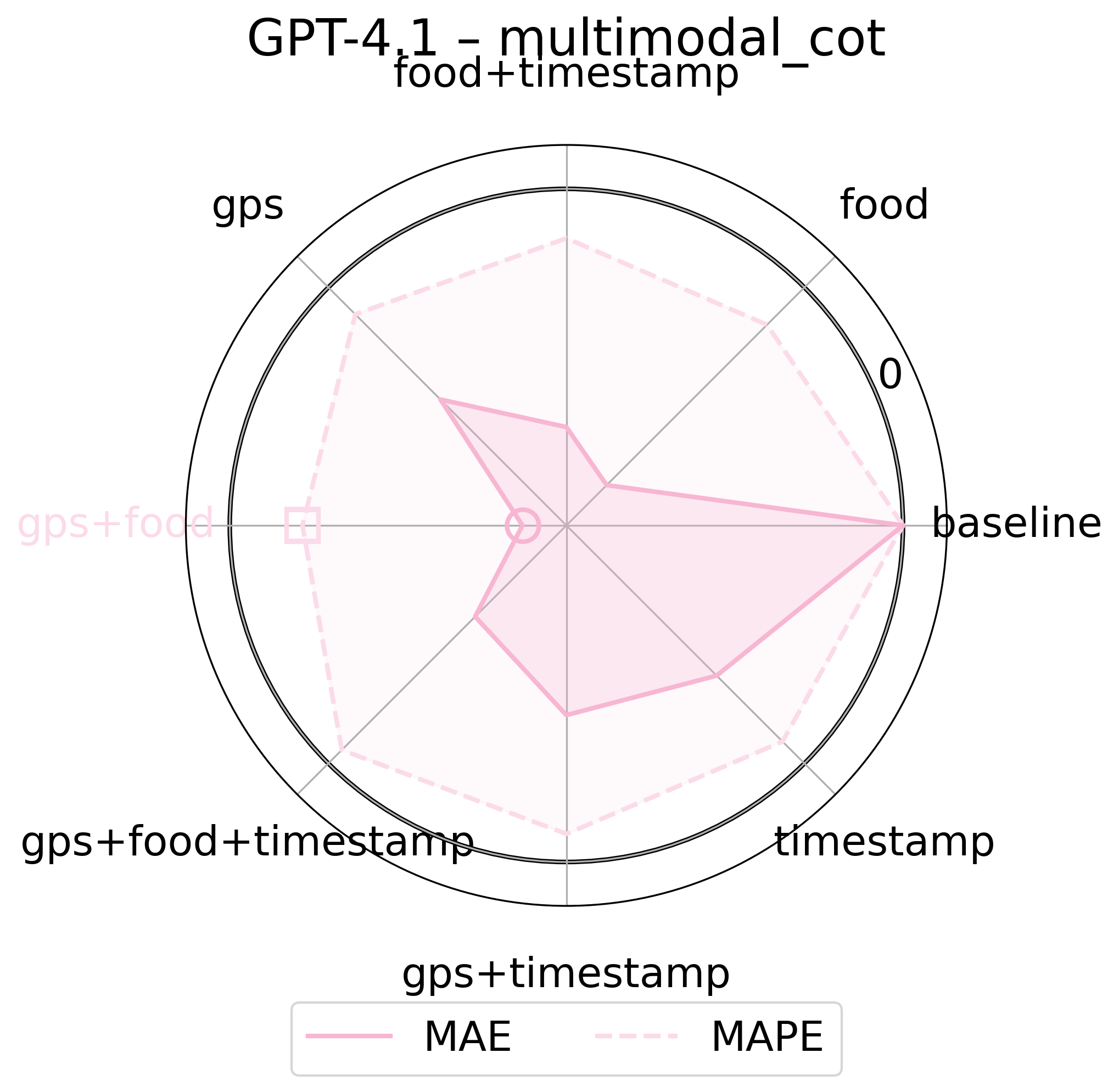}
    \caption{GPT-4.1 Multimodal CoT}
    \label{fig:img4}
  \end{subfigure}
  \begin{subfigure}[t]{0.195\linewidth}
    \centering
    \includegraphics[width=\linewidth]{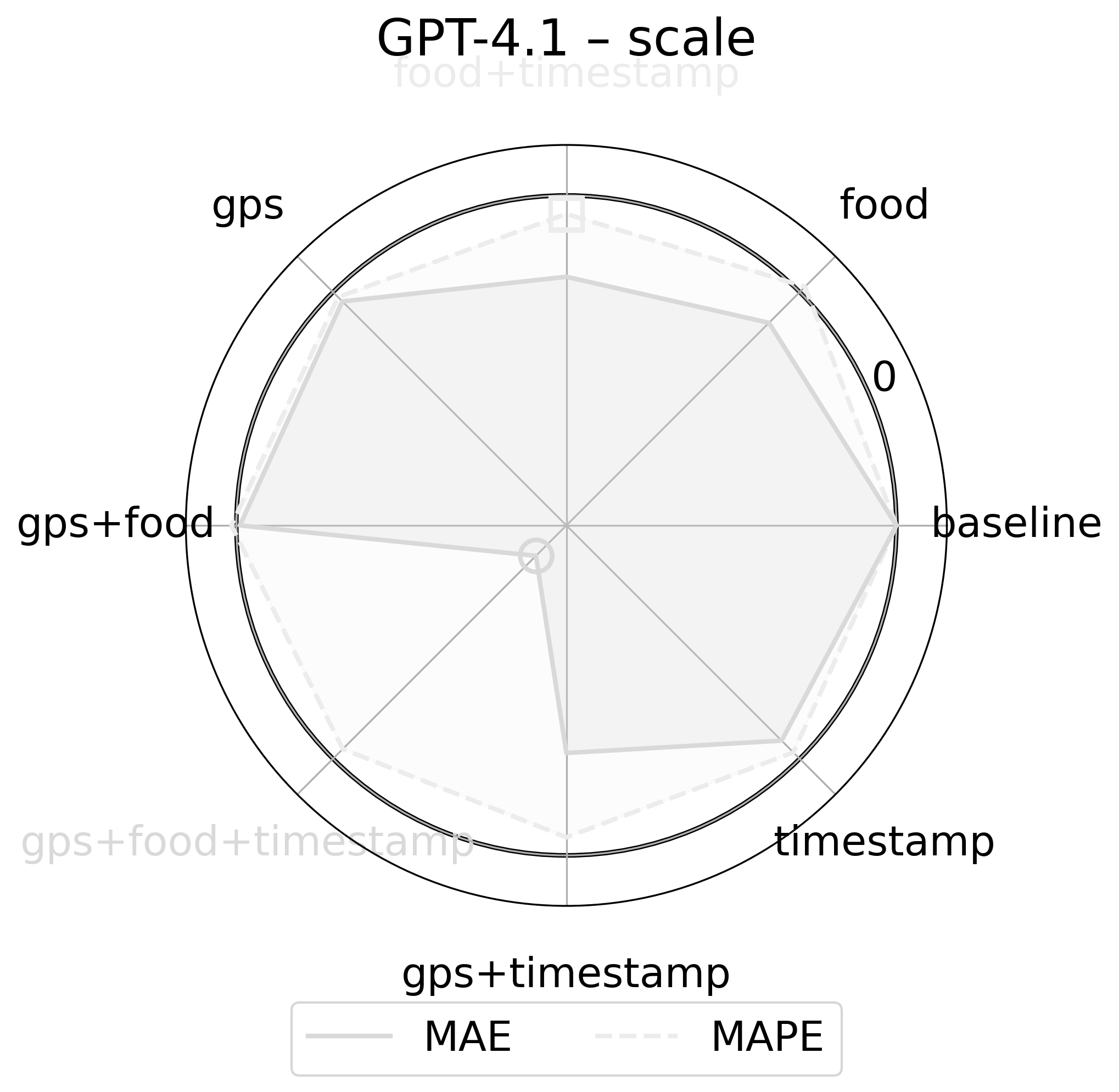}
    \caption{GPT-4.1 Scale-Hint}
    \label{fig:img4}
  \end{subfigure}

  \begin{subfigure}[t]{0.195\linewidth}
    \centering
    \includegraphics[width=\linewidth]{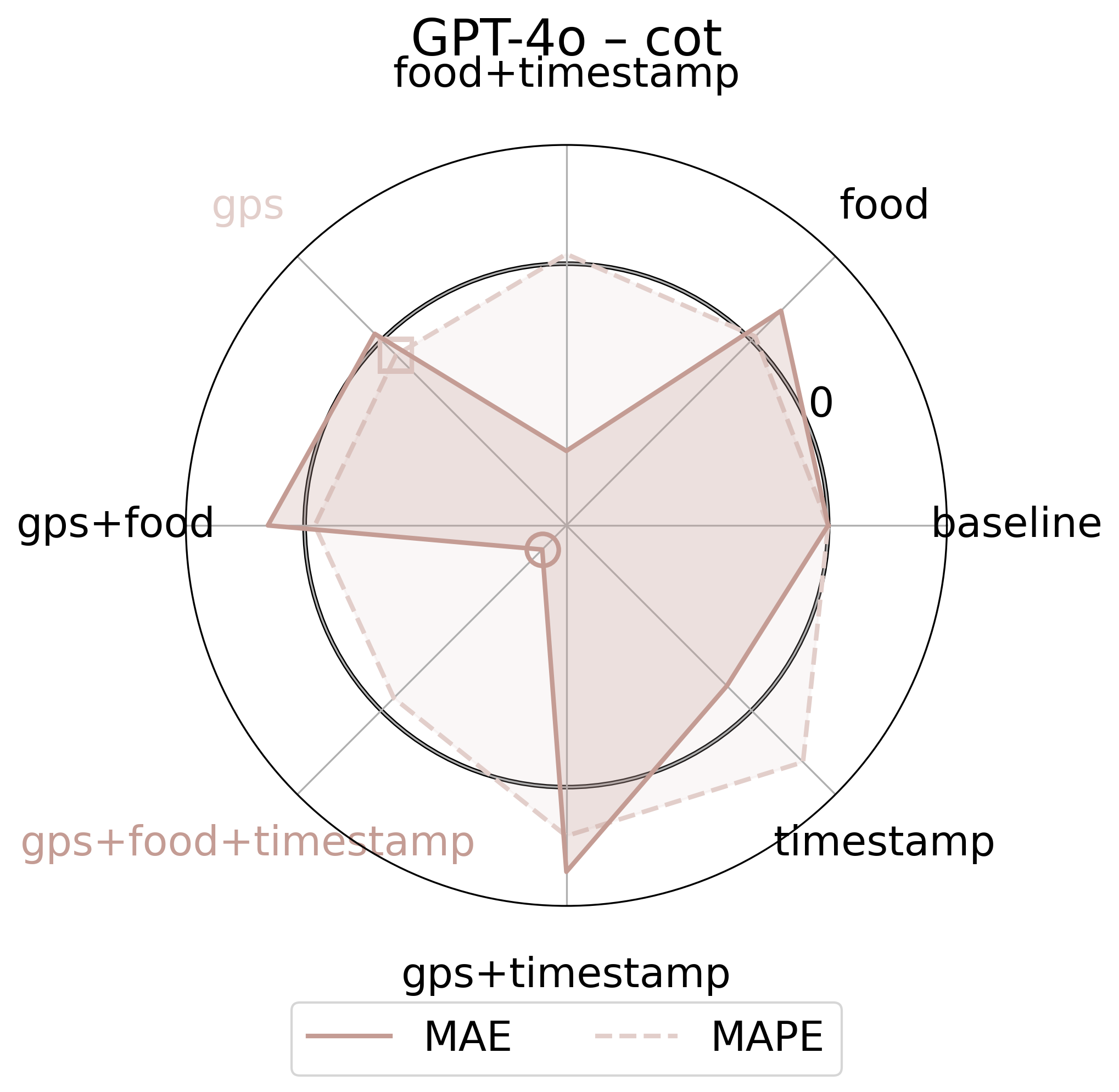}
    \caption{GPT 4o Chain-of-Thought}
    \label{fig:img1}
  \end{subfigure}
  \begin{subfigure}[t]{0.195\linewidth}
    \centering
    \includegraphics[width=\linewidth]{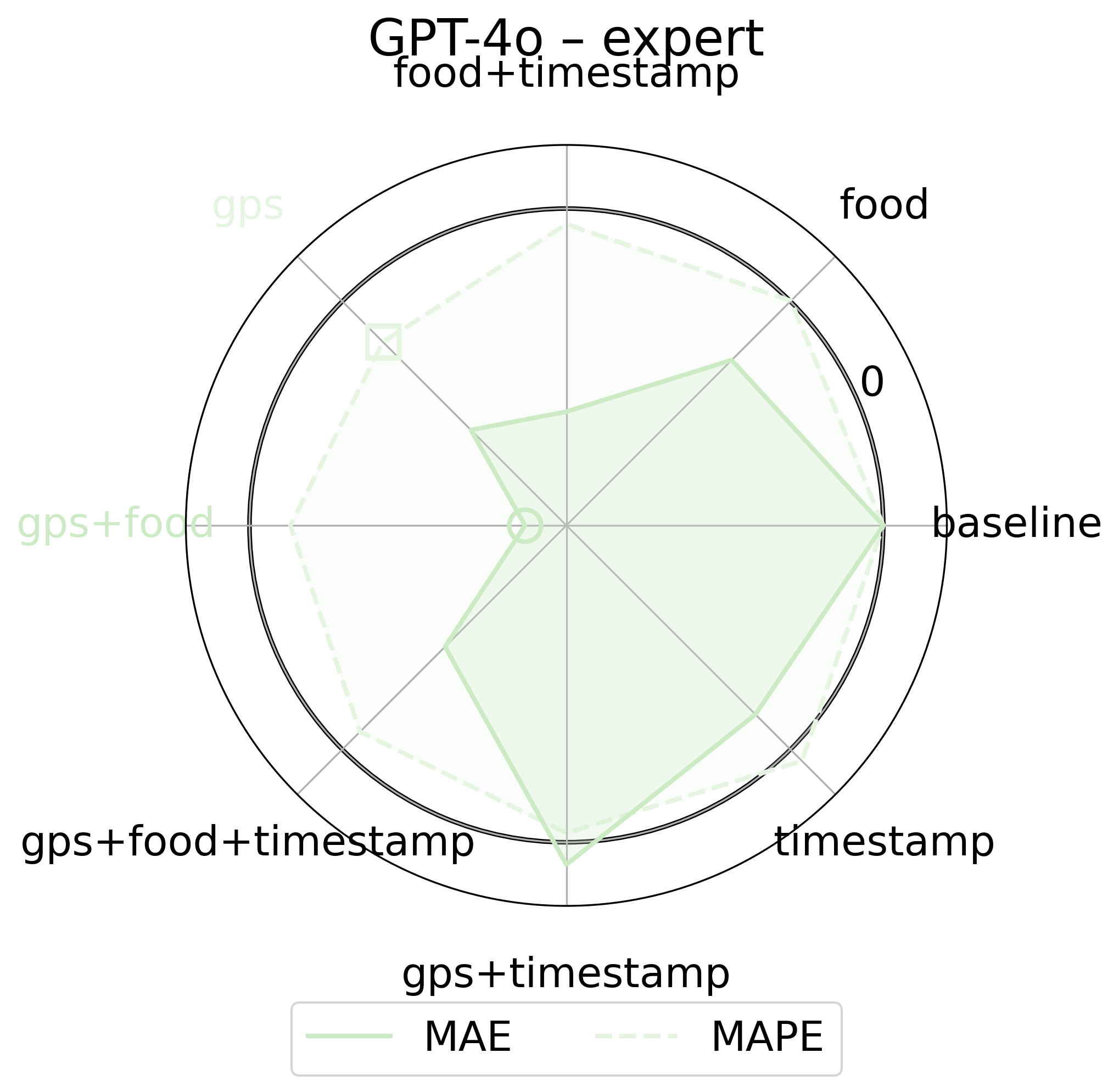}
    \caption{GPT-4o Expert Persona}
    \label{fig:img2}
  \end{subfigure}
  \begin{subfigure}[t]{0.195\linewidth}
    \centering
    \includegraphics[width=\linewidth]{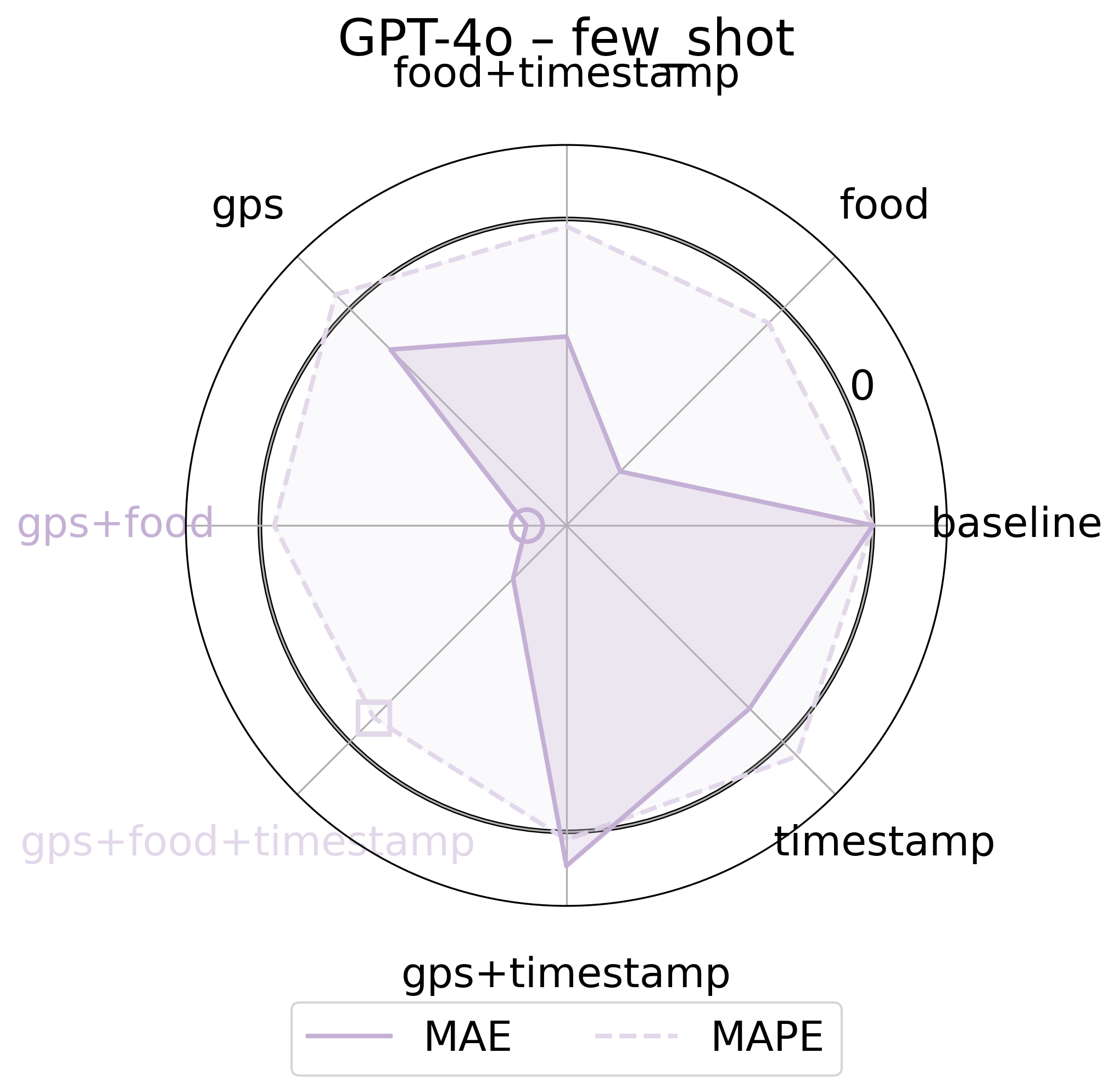}
    \caption{GPT-4o Few-Shot}
    \label{fig:img3}
  \end{subfigure}
  \begin{subfigure}[t]{0.195\linewidth}
    \centering
    \includegraphics[width=\linewidth]{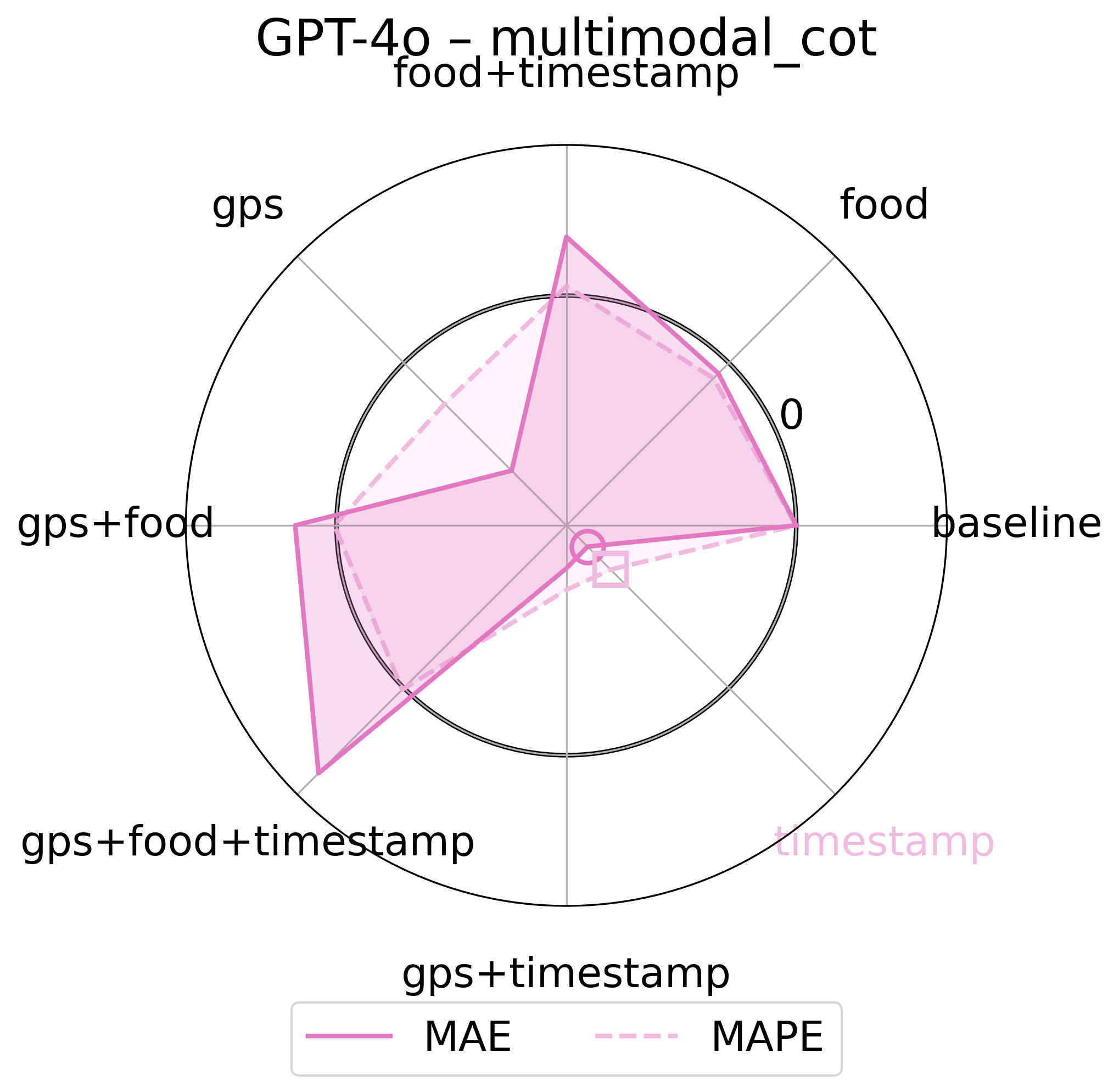}
    \caption{GPT-4o Multimodal CoT}
    \label{fig:img4}
  \end{subfigure}
  \begin{subfigure}[t]{0.195\linewidth}
    \centering
    \includegraphics[width=\linewidth]{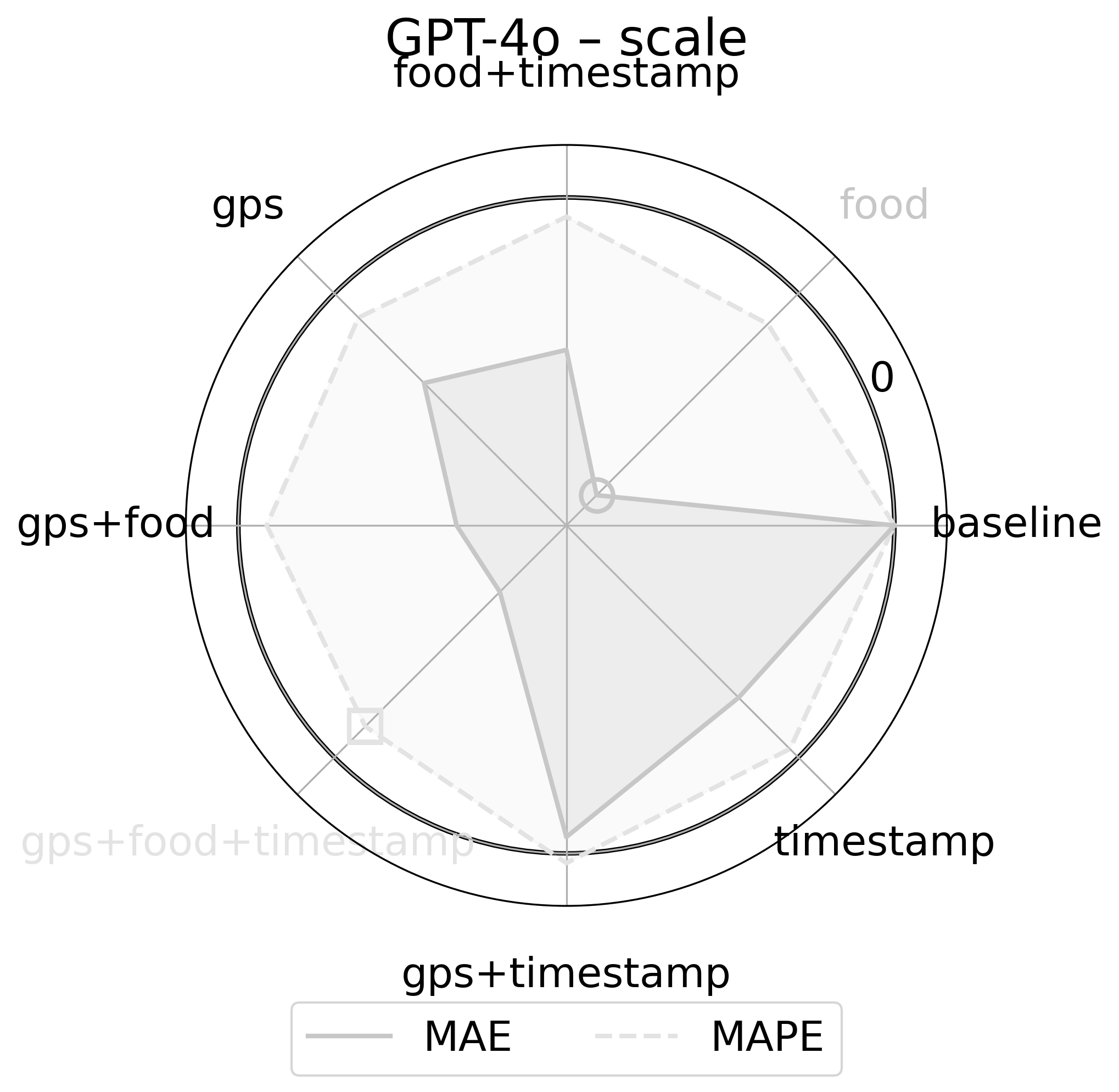}
    \caption{GPT-4o Scale-Hint}
    \label{fig:img4}
  \end{subfigure}

  \begin{subfigure}[t]{0.195\linewidth}
    \centering
    \includegraphics[width=\linewidth]{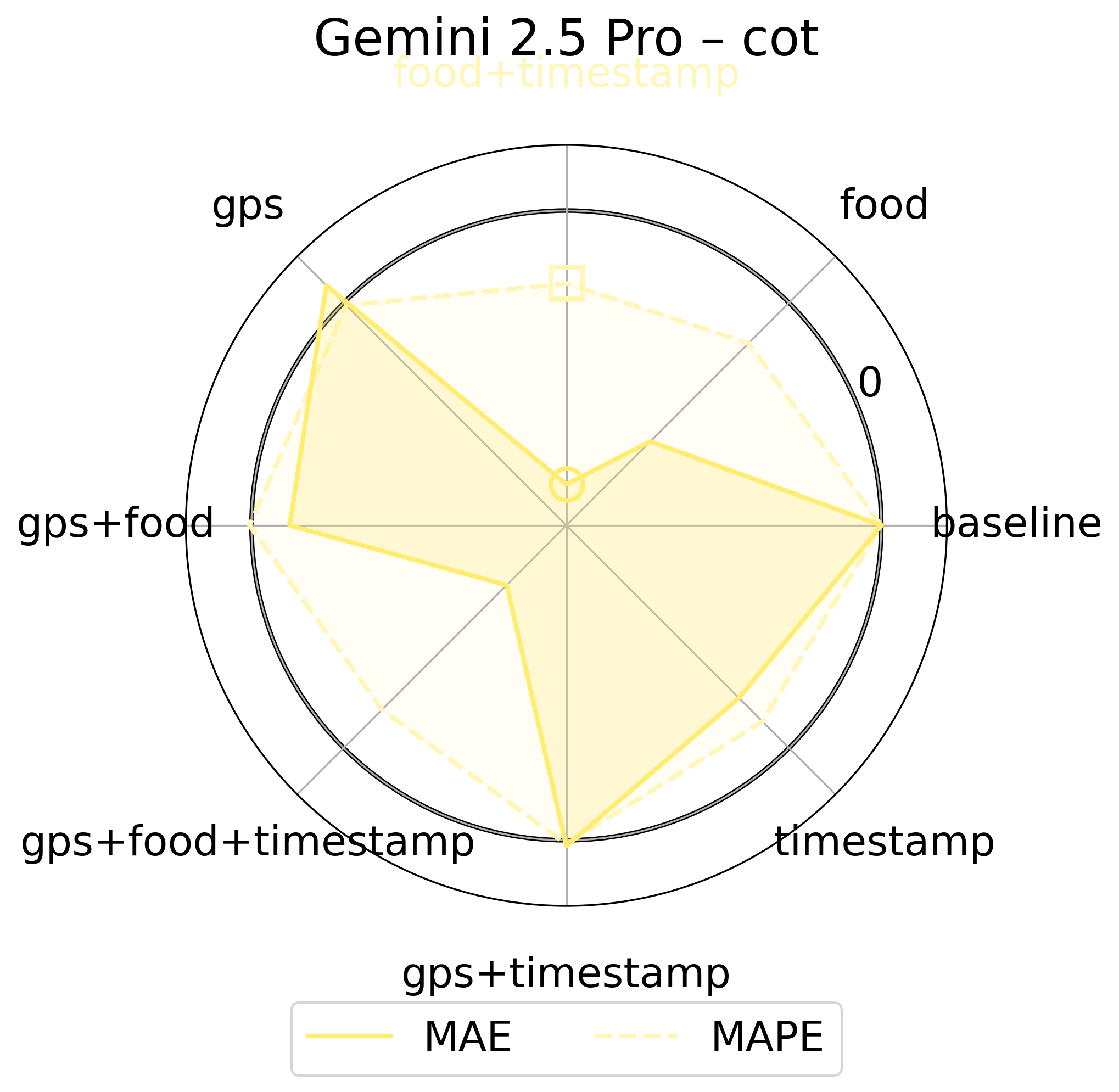}
    \caption{Gemini-2.5-Pro Chain-of-Thought}
    \label{fig:img1}
  \end{subfigure}
  \begin{subfigure}[t]{0.195\linewidth}
    \centering
    \includegraphics[width=\linewidth]{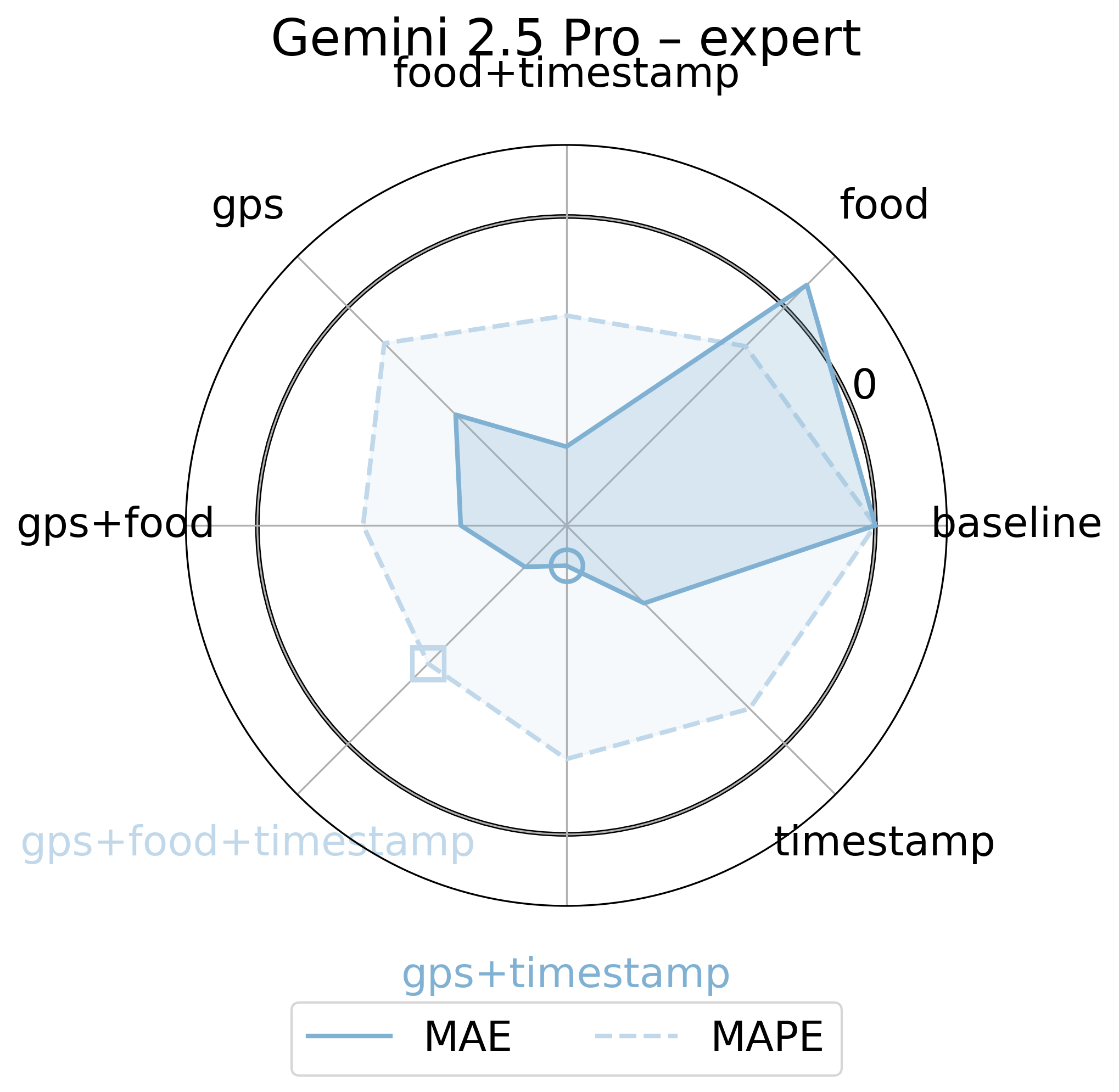}
    \caption{Gemini-2.5-Pro Expert Persona}
    \label{fig:img2}
  \end{subfigure}
  \begin{subfigure}[t]{0.195\linewidth}
    \centering
    \includegraphics[width=\linewidth]{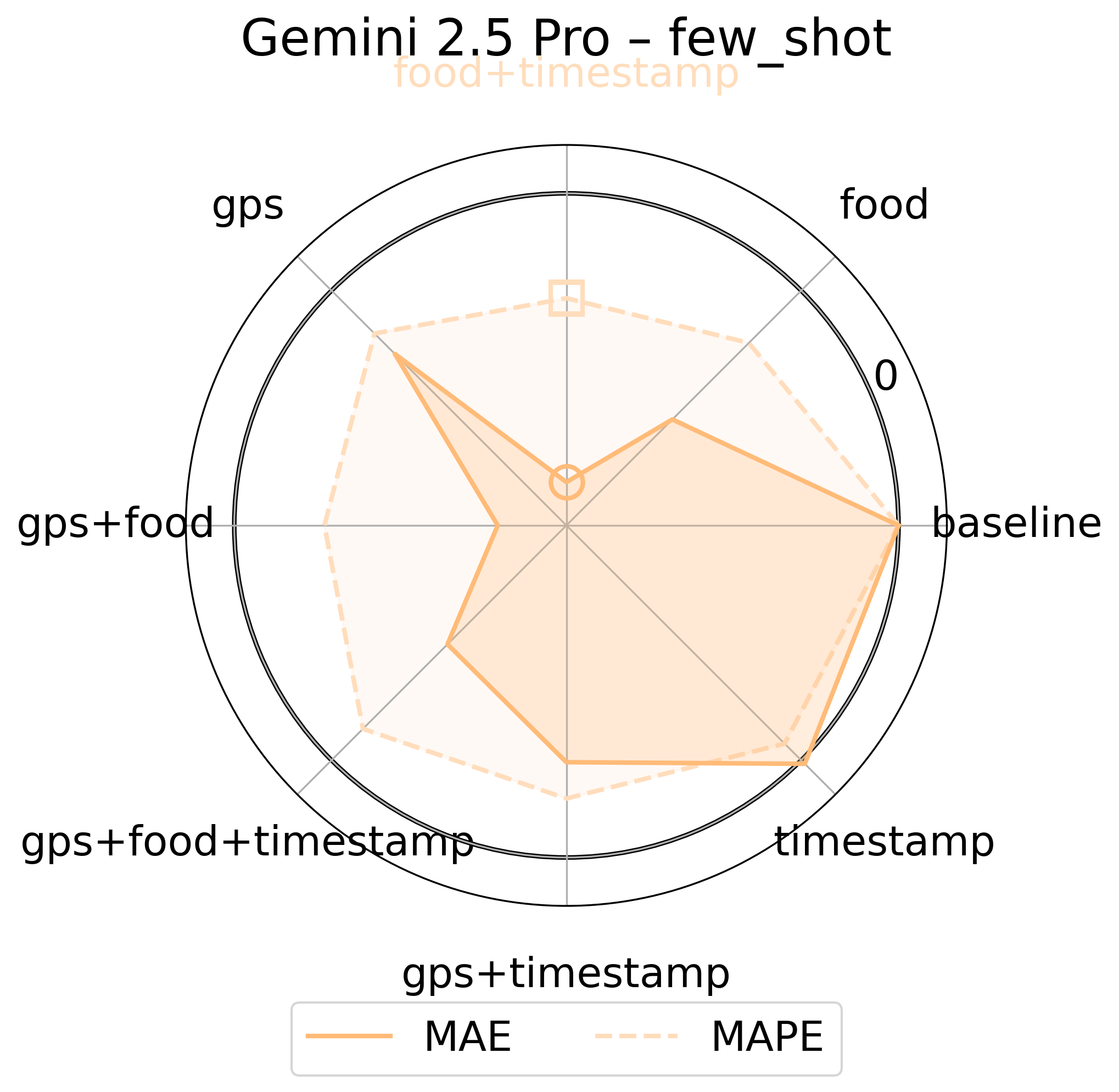}
    \caption{Gemini-2.5-Pro Few-Shot}
    \label{fig:img3}
  \end{subfigure}
  \begin{subfigure}[t]{0.195\linewidth}
    \centering
    \includegraphics[width=\linewidth]{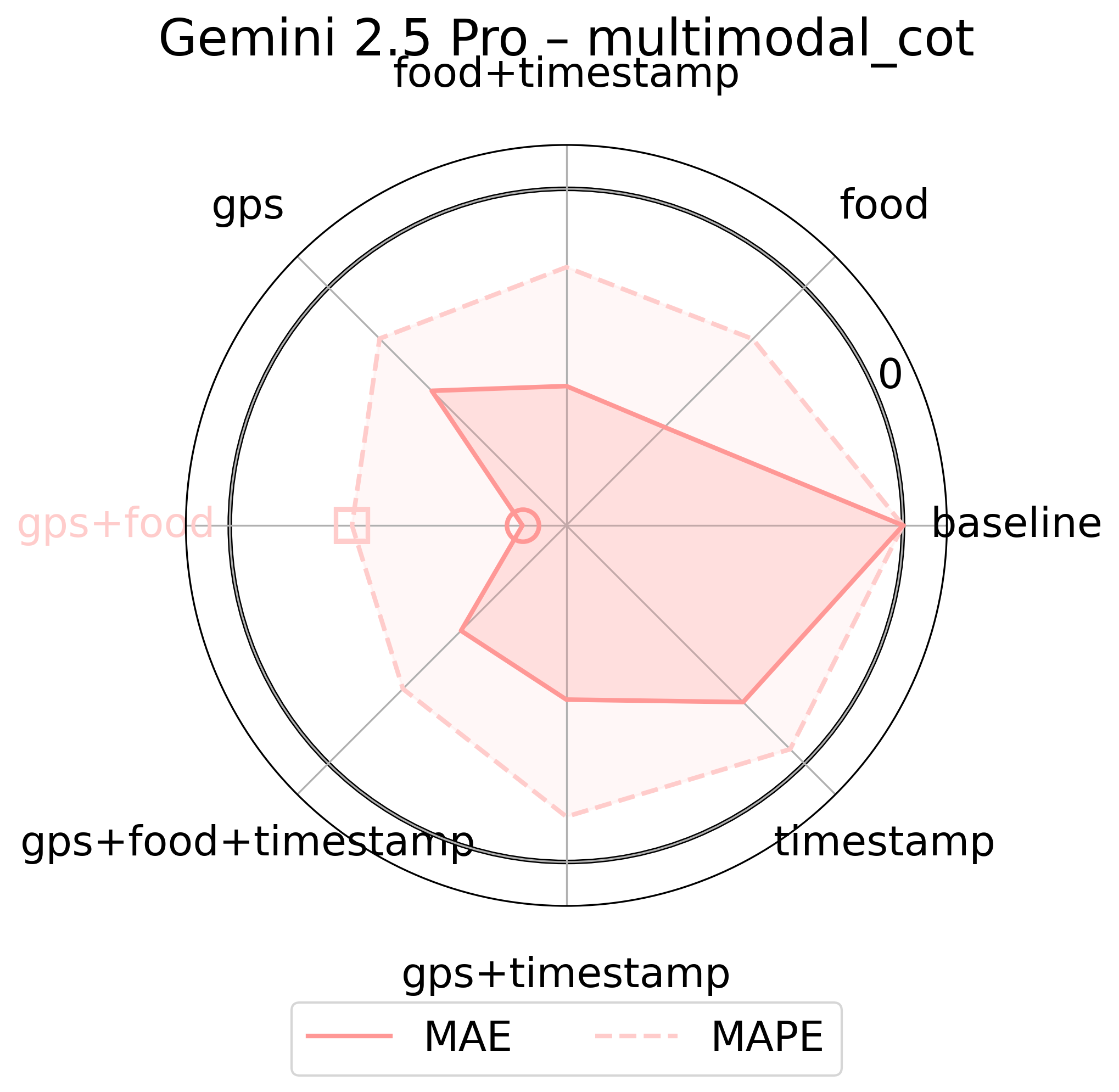}
    \caption{Gemini-2.5-Pro Multimodal CoT}
    \label{fig:img4}
  \end{subfigure}
  \begin{subfigure}[t]{0.195\linewidth}
    \centering
    \includegraphics[width=\linewidth]{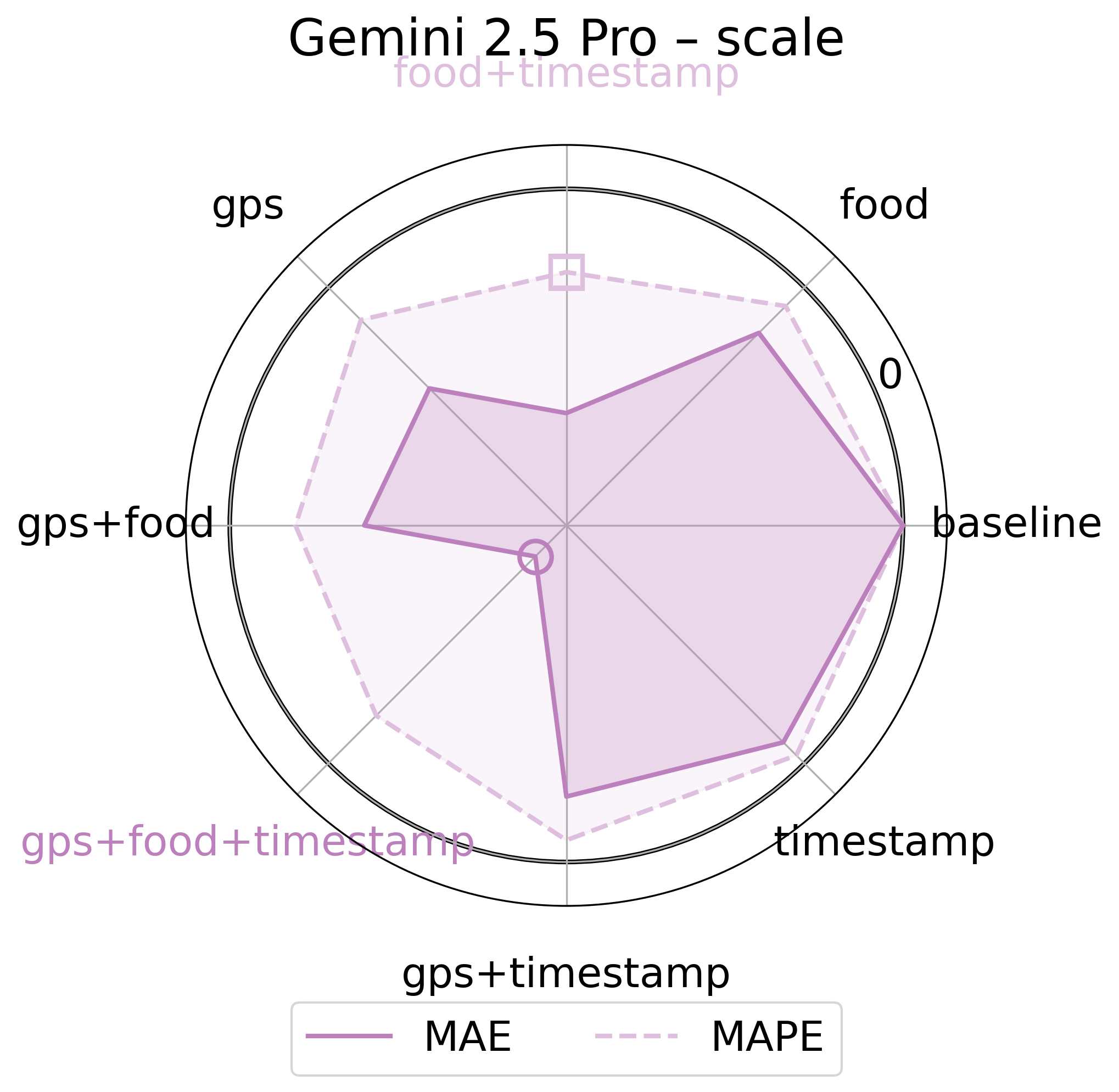}
    \caption{Gemini-2.5-Pro Scale-Hint}
    \label{fig:img4}
  \end{subfigure}

  \caption{\textbf{Averaged Experiment 2 Results For Closed-Weight Models.} MAE (solid lines) and MAPE (dashed lines) are plotted for various contextual metadata combinations for a given closed-weight model. Each spoke represents an error in a metadata combination; closer proximity to the center signifies a reduction in error relative to the baseline prompt. Colored markers denote the \textsc{Best-Metadata} configuration for each metric.}
  \label{fig:exp_2_closed_weight_models}
\end{figure*}

\begin{figure*}[htbp] 
  \centering

  \begin{subfigure}[t]{0.195\linewidth}
    \centering
    \includegraphics[width=\linewidth]{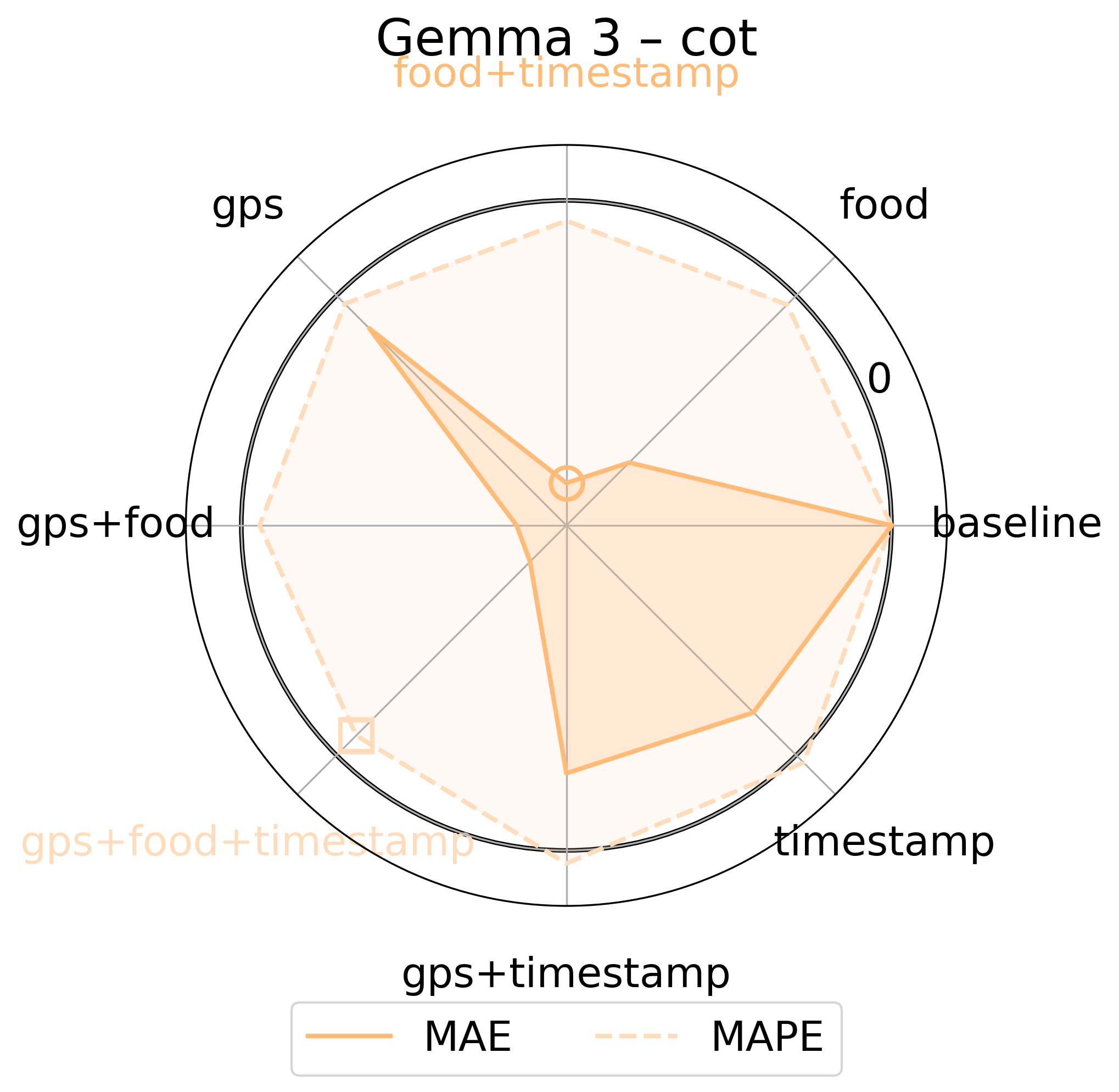}
    \caption{Gemma3 Chain-of-Thought}
    \label{fig:img1}
  \end{subfigure}
  \begin{subfigure}[t]{0.195\linewidth}
    \centering
    \includegraphics[width=\linewidth]{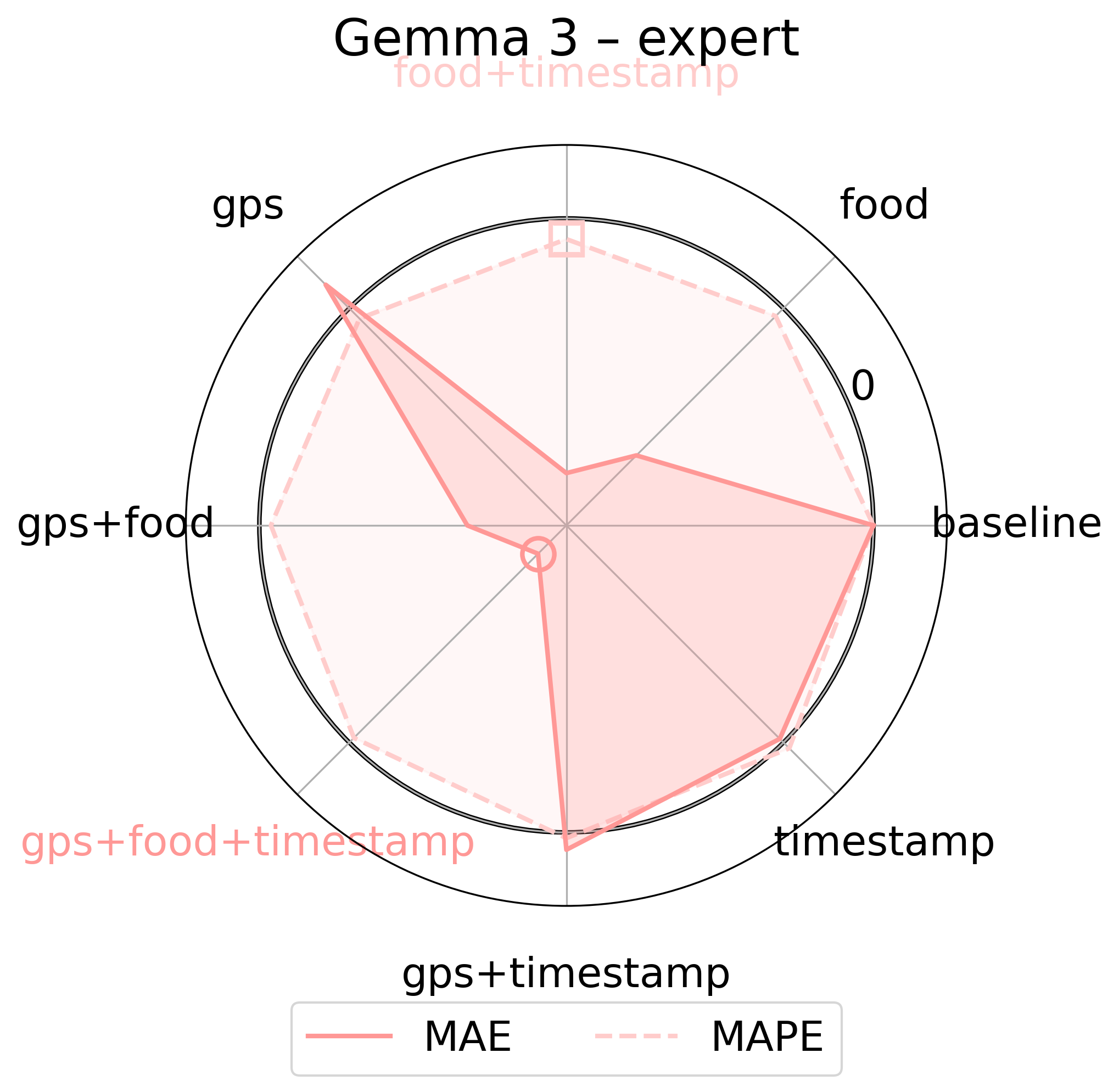}
    \caption{Gemma3 Expert Persona}
    \label{fig:img2}
  \end{subfigure}
  \begin{subfigure}[t]{0.195\linewidth}
    \centering
    \includegraphics[width=\linewidth]{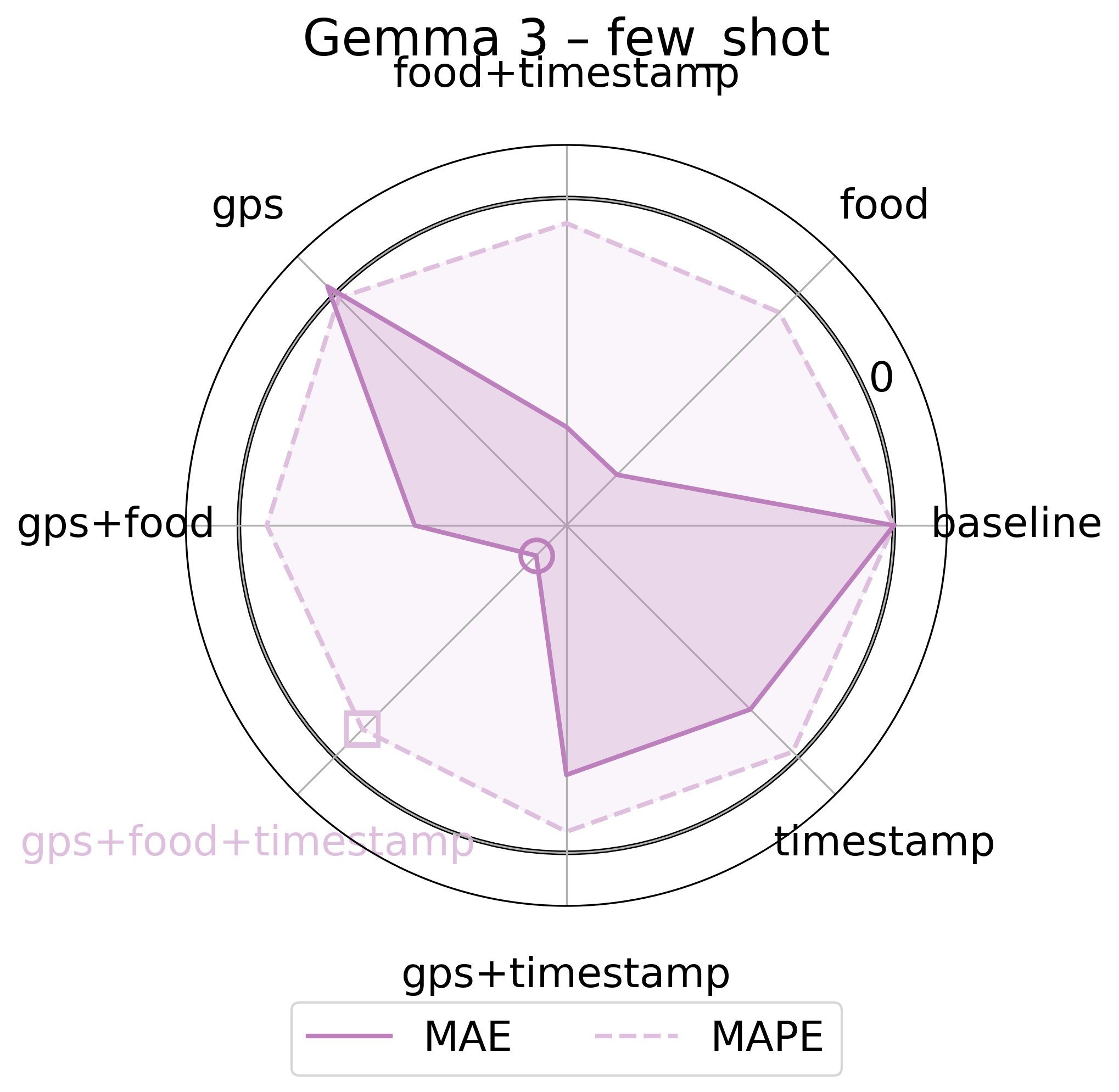}
    \caption{Gemma3 Few-Shot}
    \label{fig:img3}
  \end{subfigure}
  \begin{subfigure}[t]{0.195\linewidth}
    \centering
    \includegraphics[width=\linewidth]{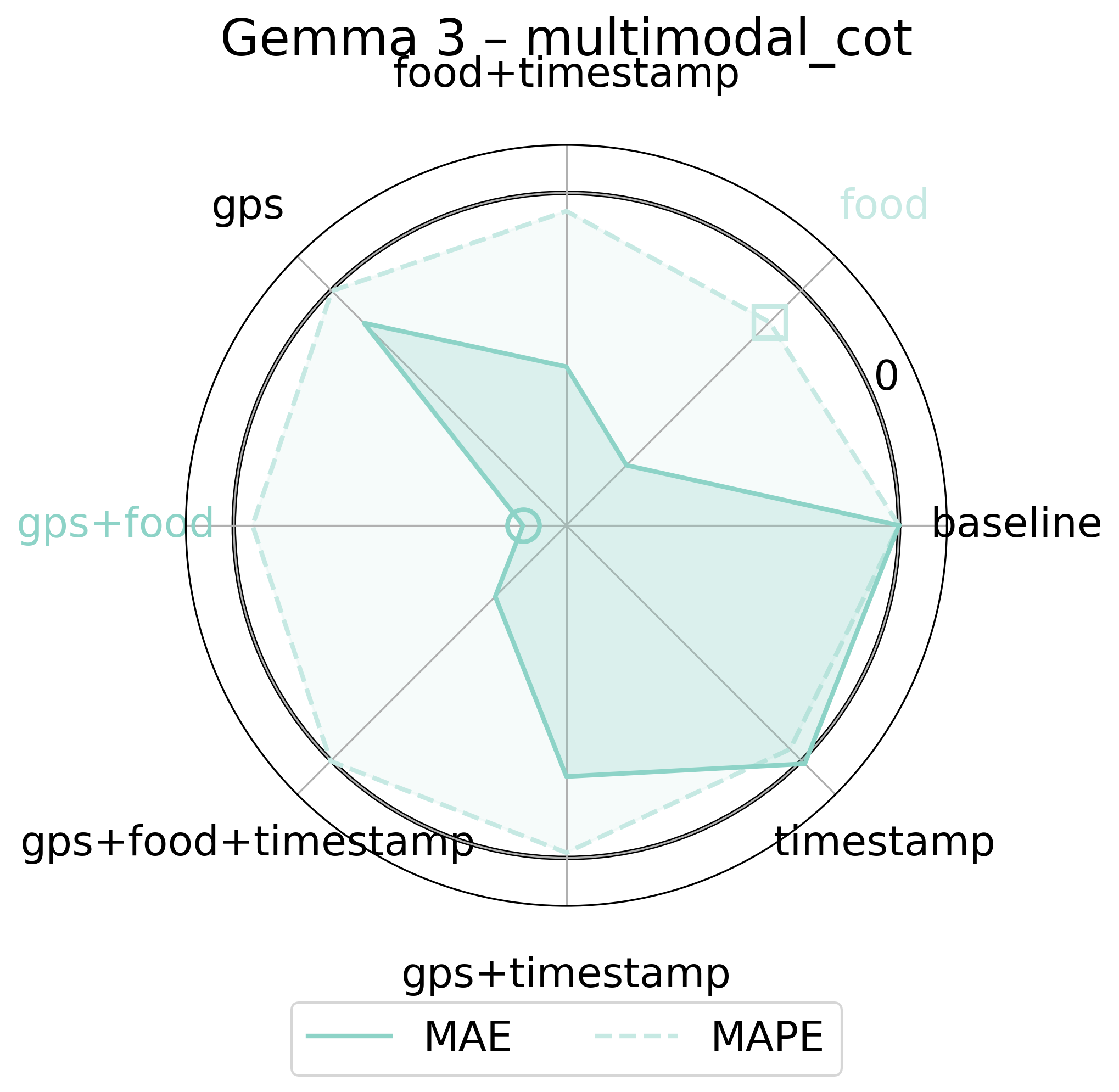}
    \caption{Gemma3 Multimodal CoT}
    \label{fig:img4}
  \end{subfigure}
  \begin{subfigure}[t]{0.195\linewidth}
    \centering
    \includegraphics[width=\linewidth]{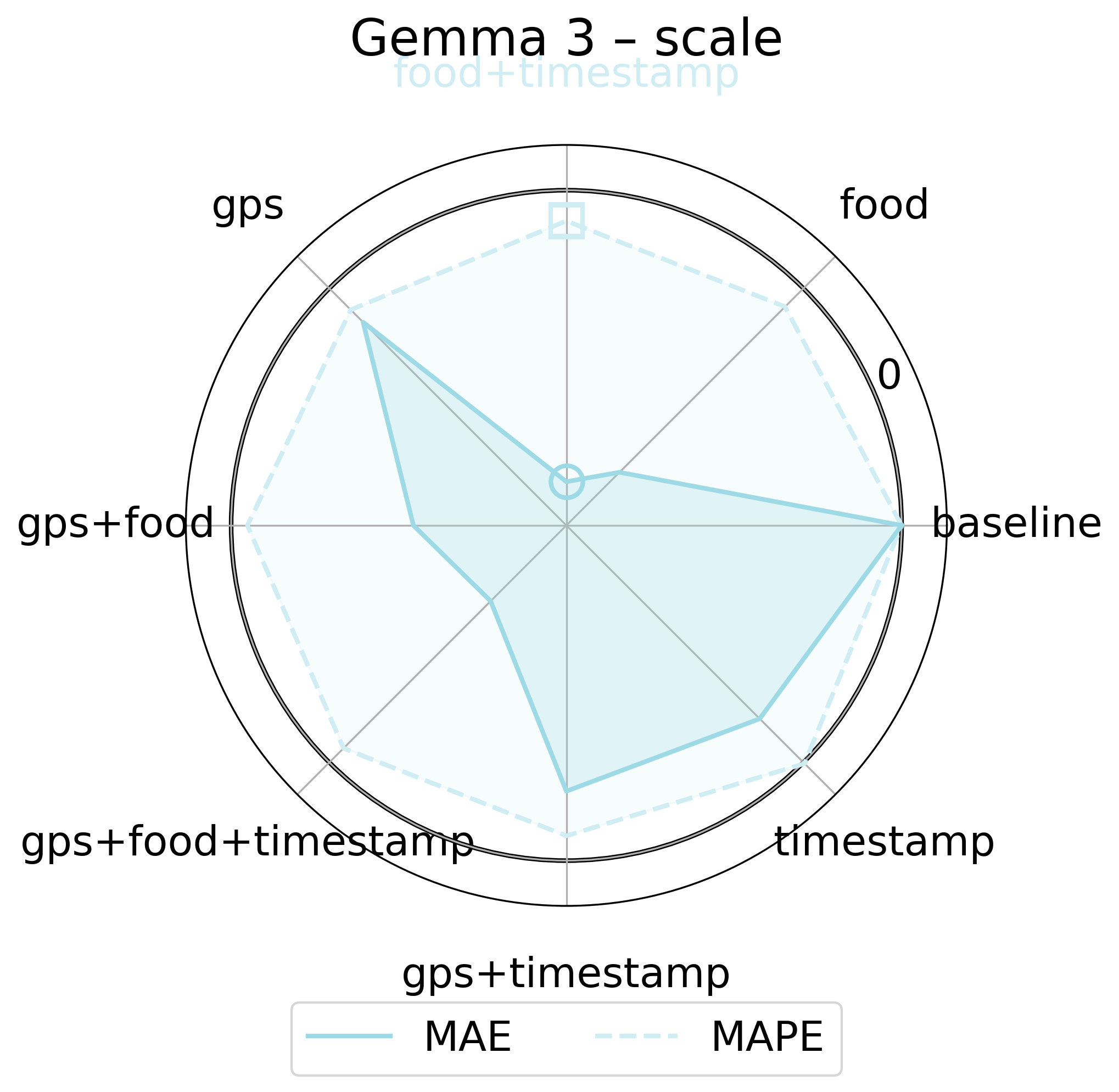}
    \caption{Gemma3 Scale-Hint}
    \label{fig:img4}
  \end{subfigure}

  \begin{subfigure}[t]{0.195\linewidth}
    \centering
    \includegraphics[width=\linewidth]{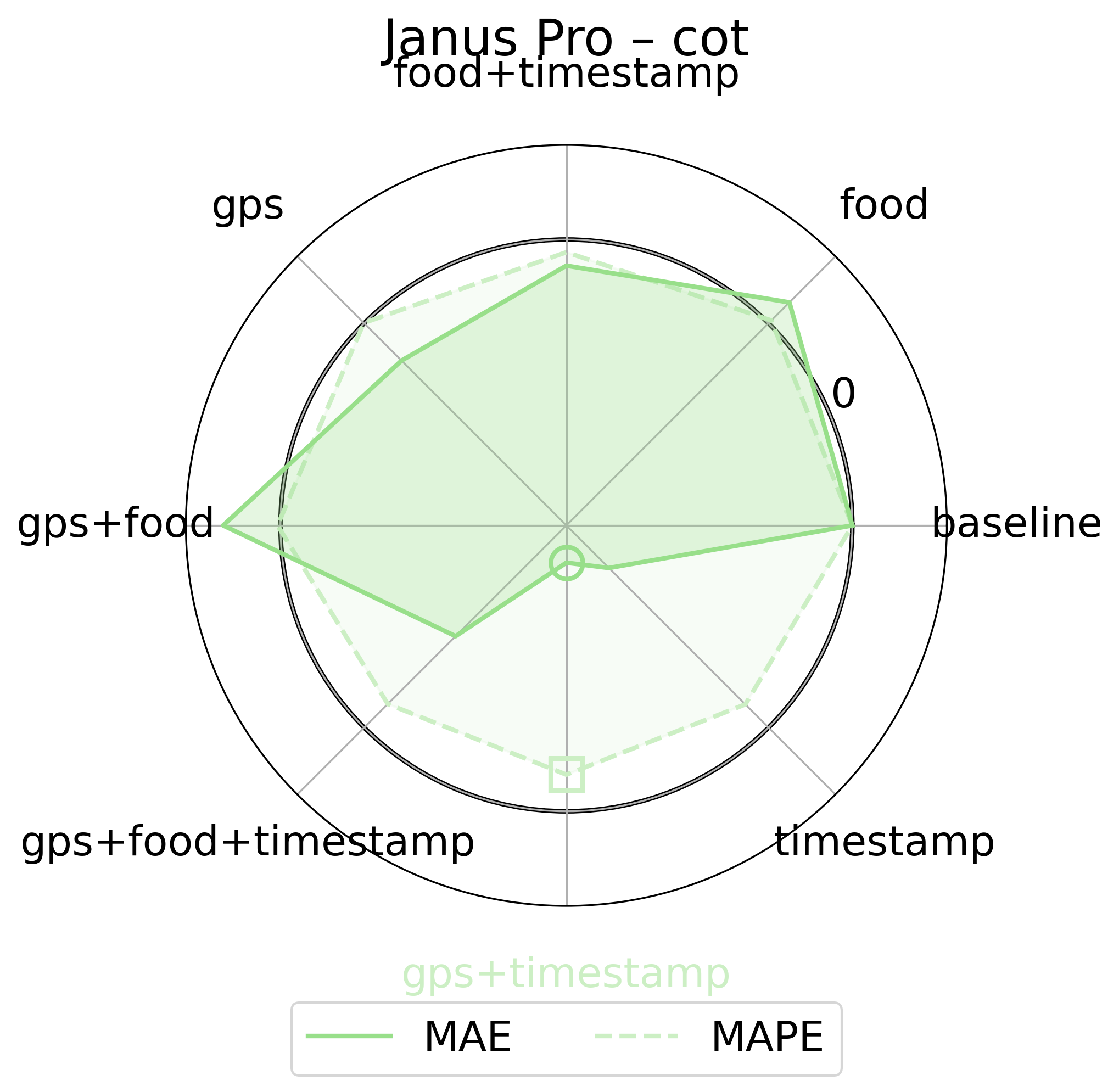}
    \caption{Janus-Pro Chain-of-Thought}
    \label{fig:img1}
  \end{subfigure}
  \begin{subfigure}[t]{0.195\linewidth}
    \centering
    \includegraphics[width=\linewidth]{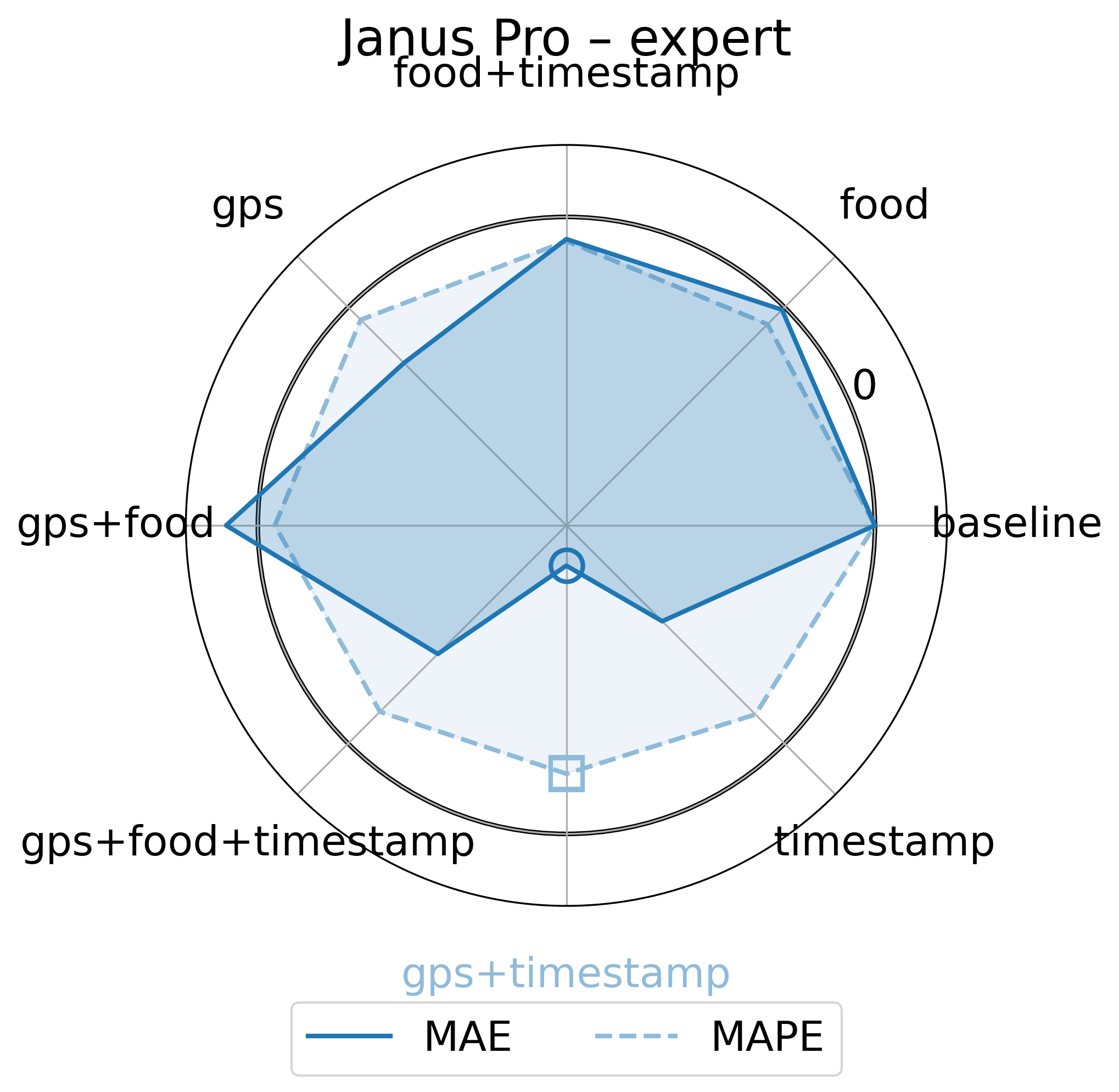}
    \caption{Janus-Pro Expert Persona}
    \label{fig:img2}
  \end{subfigure}
  \begin{subfigure}[t]{0.195\linewidth}
    \centering
    \includegraphics[width=\linewidth]{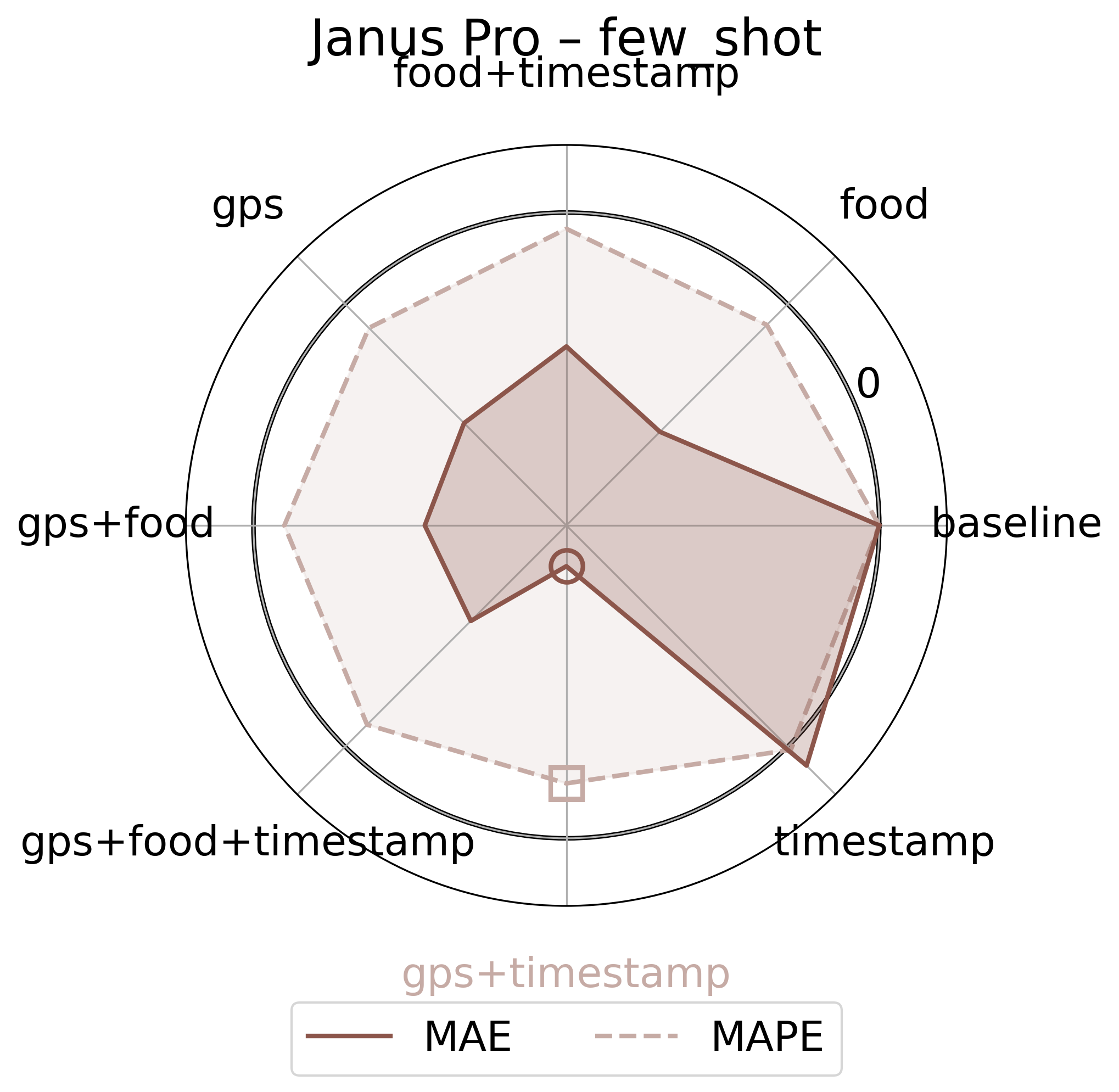}
    \caption{Janus-Pro Few-Shot}
    \label{fig:img3}
  \end{subfigure}
  \begin{subfigure}[t]{0.195\linewidth}
    \centering
    \includegraphics[width=\linewidth]{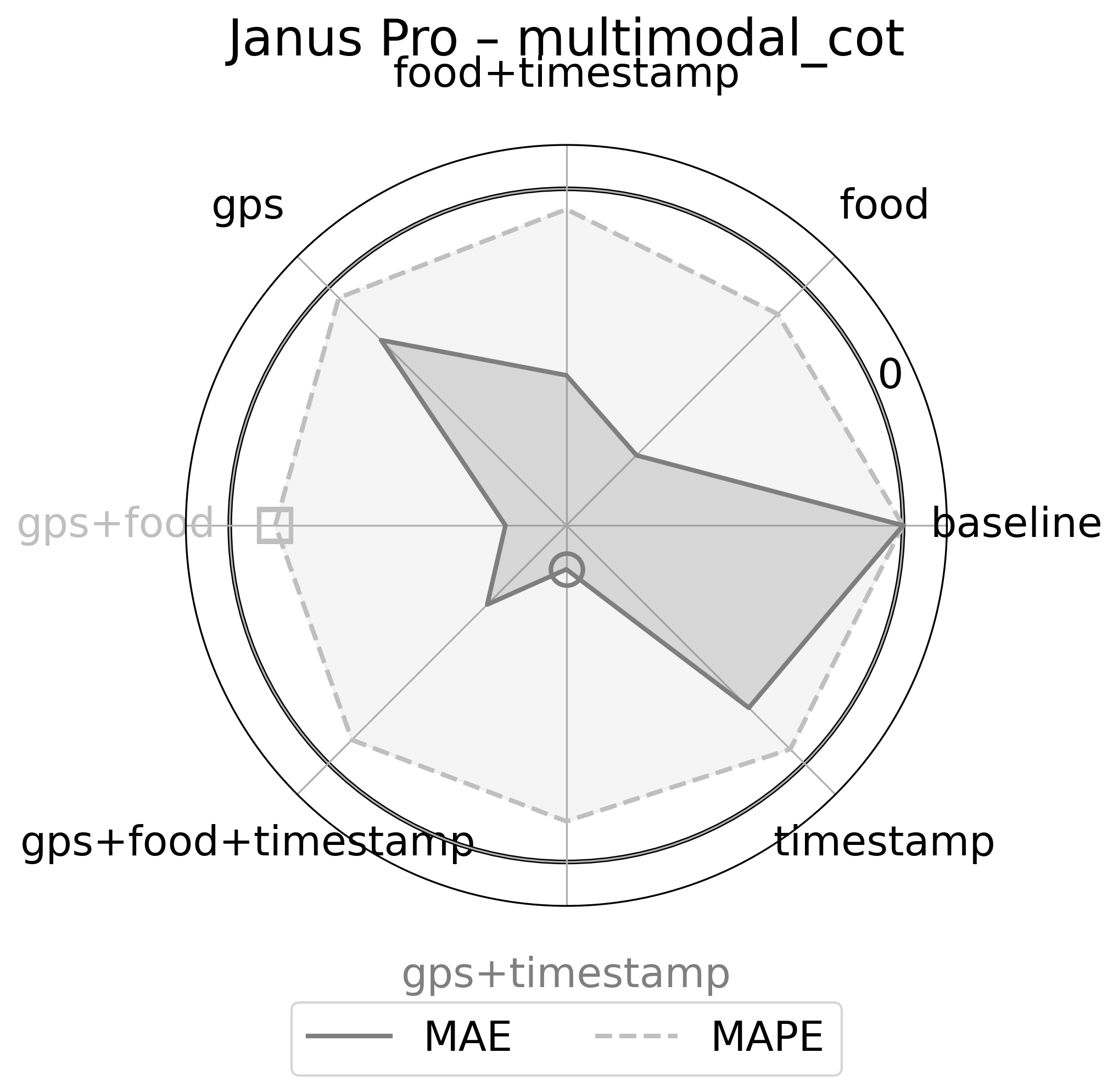}
    \caption{Janus-Pro Multimodal CoT}
    \label{fig:img4}
  \end{subfigure}
  \begin{subfigure}[t]{0.195\linewidth}
    \centering
    \includegraphics[width=\linewidth]{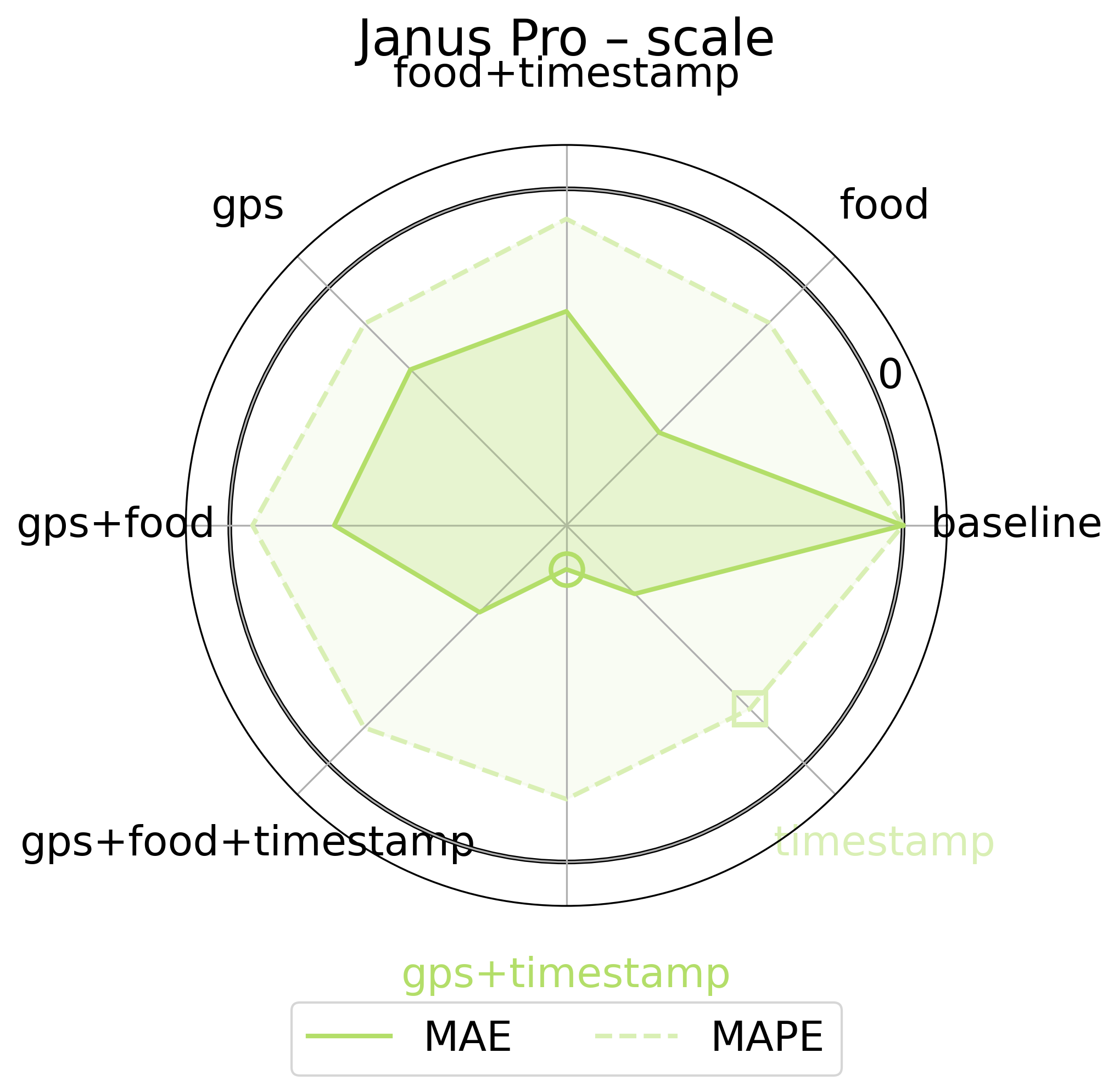}
    \caption{Janus-Pro Scale-Hint}
    \label{fig:img4}
  \end{subfigure}

  \begin{subfigure}[t]{0.195\linewidth}
    \centering
    \includegraphics[width=\linewidth]{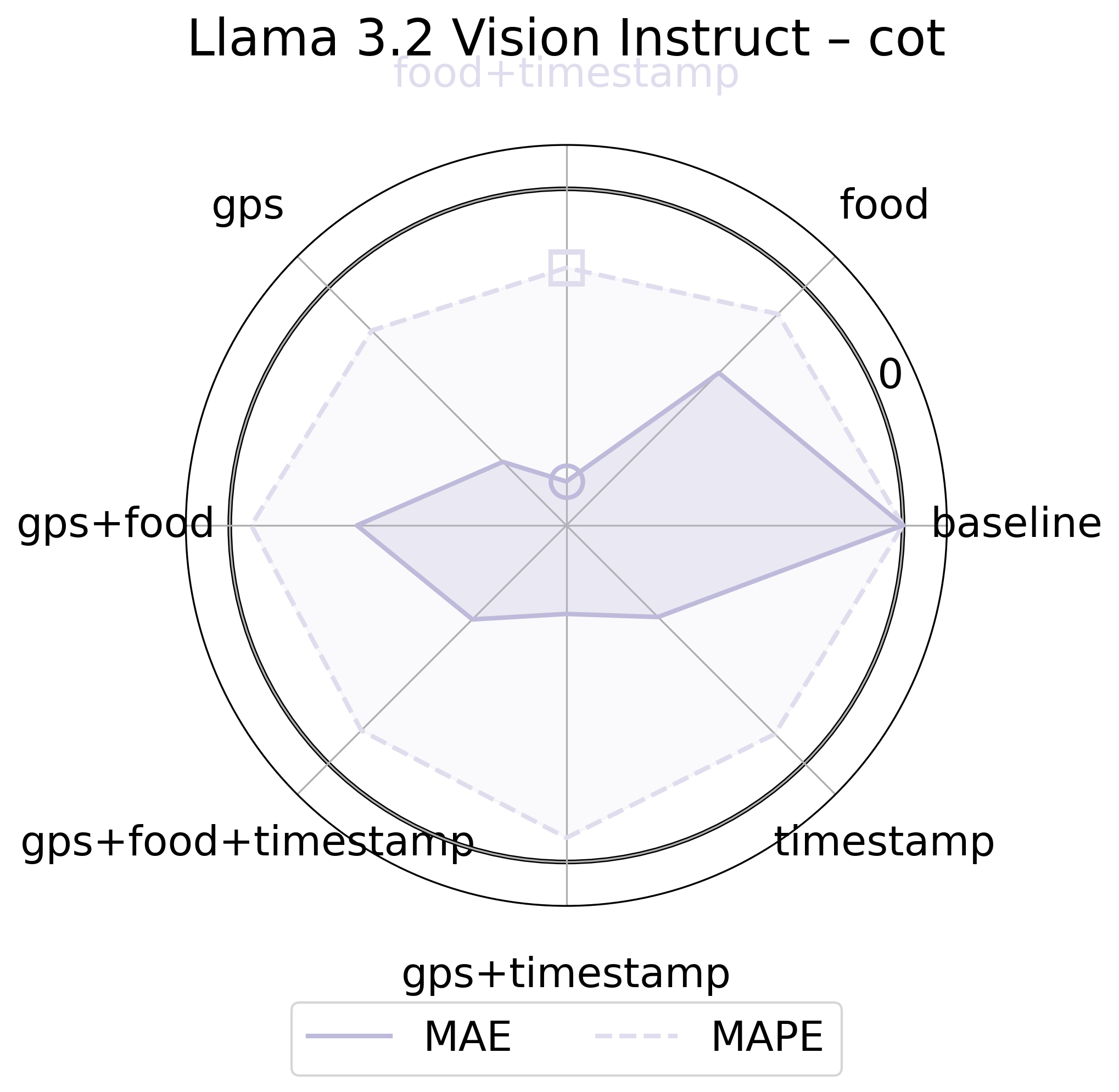}
    \caption{LLaMA-3.2-Vision-Instruct Chain-of-Thought}
    \label{fig:img1}
  \end{subfigure}
  \begin{subfigure}[t]{0.195\linewidth}
    \centering
    \includegraphics[width=\linewidth]{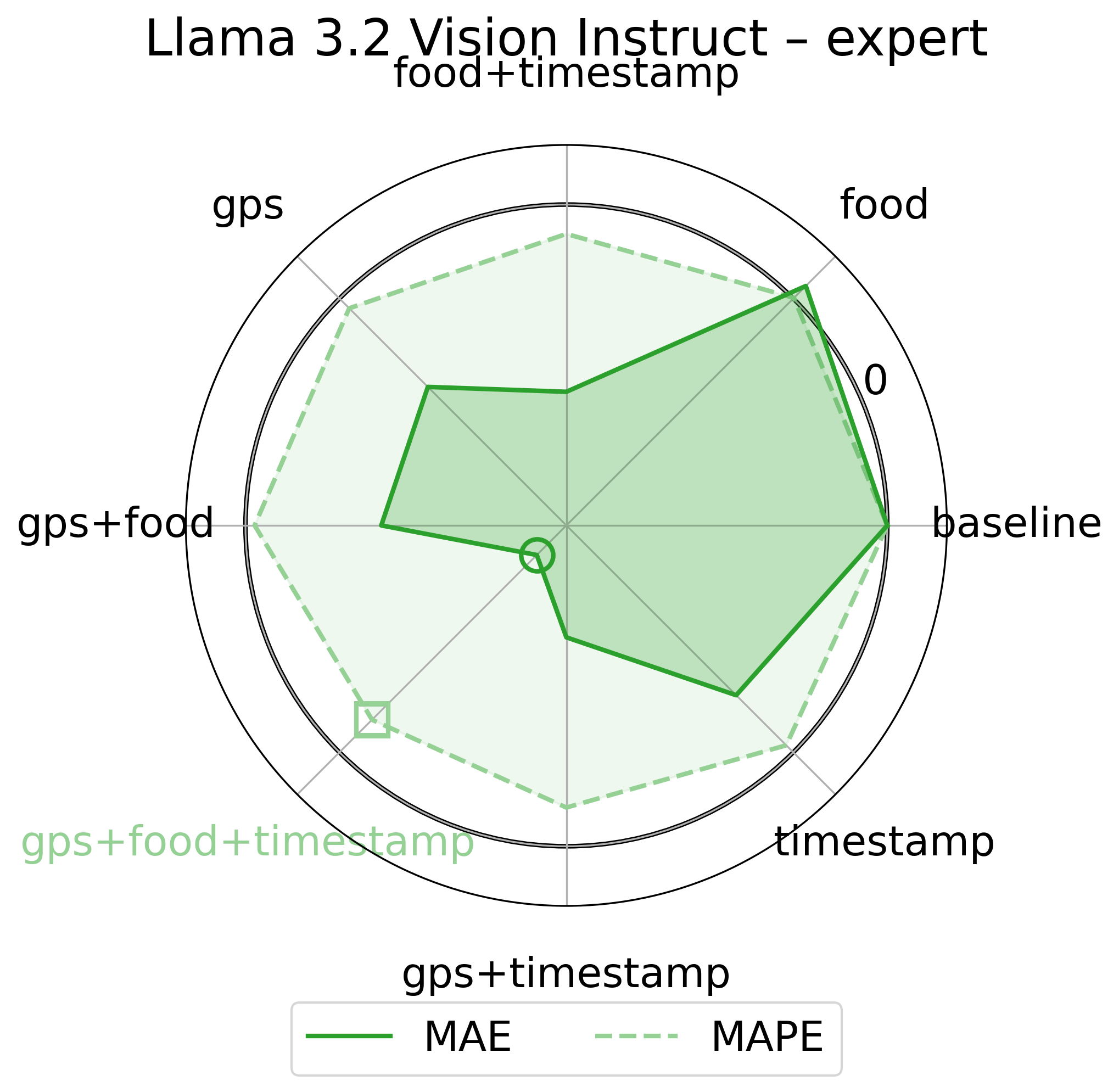}
    \caption{LLaMA-3.2-Vision-Instruct Expert Persona}
    \label{fig:img2}
  \end{subfigure}
  \begin{subfigure}[t]{0.195\linewidth}
    \centering
    \includegraphics[width=\linewidth]{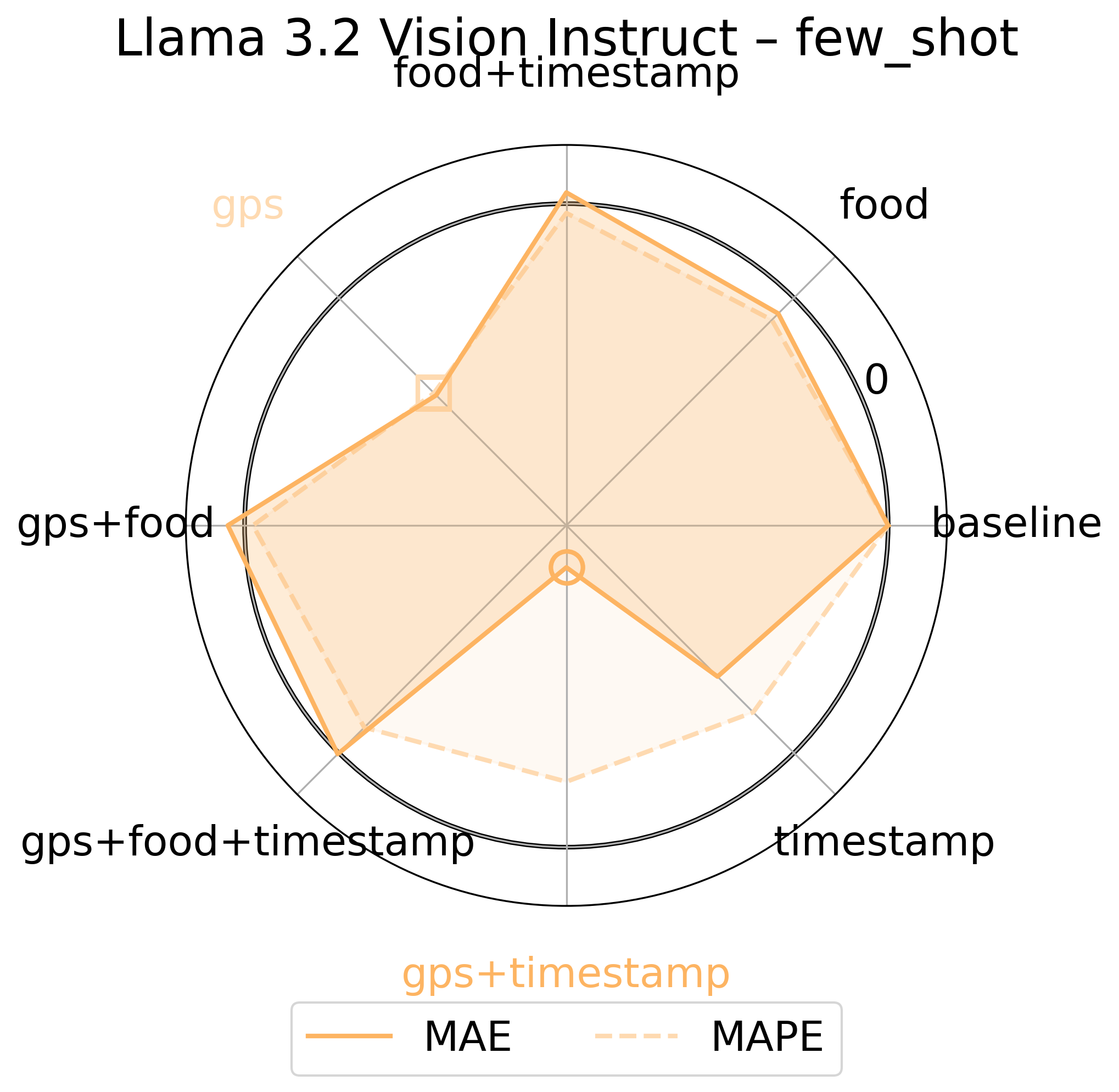}
    \caption{LLaMA-3.2-Vision-Instruct Few-Shot}
    \label{fig:img3}
  \end{subfigure}
  \begin{subfigure}[t]{0.195\linewidth}
    \centering
    \includegraphics[width=\linewidth]{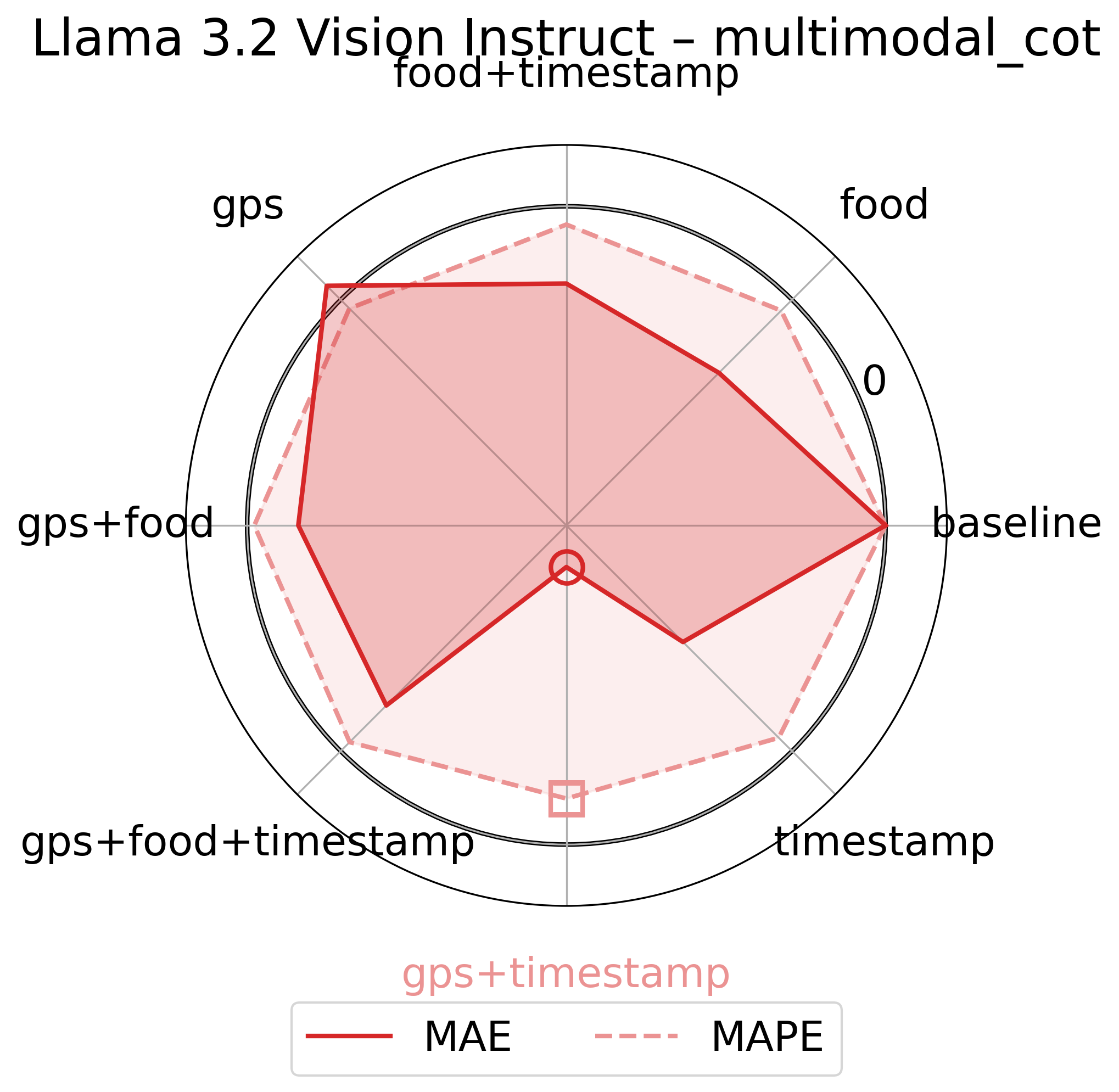}
    \caption{LLaMA-3.2-Vision-Instruct Multimodal CoT}
    \label{fig:img4}
  \end{subfigure}
  \begin{subfigure}[t]{0.195\linewidth}
    \centering
    \includegraphics[width=\linewidth]{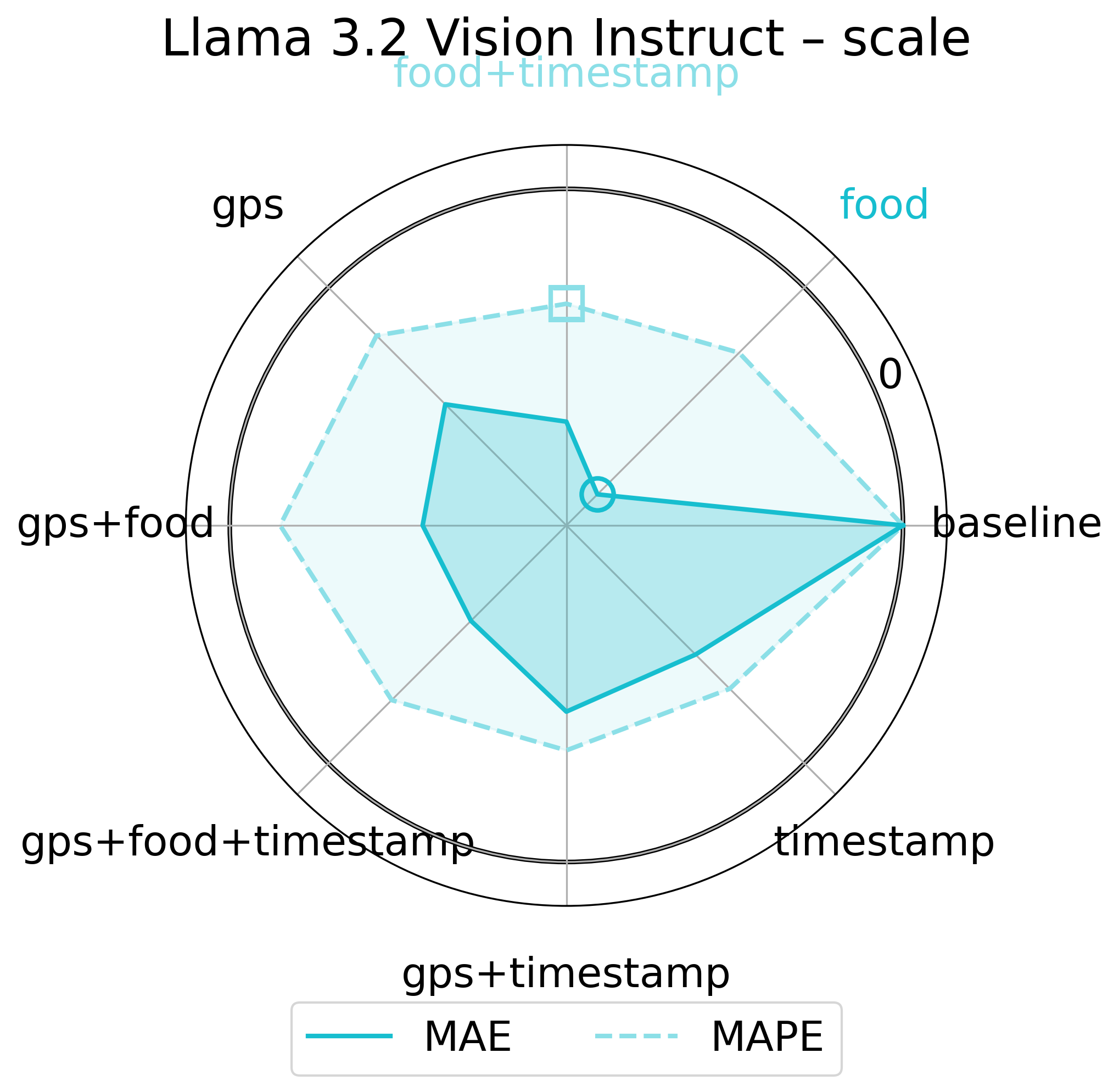}
    \caption{LLaMA-3.2-Vision-Instruct Scale-Hint}
    \label{fig:img4}
  \end{subfigure}

  \begin{subfigure}[t]{0.195\linewidth}
    \centering
    \includegraphics[width=\linewidth]{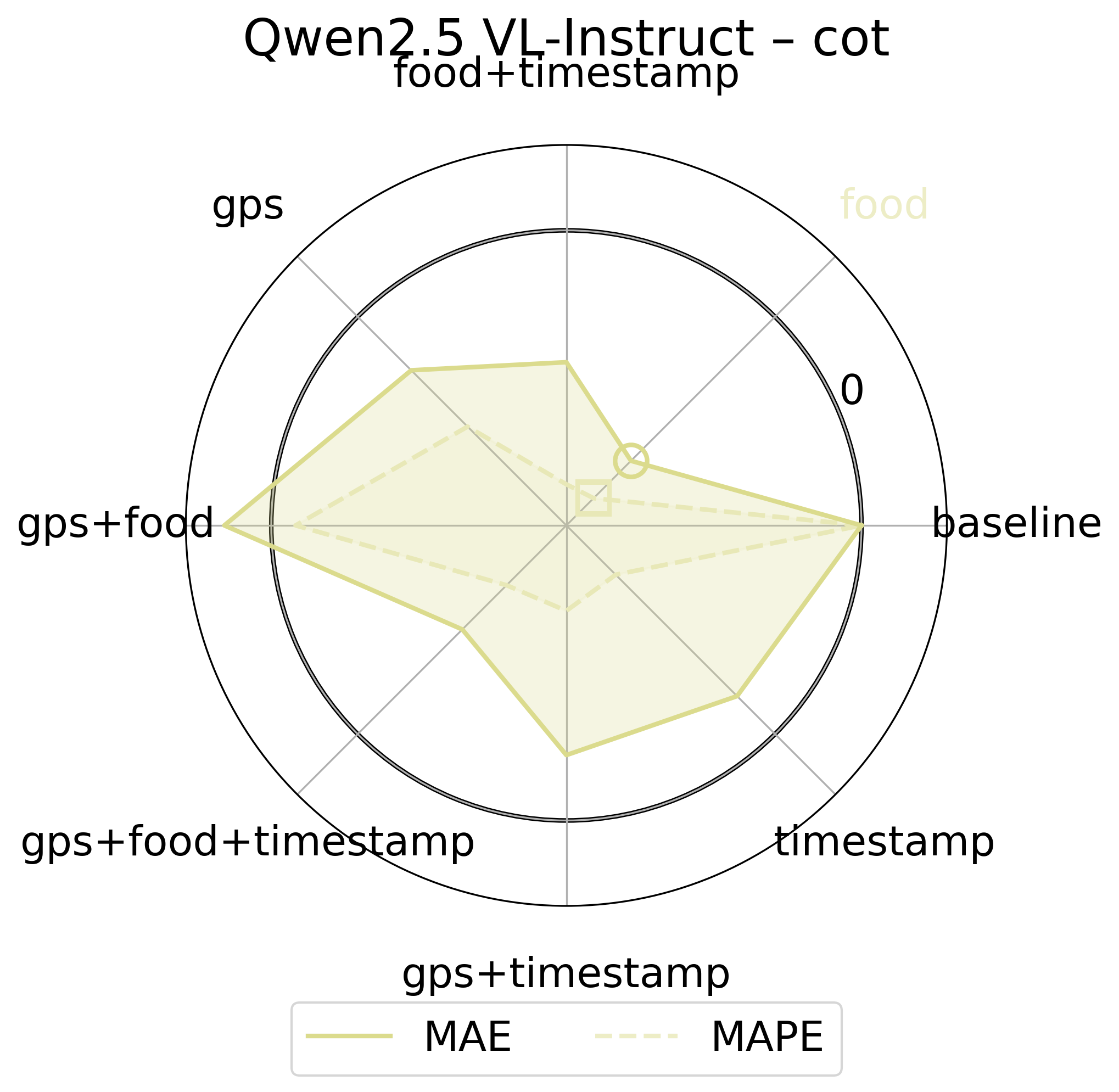}
    \caption{Qwen2.5-VL Chain-of-Thought}
    \label{fig:img1}
  \end{subfigure}
  \begin{subfigure}[t]{0.195\linewidth}
    \centering
    \includegraphics[width=\linewidth]{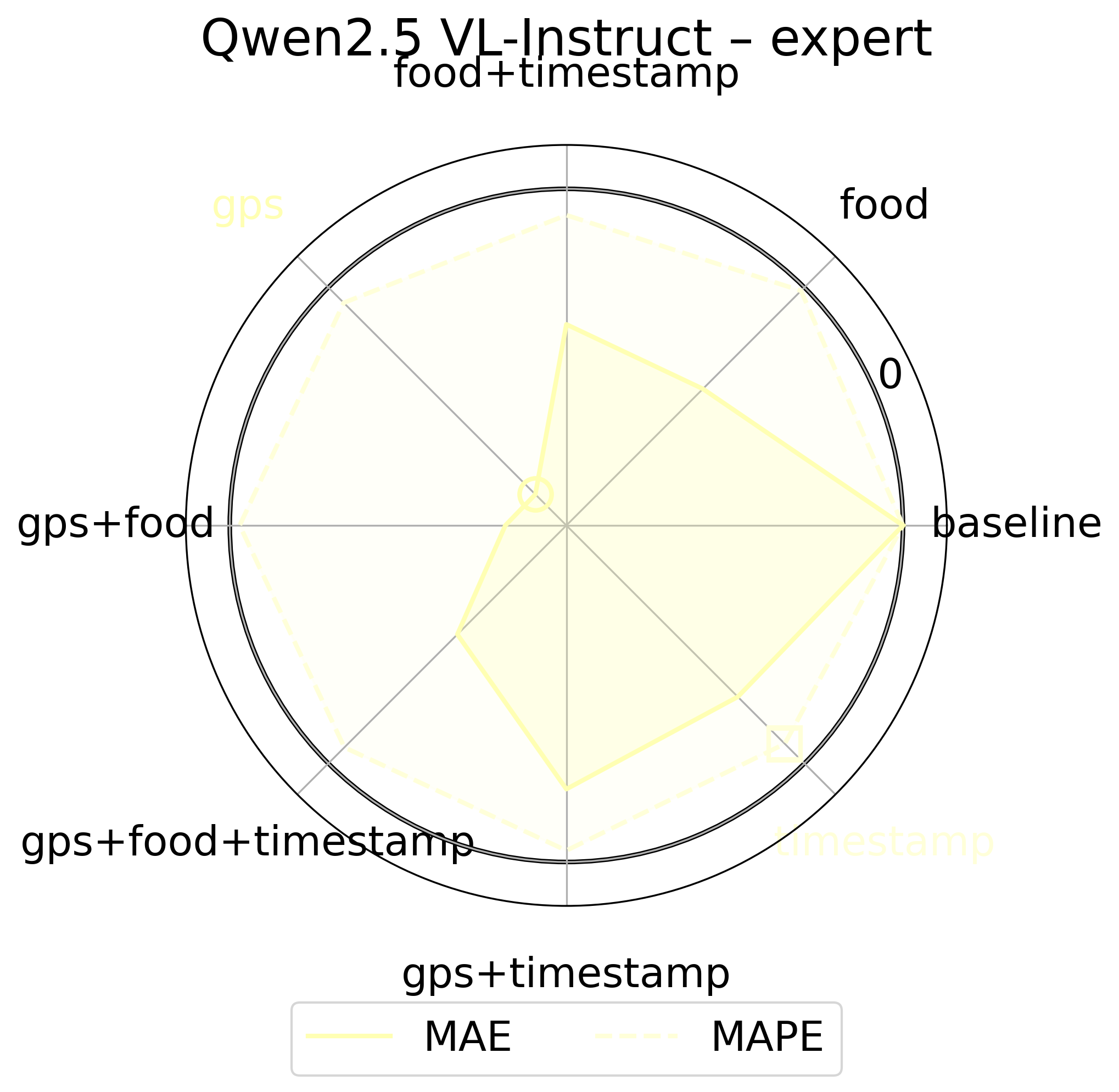}
    \caption{Qwen2.5-VL Expert Persona}
    \label{fig:img2}
  \end{subfigure}
  \begin{subfigure}[t]{0.195\linewidth}
    \centering
    \includegraphics[width=\linewidth]{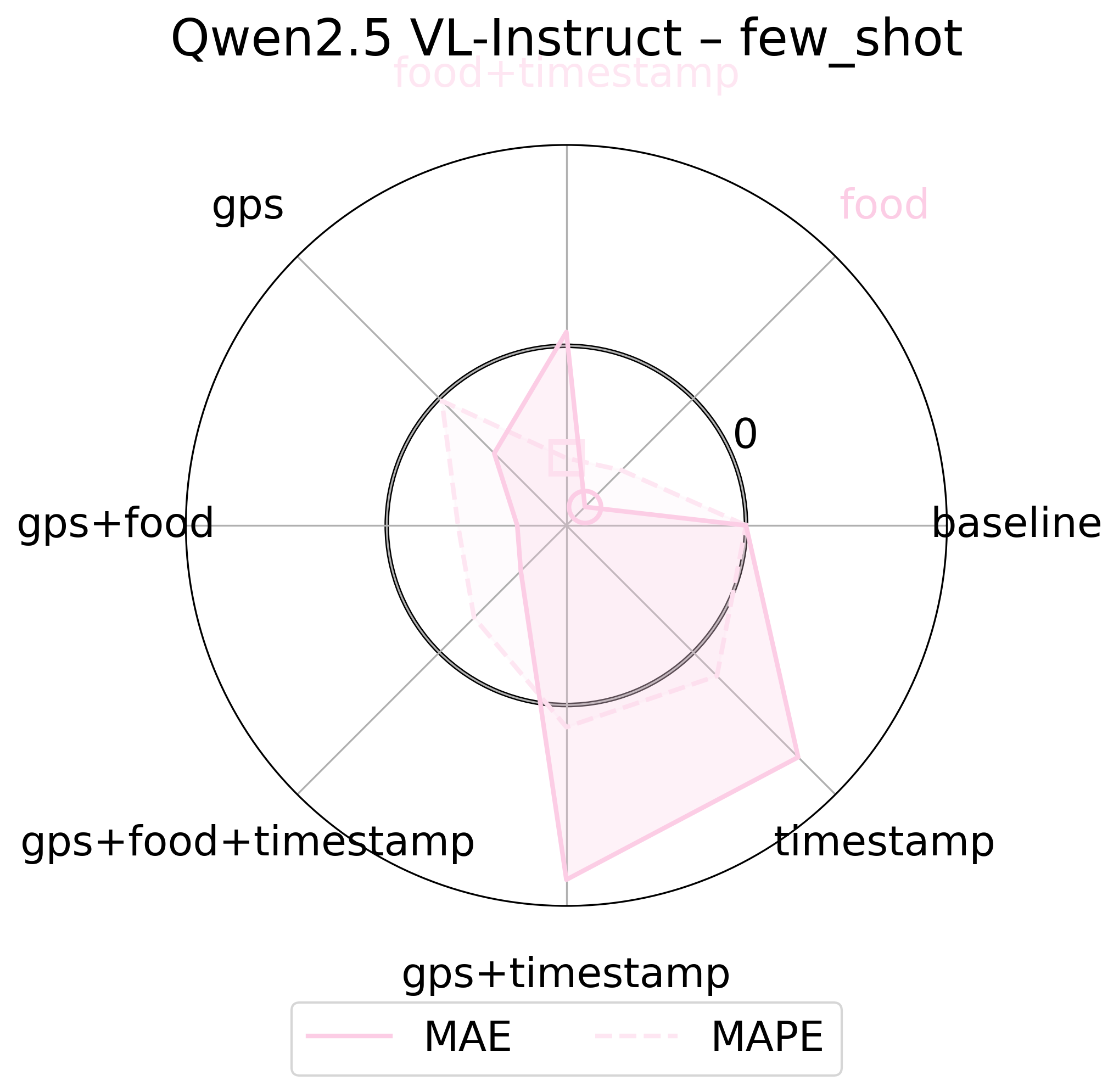}
    \caption{Qwen2.5-VL Few-Shot}
    \label{fig:qwen-few-shot}
  \end{subfigure}
  \begin{subfigure}[t]{0.195\linewidth}
    \centering
    \includegraphics[width=\linewidth]{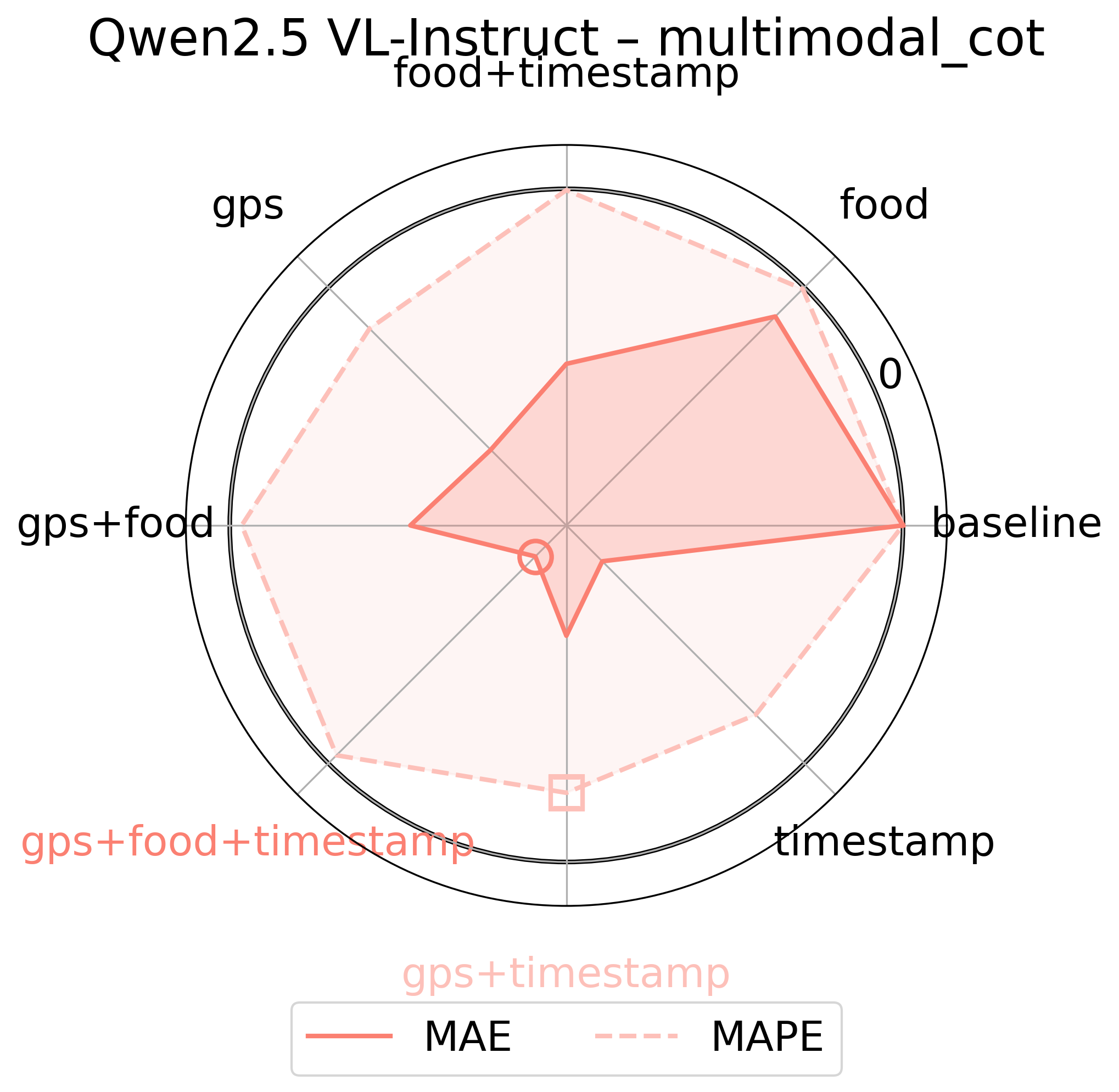}
    \caption{Qwen2.5-VL Multimodal CoT}
    \label{fig:img4}
  \end{subfigure}
  \begin{subfigure}[t]{0.195\linewidth}
    \centering
    \includegraphics[width=\linewidth]{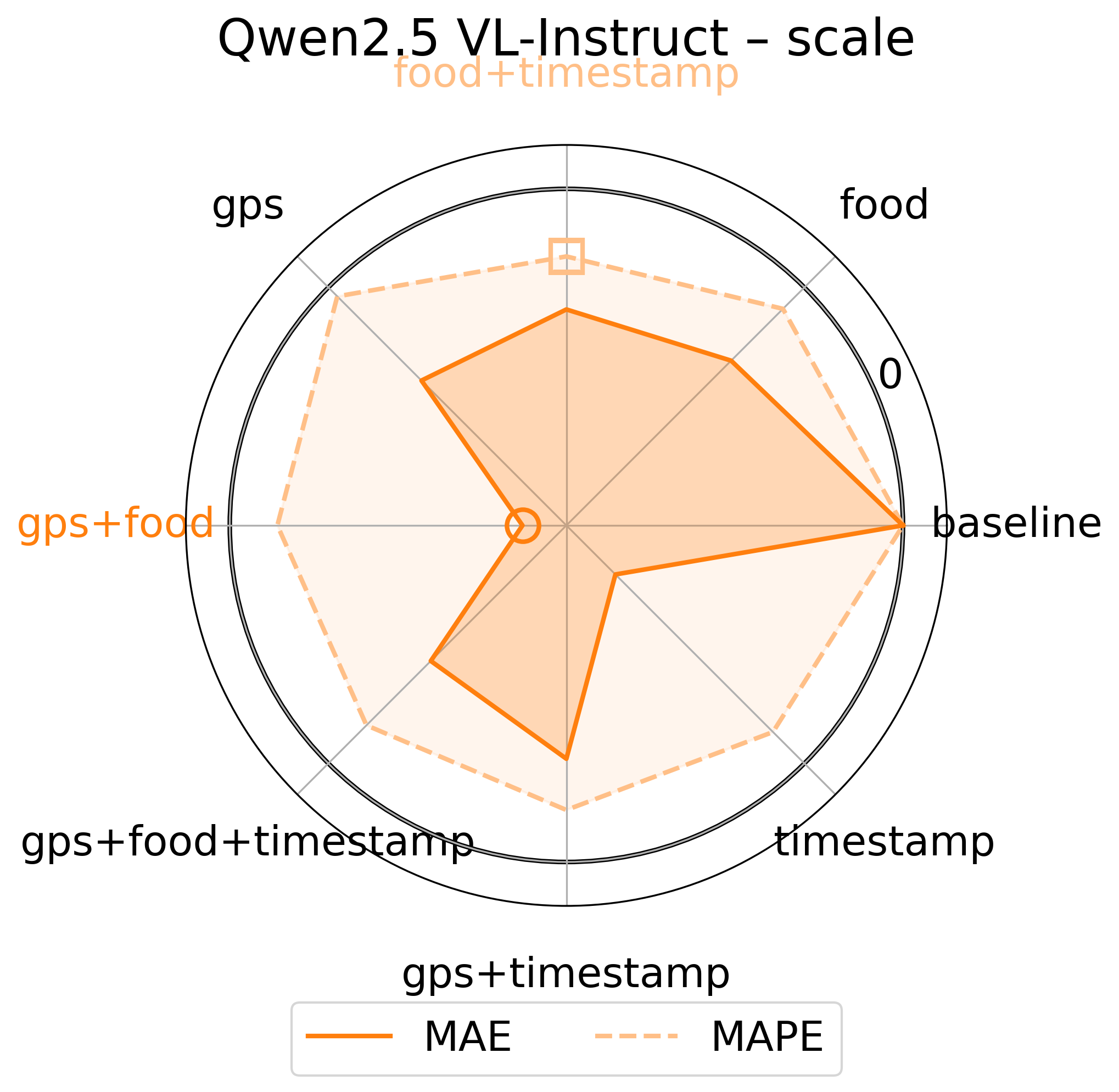}
    \caption{Qwen2.5-VL Scale-Hint}
    \label{fig:img4}
  \end{subfigure}

  \caption{\textbf{Averaged Experiment 2 Results For Open-Weight Models.} MAE (solid lines) and MAPE (dashed lines) are plotted for various contextual metadata combinations for a given open-weight model. Each spoke represents an error in a metadata combination; the closer proximity to the center signifies a reduction in error relative to the baseline prompt. Colored markers denote the \textsc{Best-Metadata} configuration for each metric.}
  \label{fig:exp_2_open_weight_models}
\end{figure*}

Figures~\ref{fig:exp_2_closed_weight_models} and~\ref{fig:exp_2_open_weight_models} offer more detailed visualizations of metadata's effects on nutrient MAE and MAPE across all reasoning modifiers and models. Consistent with the findings in Figure~\ref{fig:exp_1_all_models}, including metadata generally leads to a reduction in nutrient MAE and MAPE. However, it is important to note the exceptionally high MAE values for \texttt{gps+timestamp} and \texttt{timestamp} in the \texttt{few\_shot} configuration of Qwen2.5, as depicted in Figure~\ref{fig:qwen-few-shot}.

\subsection{Prompt Templates}
\label{app:prompt_templates}

To facilitate reproducibility, we enumerate the exact text snippets that form our prompts. Each prompt is assembled from three components, presented in order of appearance: \emph{(i)} a baseline nutrition–analysis request, \emph{(ii)} an optional reasoning-modifier prefix, and \emph{(iii)} optional metadata-augmentation fragments. The metadata fragments are appended \emph{after} the baseline request, whereas the reasoning modifiers are prepended \emph{before} it.  

Table~\ref{tab:baseline_prompt} shows the baseline request that is always included. The available reasoning-modifier prefixes are listed in Table~\ref{tab:reason_templates}, and the metadata fragments appear in Table~\ref{tab:meta_templates}. At run time, angle-bracket placeholders (e.g., \texttt{\textless{}Month DD, YYYY\textgreater{}}) are replaced with study-specific values before the prompt is sent to the model.

\begin{table}[htbp]
\caption{Baseline nutrition-analysis prompt (always included).}
\label{tab:baseline_prompt} 
\footnotesize
\setlength{\tabcolsep}{4pt} 
\begin{tabularx}{\columnwidth}{@{}X@{}} 
\toprule
\textbf{Prompt (verbatim)}\\
\midrule
\texttt{Analyze this food image and estimate the following nutritional content: calories (kcal), protein (g), carbohydrates (g), fat (g), portion (g).\newline
Provide a single point estimate (no ranges) for each metric, using exactly this format:\newline
kcal: \textless{}number\textgreater{}, protein: \textless{}number\textgreater{}, carbs: \textless{}number\textgreater{}, fat: \textless{}number\textgreater{}, portion: \textless{}number\textgreater{}\newline
Do not include any ranges or hyphens, only one number per metric. ONLY THE NUMERIC ESTIMATES SHOULD BE INCLUDED. THE UNITS FOR EACH METRIC WILL BE ASSUMED; DO NOT INCLUDE THE UNITS.} \\
\bottomrule
\end{tabularx}
\end{table}

\begin{table}[htbp]
\caption{Metadata-augmentation fragments (appended \emph{after} the baseline
prompt).  Angle brackets denote run-time substitution.}
\label{tab:meta_templates}
\footnotesize
\setlength{\tabcolsep}{4pt}
\begin{tabularx}{\columnwidth}{@{}lX@{}}
\toprule
\textbf{Flag} & \textbf{Prompt fragment (verbatim)}\\ \midrule
\texttt{gps} &
\texttt{Location context: \textless{}venue\textgreater{}, \textless{}city\textgreater{}, \textless{}country\textgreater{}.}\\[2pt]
\texttt{timestamp} &
\texttt{Time context: \textless{}Month DD, YYYY at hh:mm AM/PM\textgreater{} (\textless{}MealType\textgreater{}, \textless{}DayType\textgreater{}).}\\[2pt]
\texttt{food} &
\texttt{Visible foods include: \textless{}comma-separated list\textgreater{}.}\\
\bottomrule
\end{tabularx}
\end{table}

\begin{table}[htbp]
\caption{Reasoning-modifier prefixes (prepended \emph{before} metadata or baseline prompt).}
\label{tab:reason_templates}
\footnotesize
\setlength{\tabcolsep}{4pt}
\begin{tabularx}{\columnwidth}{@{}lX@{}}
\toprule
\textbf{Flag} & \textbf{Prompt fragment (verbatim)}\\ \midrule
\texttt{expert} &
\texttt{You are a nutrition expert.}\\
\addlinespace[4pt] 

\texttt{few\_shot} &
\begin{minipage}[t]{\linewidth}\footnotesize\ttfamily
Analyze the attached meal photo. Here are examples:\newline
Example~1:\newline
\,[IMAGE: two slices of pepperoni pizza]\newline
\,Output: Calories: \textasciitilde500 kcal; Protein: \textasciitilde20 g; Fat: \textasciitilde22 g; Carbs: \textasciitilde50 g\newline
Example~2:\newline
\,[IMAGE: a bowl of salad with chicken and avocado]\newline
\,Output: Calories: \textasciitilde350 kcal; Protein: \textasciitilde30 g; Fat: \textasciitilde15 g; Carbs: \textasciitilde20 g\newline[2pt]
Now analyze the next image.
\end{minipage}\\
\addlinespace[4pt] 

\texttt{multimodal\_cot} &
\begin{minipage}[t]{\linewidth}\footnotesize\ttfamily
Analyze this meal photo stepwise to avoid mistakes.\newline
1.\ List the foods present and their approximate amounts.\newline
2.\ For each item, estimate calories, protein, carbs, fat, and portion size in this exact format (calories-item: X, protein-item: Y, carbs-item: Z, fat-item: W, portion-item: P).\newline
3.\ Compute and provide the total values in the format described below.
\end{minipage}\\
\addlinespace[4pt] 

\texttt{cot} &
\texttt{Let's think step by step about the nutritional and portion estimation.}\\
\addlinespace[4pt] 

\texttt{scale} &
\texttt{Before estimating, please describe what other objects or context you see in the image that could help gauge the size or scale of the food.}\\
\bottomrule
\end{tabularx}
\end{table}